%% file: main.tex
\documentclass[10pt,twocolumn,letterpaper]{article}

\usepackage[pagenumbers]{cvpr} 
\usepackage{multirow}

\usepackage[norelsize, linesnumbered, ruled, lined, boxed, commentsnumbered]{algorithm2e}

\usepackage[ruled]{algorithm2e}
\usepackage{bbm}
\usepackage{bm}
\usepackage{color}
\usepackage{multirow}
\usepackage{array}
\usepackage{booktabs}
\usepackage{diagbox}
\usepackage{colortbl}
\usepackage{tabularx}
\usepackage[accsupp]{axessibility} 

\usepackage{graphicx}
\usepackage{amsmath}
\usepackage{amsthm}
\usepackage{amssymb}

\definecolor{cvprblue}{rgb}{0.21,0.49,0.74}
\usepackage[pagebackref,breaklinks,colorlinks,citecolor=cvprblue]{hyperref}
\usepackage{diagbox}
\def\paperID{4144} 
\def\confName{CVPR}
\def\confYear{2024}

\usepackage{cvpr}              
\input{preamble}

\usepackage{multirow}

\usepackage[norelsize, linesnumbered, ruled, lined, boxed, commentsnumbered]{algorithm2e}

\usepackage[ruled]{algorithm2e}
\usepackage{bbm}
\usepackage{bm}
\usepackage{color}
\usepackage{multirow}
\usepackage{array}
\usepackage{booktabs}
\usepackage{diagbox}
\usepackage{colortbl}
\usepackage{tabularx}
\usepackage[accsupp]{axessibility} 

\usepackage{graphicx}
\usepackage{amsmath}
\usepackage{amsthm}
\usepackage{amssymb}

\definecolor{cvprblue}{rgb}{0.21,0.49,0.74}
\usepackage[pagebackref,breaklinks,colorlinks,citecolor=cvprblue]{hyperref}
\usepackage{diagbox}
\def\paperID{4144} 
\def\confName{CVPR}
\def\confYear{2024}

\title{SimAC: A Simple Anti-Customization Method for Protecting Face Privacy against Text-to-Image Synthesis of Diffusion Models}

\author{
Feifei Wang\textsuperscript{\rm 1,2,}\thanks{Work done during an internship in Alibaba Cloud.},\quad
Zhentao Tan\textsuperscript{\rm 2,1},\quad
Tianyi Wei\textsuperscript{\rm 1},\quad
Yue Wu\textsuperscript{\rm 2},\quad
Qidong Huang\textsuperscript{\rm 1,}\thanks{Corresponding author.} \\
\textsuperscript{\rm 1}University of Science and Technology of China \quad
\textsuperscript{\rm 2}Alibaba Cloud \\
{\tt\small \{wangfeifei@, tzt@, bestwty@, hqd0037@\}mail.ustc.edu.cn}\quad
{\tt\small matthew.wy@alibaba-inc.com}
}

\begin{document}
\maketitle

\begin{abstract}
Despite the success of diffusion-based customization methods on visual content creation, increasing concerns have been raised about such techniques from both privacy and political perspectives. 
To tackle this issue, several anti-customization methods have been proposed in very recent months, predominantly grounded in adversarial attacks. 
Unfortunately, most of these methods adopt straightforward designs, such as end-to-end optimization with a focus on adversarially maximizing the original training loss, thereby neglecting nuanced internal properties intrinsic to the diffusion model, and even leading to ineffective optimization in some diffusion time steps.
In this paper, we strive to bridge this gap by undertaking a comprehensive exploration of these inherent properties, to boost the performance of current anti-customization approaches. 
Two aspects of properties are investigated: 
1) We examine the relationship between time step selection and the model's perception in the frequency domain of images and find that lower time steps can give much more contributions to adversarial noises. 
This inspires us to propose an adaptive greedy search for optimal time steps that seamlessly integrates with existing anti-customization methods. 
2) We scrutinize the roles of features at different layers during denoising and devise a sophisticated feature-based optimization framework for anti-customization.
Experiments on facial benchmarks demonstrate that our approach significantly increases identity disruption, thereby protecting user privacy and copyright. 
Our code is available at: \href{https://github.com/somuchtome/SimAC}{https://github.com/somuchtome/SimAC}.

\end{abstract}

\section{Introduction}\label{1}
\label{sec:intro}

\begin{figure}[t]
\centering
\includegraphics[width=1.0\linewidth]{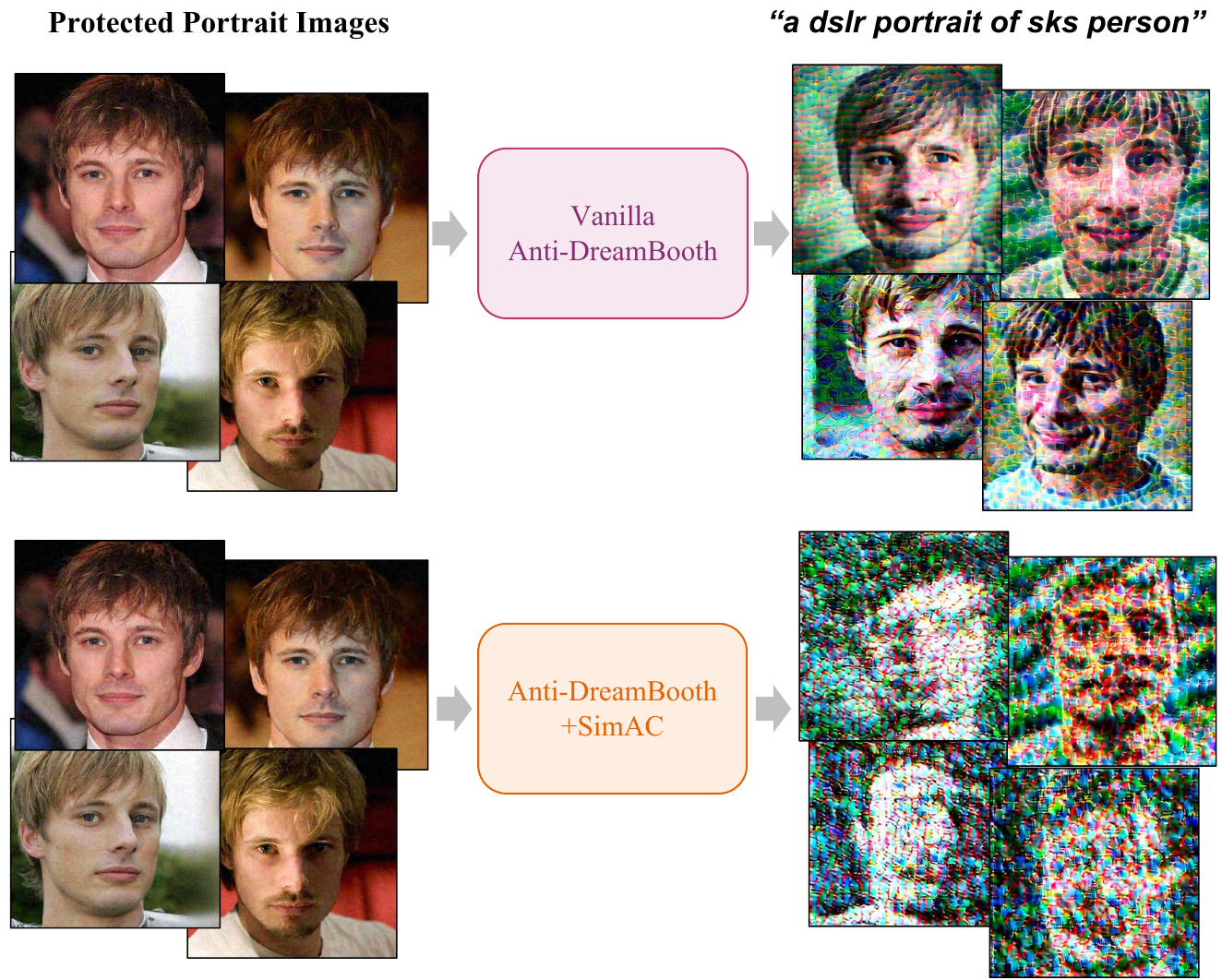}
\vspace{-2em}
\caption{Comparison between Anti-DreamBooth before and after adding SimAC. Our method further boosts its ability to de-identity.}
\label{fig:teaser}
\end{figure}


Latent Diffusion model (LDM) \cite{sohl2015deep,ho2020denoising, DBLP:conf/iclr/SongME21} has been recently proven as a strong paradigm for photorealistic visual content generation. 
The emergence of open-source Stable Diffusion encourages users to explore creative possibilities with LDMs.  
Users only need to provide several images representing the same subject along with a rare identifier to customize their diffusion
models \cite{gal2022image,ruiz2023dreambooth,kumari2023multi, hu2021lora}. 
After fine-tuning the model or optimizing the text embedding of rare identifiers, it can flexibly generate high-quality images containing the specified object.

The convenience of customizing large-scale text-to-image models also allows the malicious users to generate forgery images that violate the truth. 
Some of them may use such technique to steal others' painting styles and generate new painting content. In addition without permission. 
While some of them may collect one's portrait photos that are published on the social platform, and generate fake images of this person through customization. 
In other words, the lawbreakers can easily produce fake news that contains celebrity photos \cite{trump}, greatly misleading the public. 
The infringement poses threats to user privacy and intellectual property. 
Hence, it is essential to devise effective countermeasures to safeguard users against such malicious usage.



\begin{figure*}
    \centering
    \begin{minipage}{0.31\linewidth}
    \centering
    \includegraphics[width=1.0\linewidth]{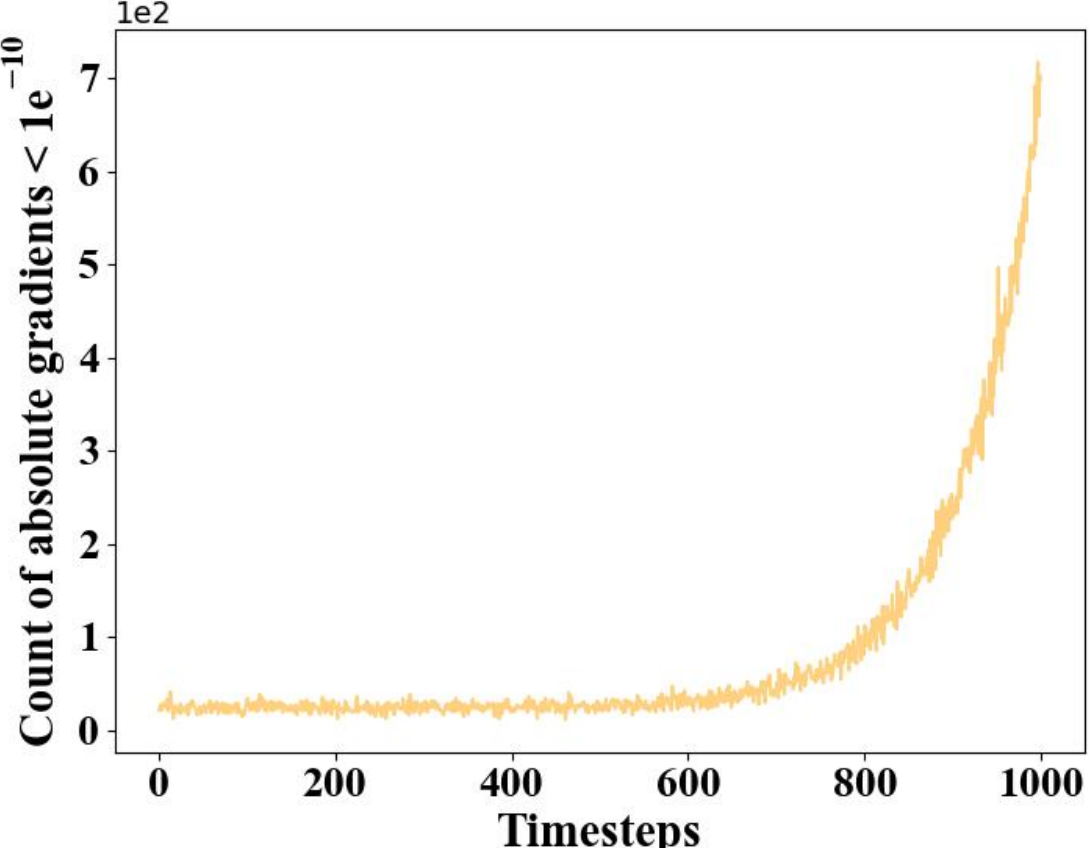}
    \centerline{\scriptsize \textbf{(a) the number of absolute gradients \boldmath$<1e^{-10}$}}
    \label{fig:1e-10}
    \end{minipage}
    \hfill
    \begin{minipage}{0.31\linewidth}
    \centering
    \includegraphics[width=1.0\linewidth]{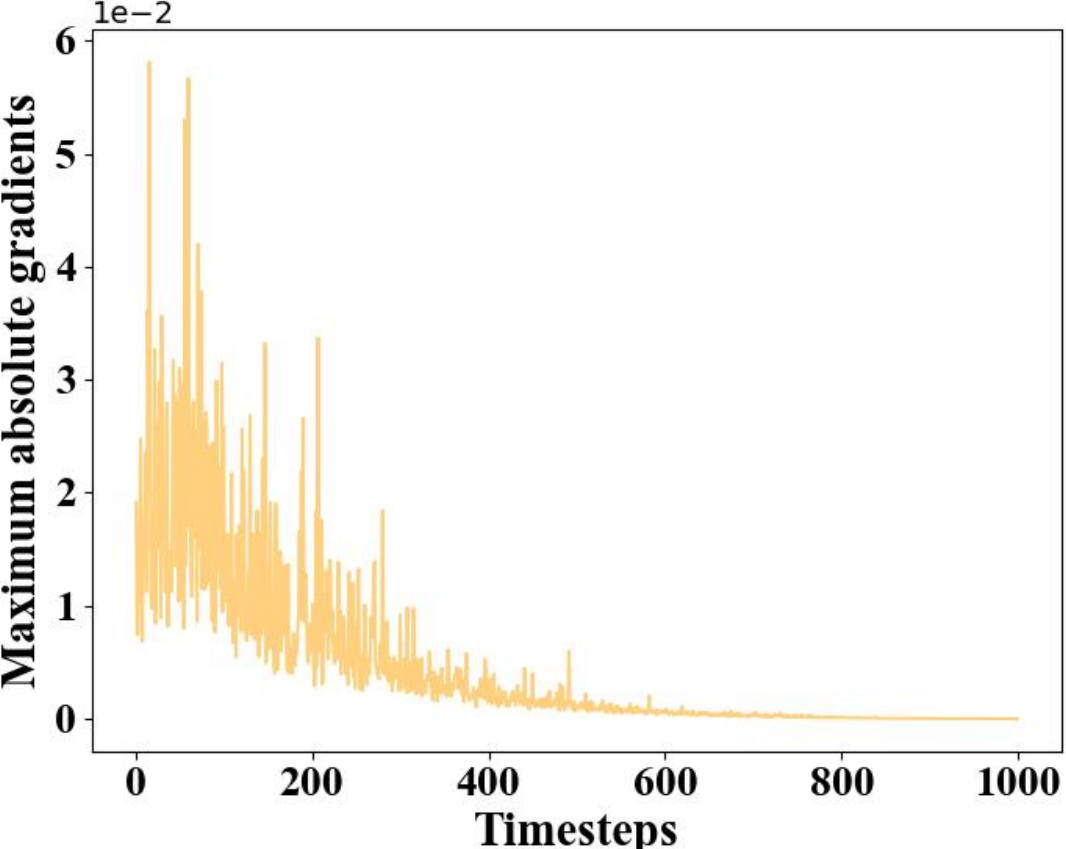}
    \centerline{\scriptsize \textbf {(b) the maximum absolute gradients}}
    \label{fig:max}
    \end{minipage}
    \hfill
    \begin{minipage}{0.35\linewidth}
    \centering
    \includegraphics[width=1.0\linewidth]{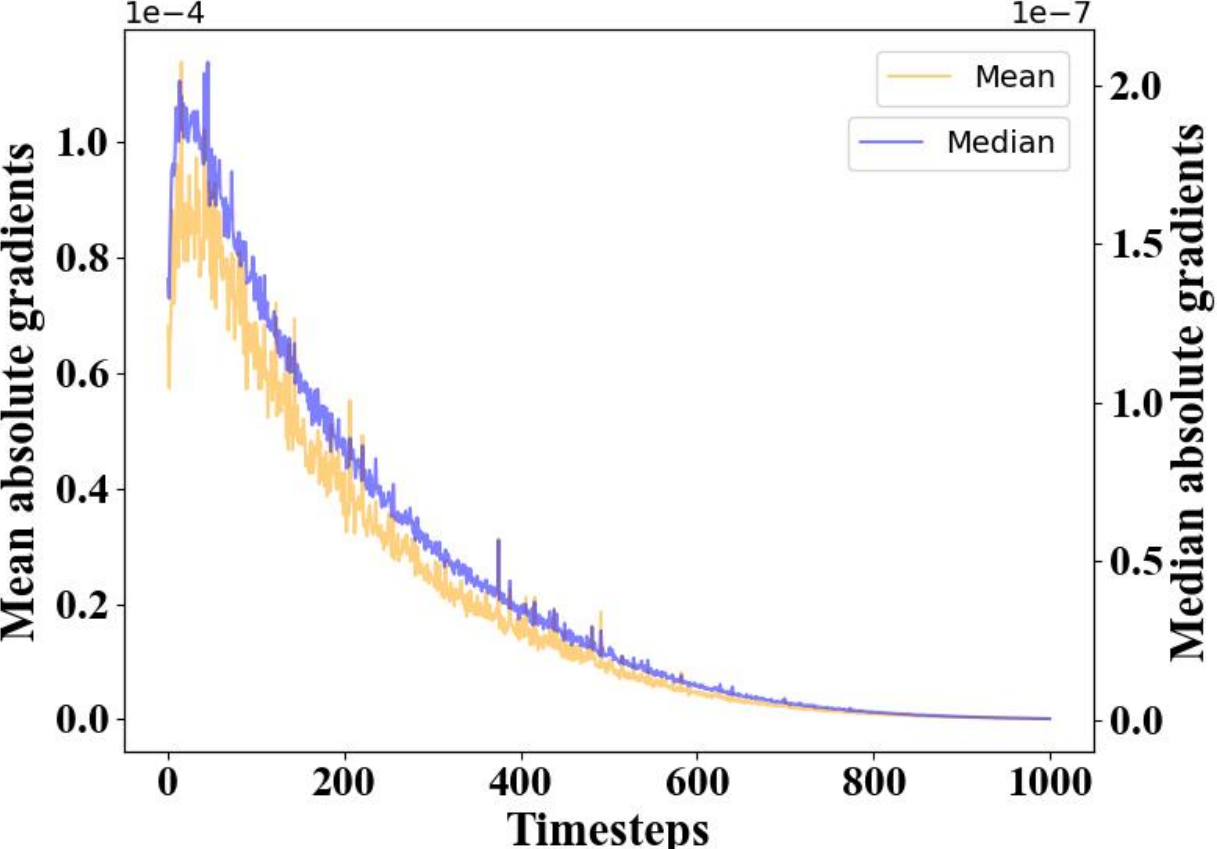}
    \centerline{\scriptsize \textbf {(c) the mean and median of the absolute gradients}}
    \label{fig:mean}
    \end{minipage}
    \hfill
    \vspace{-1.5em}
    \captionof{figure}{Distribution of Anti-DreamBooth attack gradients on different diffusion timesteps, where (a) counts the number of absolute gradients below the threshold $1e^{-10}$ at each timestep, (b) presents the maximum absolute gradients at different timesteps, and (c) demonstrates how the mean and median of the absolute  gradients change over timesteps. 
    Apparently, the absolute gradients shows great discrepancy on the varying timesteps and nearly zero values appear at large timesteps, which leads to ineffective noise optimization.}
    \vspace{1.5em}
    \label{fig:grad}
\end{figure*}

 Anti-customization methods are generally based on adversarial attack\cite{szegedy2013intriguing,goodfellow2015explaining} against text-to-image models.
AdvDM \cite{DBLP:conf/icml/LiangWHZXSXMG23} is the pioneering work that uses adversarial noise to protect user images from being customized by diffusion models. 
It ingeniously combines the diffusion model with adversarial samples, firstly achieving user privacy protection. 
Anti-DreamBooth \cite{van2023anti} further enhances the protective effect by employing alternate training. 
Both of them randomly select the timesteps from (0, \textit{max denoising steps}) regarding the noise added to LDM latent, and directly employ the maximization of the LDM training loss as the optimization objective.

However, as showcased in Figure~\ref{fig:grad}, we notice that the gradient of the perturbed images is quite small and even \textbf{zero} when the randomly sampled timesteps are relatively large, \ie, the noisy latent is closer to Gaussian noise. 
This implies that within the limited steps of an adversarial attack, a large portion of steps are ineffective since the perturbed image cannot optimize the objective with such zero gradients, leading to a decrease in both protection effectiveness and efficiency. 
Hence, the images protected by these current methods, when customized, still allow the model to capture many details from user-input images and leak the user's privacy, as shown in Figure~\ref{fig:teaser}.
 
The fundamental reason is that these methods fail to combine the adversarial attacks with properties inherent in diffusion models. 
We aim to conduct an in-depth analysis of why the gradients of perturbed images have such discrepancy at different time steps and how diffusion models perceive input images at different intermediate layers. 
Then, based on the in-depth analysis,  we propose improvements to existing customization methods from both the temporal and feature dimensions. 

To observe the relationship between timesteps and the gradient of the perturbed images, we first reconstruct images predicted from noisy latents at different time steps and compare them with the input images. 
Considering regular adversarial noises mainly affect the high frequency of images, our comparison is conducted in the frequency domain, aiming at investigating whether the model's perception lies on lower-level or higher-level. 
We observe that, the model focuses on the higher frequency components of images when selecting smaller time steps, and \textit{vice versa}. 
Therefore, introducing adversarial noises at larger timesteps is ineffective, since the subtle changes perturbed on images can hardly affect the low-frequency of the generated images. 
To improve the effectiveness, we propose an adaptive timestep selection method to find optimal time intervals, where we iteratively update the range of selection in a greedy way. 
To explore the effect of different layers' features in the U-Net decoder during denoising, we employ PCA (Principal Component Analysis) to visualize them and show the discrepancy. 
We clearly find that within the decoder, the feature extraction gradually shifts from low-frequency to high-frequency as the layer goes deeper. 
It indicates the higher layers concentrates more on the texture of images, while the lower layers focus on the structure. 
Consequently, we select the features representing high-frequency information during optimization, since regular adversarial noises concentrate on the high frequency of images. 
And we construct the feature interference loss and integrate it with the diffusion denoising loss as the objective function, to improve the ability of identity interference.

Our main contributions are as follows:
\begin{itemize}
    \item  We reveal the inadequate optimization steps that exists in current anti-customization methods, and gives detailed analysis regarding the perceptual discrepancy of diffusion models at different timesteps and intermediate layers, we have better aligned the optimization of adversarial gradients with diffusion models.
    \item Based on analysis, we propose a simple but effective anti-customization method, including adaptive greedy time selection and a feature interference loss to improve the protection ability. Our method can be easily generalized to different anti-customization frameworks and improve their performance.
    
    \item Extensive evaluations on two face datasets demonstrate that our method achieves more obvious disruption of the user's identity and provides better privacy protection.

\end{itemize}

\section{Related work}
\subsection{Generative Models and Diffusion Models}
Variational autoencoders (VAEs) \cite{kingma2013auto} and Generative adversarial networks (GANs) \cite{goodfellow2014generative} are popular frameworks among generative models which have strong generative ability. These models encode the data $x$ as latent variables $z$ and model the joint distribution $p_{\theta}(x, z)$. However, the quality of VAE samples is not competitive with GANs which are suffering from training instability \cite{gulrajani2017improved}. Since diffusion probabilistic models (DM) \cite{sohl2015deep} progressively add noise to the data from the joint distribution $q_{\theta}(x_{0:T}|x_{0})$ and denoise step by step, the efficiency of training and quality of samples have achieved state-of-work.

Unconditional generative models cannot produce desired samples and then models take different input as guidance have sprung up. Based on GAN, cGAN \cite{mirza2014conditional} generates images conditioned on the given labels $y$ and Cycle-GAN \cite{zhu2017unpaired} implements unpaired image translation considering the given image. Equipped with some techniques like classifier-free guidance \cite{ho2021classifier}, the diffusion model gains the ability to follow diverse prompts as conditions during generation.

The open-sourced Latent Diffusion Models (LDMs) \cite{rombach2022high} operate images in the latent space of low dimensions rather than pixel space which greatly reduces training computation. To support different condition inputs, they add cross-attention layers into the underlying U-Net backbone as conditioning mechanisms $\tau_{\theta}$. The above delicate designs make diffusion more friendly for users to create what they need. It can also assist them in designing their diffusions which synthesize visual contents that contain specific concepts (\eg, objects or styles) given by users during inference.

\subsection{Customization}
DreamBooth \cite{ruiz2023dreambooth} stands out as a popular diffusion-based method for customizing text-to-image generation. This approach involves presenting 3 $\sim$ 5 images depicting a specific concept (\eg, a particular dog) alongside a corresponding identifier (\eg, ``a sks dog''). DreamBooth utilizes these images to fine-tune the pre-trained Stable Diffusion model. This fine-tuning process encourages the model to ``memorize'' the concept and its identifier, enabling it to reproduce this concept in new contexts during inference.

On the other hand, Textual-inversion \cite{gal2022image} employs a different approach. This method freezes the U-Net and exclusively optimizes the text embedding of unique identifiers  (\eg. ``sks'') to represent the input concepts.

Inspired by DreamBooth, numerous works have been mushroomed such as custom-diffusion \cite{kumari2023multi}, Sine \cite{zhang2023sine}, among others. In the quest for more efficient fine-tuning, DreamBooth has a successful integration with LoRA \cite{hu2021lora} and has become a very influential customization project in the community. LoRA specifically decomposes the attention layer of the vision transformer model \cite{dosovitskiy2020image} into low-rank matrices reducing the cost associated with fine-tuning.

\subsection{Privacy Protection for Diffusion Models}
To alleviate the issue that private images are misused by Stable Diffusion based customization, so-called ``anti-customization'', some researchers \cite{shan2023glaze, van2023anti,DBLP:conf/icml/LiangWHZXSXMG23} have recently delved into privacy protection for diffusion models. AdvDM \cite{DBLP:conf/icml/LiangWHZXSXMG23} misleads the feature extraction of diffusion models. They analyze the training objective of fine-tuning and propose a direct way that uses the gradient of denoising loss as guidance to optimize the latent variables sampled from the denoising process. The generated adversarial examples degrade the generation ability of DreamBooth or other DM-based customized approaches. Inspired by \cite{huang2021initiative}, Anti-dreambooth \cite{van2023anti} uses the alternating surrogate and perturbation learning (ASPL) to approximate the real trained models. They train the model on clean data and use these models as the surrogate model to compute the noise added to the user-provided images. The perturbed images then as the training data for surrogate model fine-tuning which mimics the real scenes. Photoguard \cite{DBLP:conf/icml/SalmanKLIM23} specially focuses on unauthorized image inpainting which misleads the public and does harm to personal reputation. They use both VAE encoder attack and UNet attack targeted on a gray image to hinder infringers from creating fake news.

Some concerns limit the application of current anti-customization methods in more practical scenarios, such as ineffective optimization, poor identity disruption, and simply using the reconstruction loss as guidance to generate adversarial examples \cite{huang2022shape,huang2023improving,huang2022pointcat,huang2023opera,huang2023diversity}. 
Our paper aims to fill these blanks by figuring out the internal mechanisms that affects the performance of protection.

\section{Method}
\label{sec:formatting}
\subsection{Preliminaries}
\noindent\textbf{Diffusion Model (DM)} 
DM is a probabilistic generative model that samples from Gaussian distribution and then progressive denoise to learn the desired data distribution. Given $x_{0} \sim q(x)$, the forward process adds increasing noise to the input images at each time-step $t \in (0,T)$ which produces a sequence of noisy latent, $\{x_{0},x_{1},...,x_{T}\}$. The backward process trains model $\epsilon_{\theta}(x_{t},t,c)$ to predict the noise added in $x_{t}$ to infer $x_{t-1}$. During denoising, the training loss is $l_{2}$ distance, as shown in Eq.(\ref{noiseuncond}), Eq.(\ref{noise}). Although there are many implementations of text-to-image diffusion models, Stable Diffusion is one of the few open-sourced diffusion models and is widely used in the community. Thus, our method is mainly based on the checkpoints of stable diffusion provided in HuggingFace.
\begin{align} \label{noiseuncond}     
\mathcal{L}_{uncond}(\theta, x_{0})=\mathbb{E}_{x_0,t,\epsilon\in\mathcal{N}(0,1)}\|\epsilon-\epsilon_\theta(x_{t+1},t)\|_2^2, 
\\
\label{noise}
\mathcal{L}_{cond}(\theta, x_{0})=\mathbb{E}_{x_0,t,c,\epsilon\in\mathcal{N}(0,1)}\|\epsilon-\epsilon_\theta(x_{t+1},t,c)\|_2^2,
\end{align}
where $x$ is the input image, $t$ is the corresponding timestep, $c$ is condition input, (\eg, text or image), $\epsilon$ is the noise term.


\noindent\textbf{Adversarial Attack} 
The adversarial examples for classification are crafted to mislead the model to classify the given input to the wrong labels. However, the traditional attack strategies are not effective when dealing with generative models. For diffusion models, adversarial examples are some images that are added on imperceptible noise, causing diffusion models to consider them out of the generated distribution. In detail, this noise heightens the challenge of image reconstruction and hinders the customization capabilities of applications based on diffusion models (DMs). The optimized noise is often typically constrained to be smaller than a constant value $\eta$. The $\delta$ is determined through the following formation:

\begin{figure}
\centering
\includegraphics[width=1\linewidth]{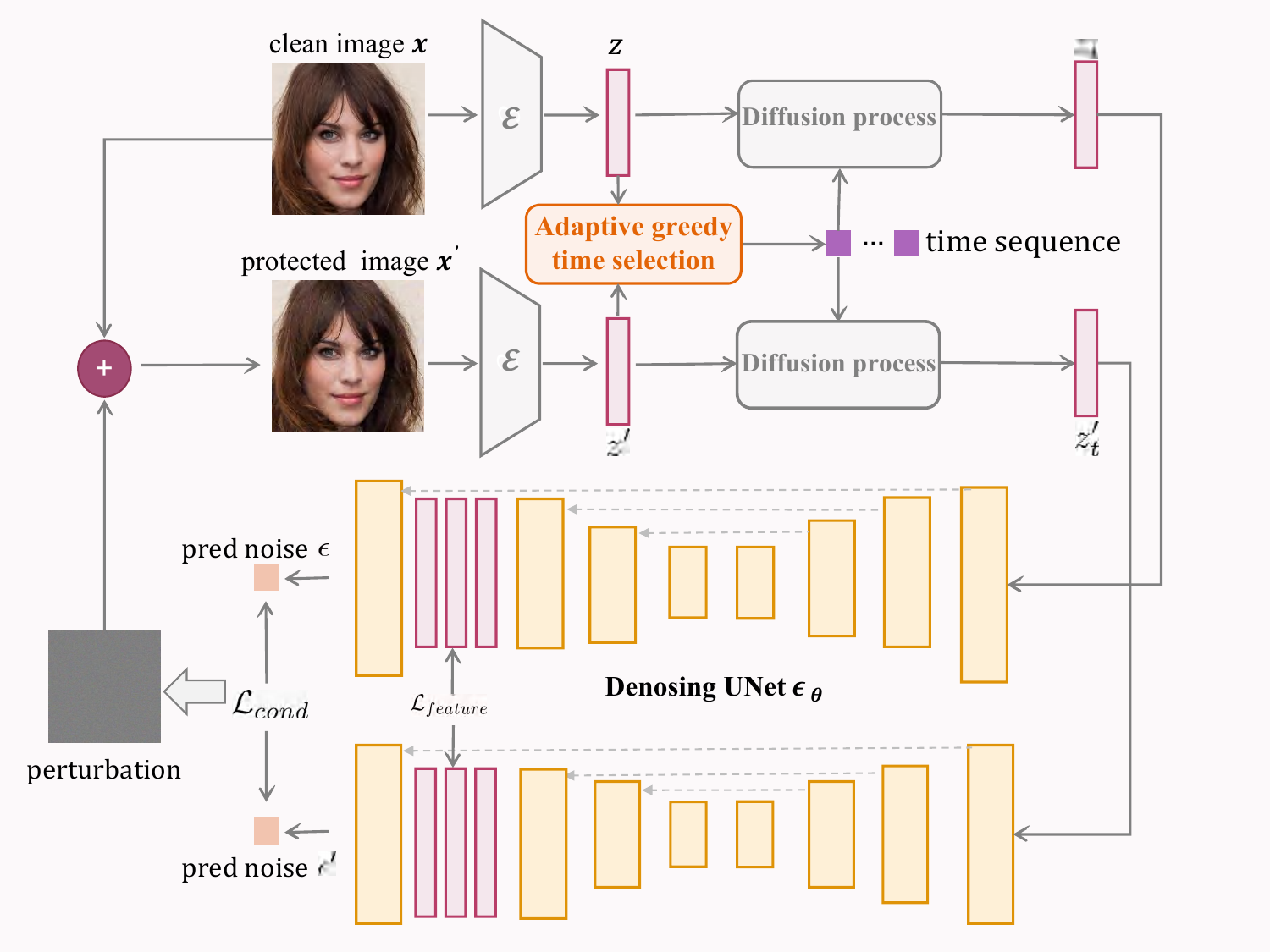}
\caption{Pipeline of SimAC. We first greedily select the time step with our adaptive strategy during the feed forward phase. Then we integrate the feature interference loss with the vanilla training loss as the final objective. The noise is iteratively optimized.}
\label{fig:pipe}
\end{figure}

\begin{equation}\label{attack}           
\delta_{adv} = \arg\mathop{\max}\limits_{\left\Vert \delta \right\Vert_{p} < \eta} L(f_{\theta}(x+\delta), y_{\mathop{real}}), 
\end{equation}
where $x$ is the input image, $y_{real}$ is the real images, and $L$ is the loss function used to evaluate the performance of adversarial examples.

Projected Gradient Descent (PGD) \cite{DBLP:conf/iclr/MadryMSTV18} is a widely used method to iteratively optimize adversarial examples. The process is formulated as
\begin{equation}\label{pgd}
x^{t+1} =\mathop{\prod}_{(x, \eta)}\left(x^t+\alpha \operatorname{sgn}\left(\nabla_x L(f_{\theta}(x+\delta), y_{\mathop{real}})\right)\right)
\end{equation}
where $x^{0} = x$ and $x$ is the input image, $\mathrm{sgn}(\cdot)$ is sign function, $(\nabla_x L(f_{\theta}(x+\delta), y_{\mathop{real}})$ is the gradient of the loss function with respect to $x+\delta$. $\alpha$ represents the step size during each iteration and $t$ is the iteration number. With the operation $\Pi_{(x, \eta)}$, the noise is limited to a $\eta$-ball ensuring the adversarial examples are acceptable.

\subsection{Overview}

Here we delve into the properties of LDMs and analyze the potential vulnerabilities that can benefit the attack. 
In Sec.~\ref{LDM}, we give a comprehensive analysis of the properties. 
In Sec.~\ref{3.4} and \ref{3.5}, we propose our method based on these analyses, mainly including two components, \ie, adaptive greedy time interval selection and the feature interference loss. 
The overall pipeline is illustrated in Figure~\ref{fig:pipe}.

\subsection{Analysis on Properties of LDMs}
\label{LDM}

\noindent\textbf{Differences at different time step.}
Our exploration focuses on analyzing gradients across various time intervals from a statistical standpoint. We've established a threshold of $1e^{-10}$ to assess the number of gradients with absolute values below this threshold in perturbed images across different timesteps. Figure ~\ref{fig:grad} illustrates a notable trend that during the latter part of the total time interval $(0, MaxTrainStep)$, there's a sharp increase in the count of absolute gradients below the specified threshold. Additionally, Figure ~\ref{fig:grad} presents the maximum value, mean, and median of absolute gradient decreases in the perturbed images when timesteps get larger.

\begin{figure}[t]
\begin{minipage}{0.312\linewidth}
    \centering
    \includegraphics[width=1\linewidth]{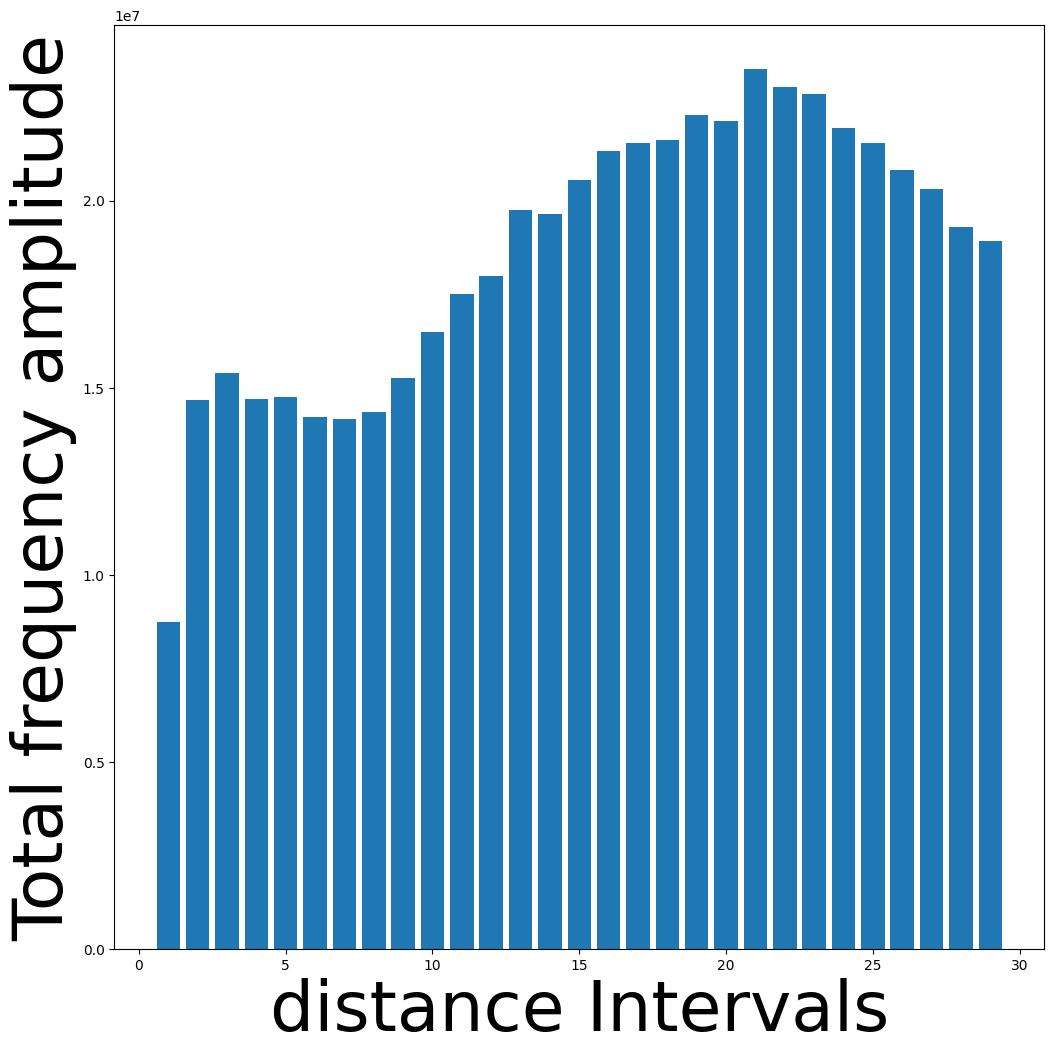}
    \centerline{\scriptsize \textbf {t $\in$ [0, 100]}}
\end{minipage}
\hfill
\begin{minipage}{0.312\linewidth}
    \centering
    \includegraphics[width=1\linewidth]{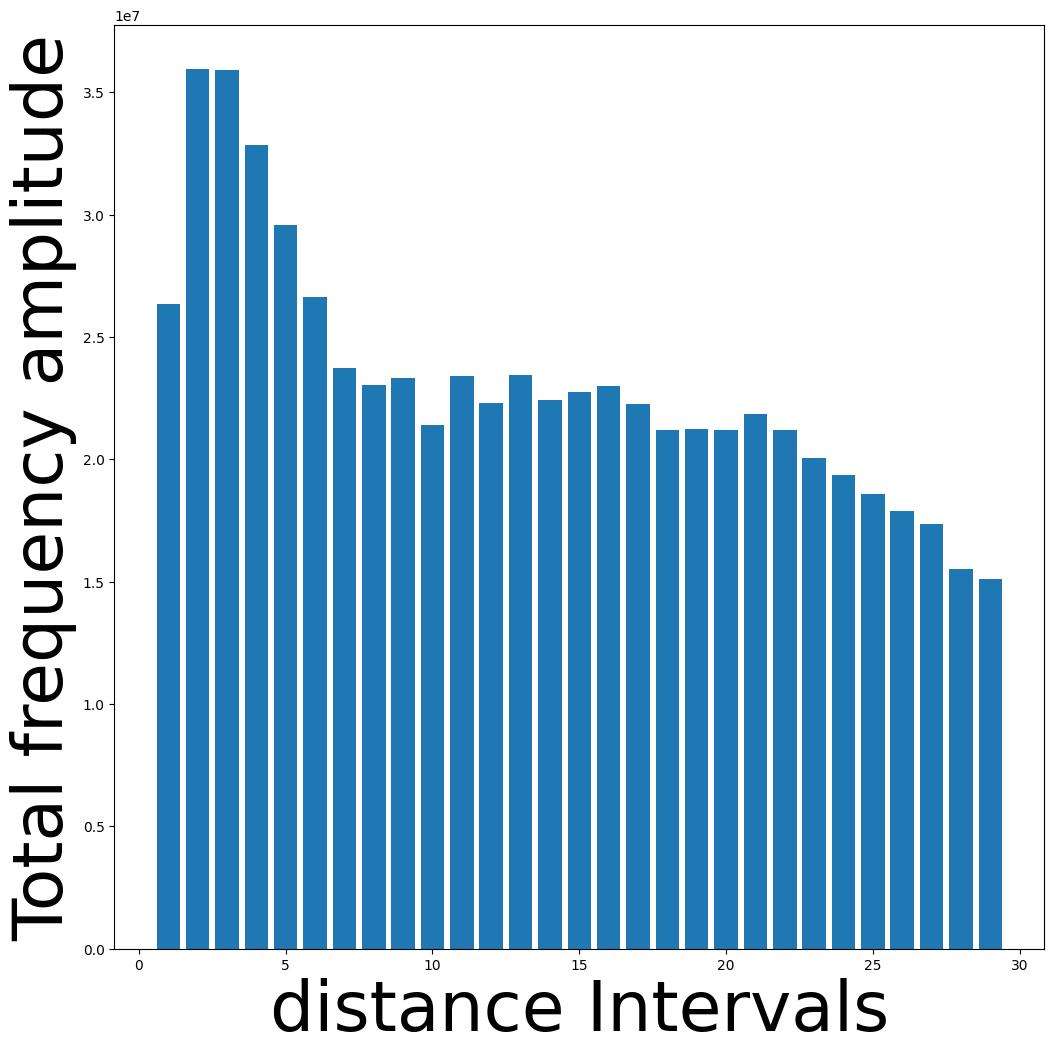}
    \centerline{\scriptsize \textbf{t $\in$ [400, 500]}}
\end{minipage}
\hfill
\begin{minipage}{0.312\linewidth}
    \centering
    \includegraphics[width=1\linewidth]{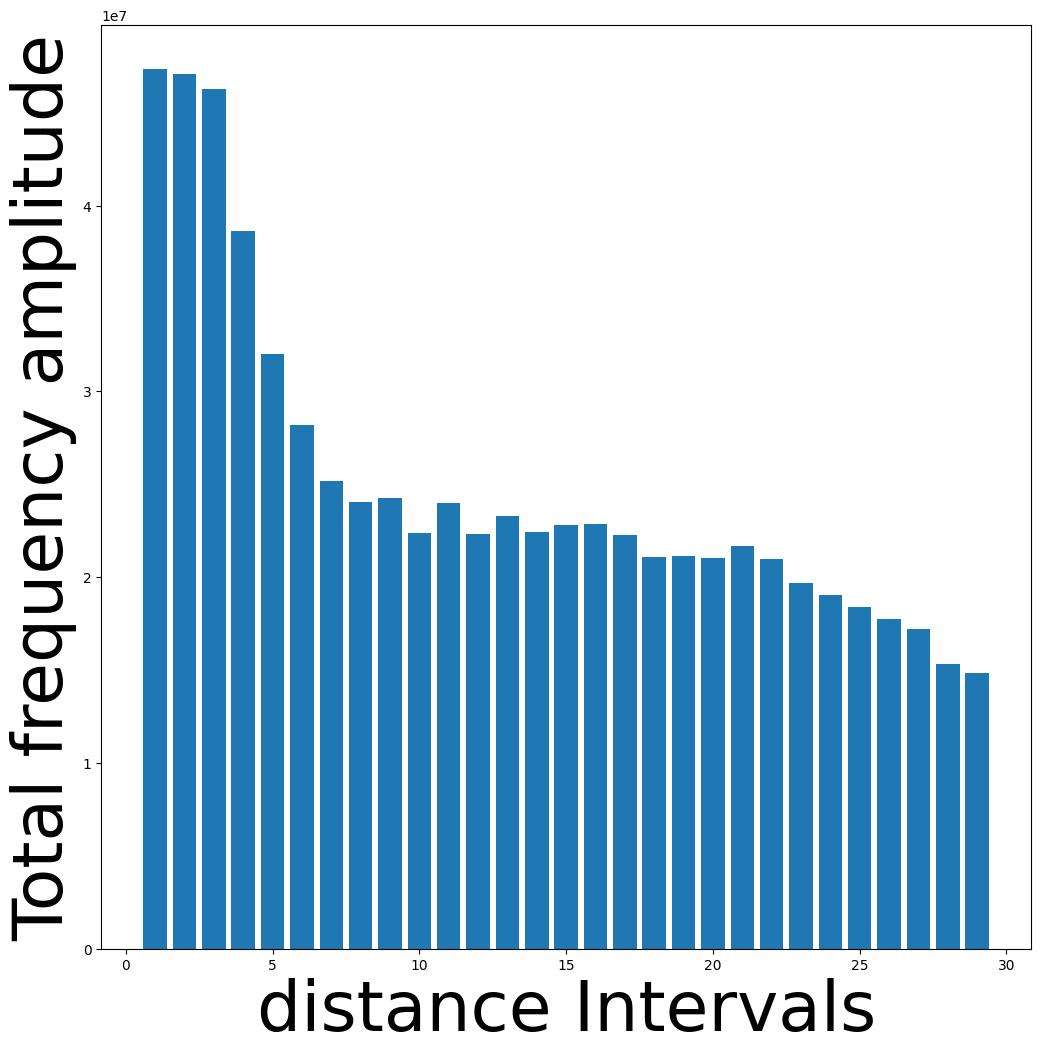}
    \centerline{\scriptsize  \textbf{t $\in$ [700, 800]}}
\end{minipage}
\caption{Frequency domain residual analysis. The X-axis is the distance from the low-frequency center (from low frequency to high frequency), and the Y-axis represents total magnitude of frequency residual. It can be seen that when $t$ is small, the high-frequency difference exceeds the low-frequency difference. As $t$ increasing, the low-frequency difference gradually dominated.}
\label{fig:freq}
\end{figure}

As shown Eq.(\ref{forward}), the forward process admits sampling $x_{t}$ at an
arbitrary timestep $t$. The magnitude of the noise corresponding to timestep is determined by a pre-defined noise schedule. To ensure that the final latent $x_{T}$ conforms to a standard normal distribution, the amplitude of noise injection gradually increases with the timestep.
 \begin{equation}\label{forward}
q\left(\mathbf{x}_t \mid \mathbf{x}_0\right)=\mathcal{N}\left(\mathbf{x}_t ; \sqrt{\bar{\alpha}_t} \mathbf{x}_0,\left(1-\bar{\alpha}_t\right) \mathbf{I}\right)
\end{equation}
 During denoising, we reconstruct the input image $x_{0}$ based on the noisy latent $z_{t}$ and visualize the difference between the reconstructed image and the input image in the frequency domain at different timesteps. As shown in Figure~\ref{fig:freq}, when $t \in [0,100]$, the main difference between the original and reconstructed images lies in the high-frequency components. Our initial expectation is to disrupt the reconstruction of the original image by introducing high-frequency adversarial noise, aiming to achieve anti-customization. However, when $t$ increases, such as $t \in  [700,800]$, the difference in low-frequency components between the reconstructed images and the input images dominates most of the frequency domain differences between the two. Therefore, this can explain why, when the timestep of noise scheduler is large, the gradient of the perturbed image is close to zero and result in ineffective noise optimization.

\begin{figure*}[t]
\begin{minipage}{0.072\linewidth}
    \centering
    \includegraphics[width=1\linewidth]{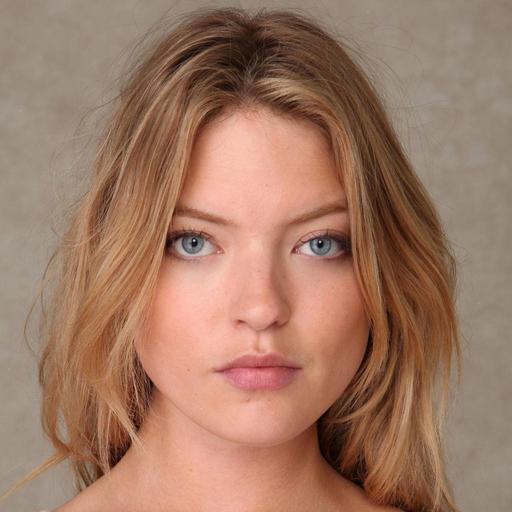}
    \centerline{\scriptsize \textbf {portrait image}}
\end{minipage}
\begin{minipage}{0.072\linewidth}
    \centering
    \includegraphics[width=1\linewidth]{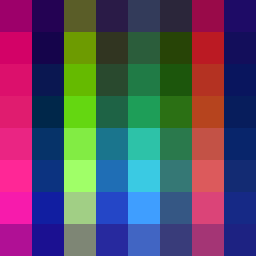}
    \centerline{\scriptsize \textbf {layer 0}}
\end{minipage}
\hfill
\begin{minipage}{0.072\linewidth}
    \centering
    \includegraphics[width=1\linewidth]{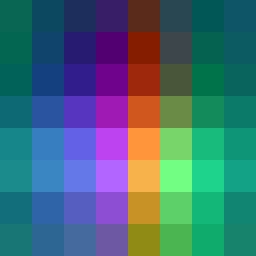}
    \centerline{\scriptsize \textbf{layer 1}}
\end{minipage}
\hfill
\begin{minipage}{0.072\linewidth}
    \centering
    \includegraphics[width=1\linewidth]{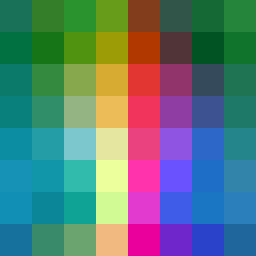}
    \centerline{\scriptsize \textbf{layer 2}}
\end{minipage}
\hfill
\begin{minipage}{0.072\linewidth}
    \centering
    \includegraphics[width=1\linewidth]{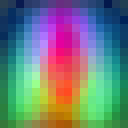}
    \centerline{\scriptsize \textbf{layer 3}}
\end{minipage}
\hfill
\begin{minipage}{0.072\linewidth}
    \centering
    \includegraphics[width=1\linewidth]{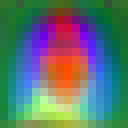}
    \centerline{\scriptsize \textbf{layer 4}}
\end{minipage}
\hfill
\begin{minipage}{0.072\linewidth}
    \centering
    \includegraphics[width=1\linewidth]{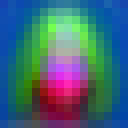}
    \centerline{\scriptsize  \textbf{layer 5}}
\end{minipage}
\hfill
\begin{minipage}{0.072\linewidth}
    \centering
    \includegraphics[width=1\linewidth]{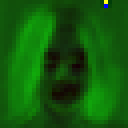}
    \centerline{\scriptsize  \textbf{layer 6}}
\end{minipage}
\hfill
\begin{minipage}{0.072\linewidth}
    \centering
    \includegraphics[width=1\linewidth]{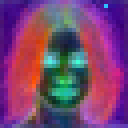}
    \centerline{\scriptsize  \textbf{layer 7}}
\end{minipage}
\hfill
\begin{minipage}{0.072\linewidth}
    \centering
    \includegraphics[width=1\linewidth]{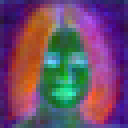}
    \centerline{\scriptsize  \textbf{layer 8}}
\end{minipage}
\hfill
\begin{minipage}{0.072\linewidth}
    \centering
    \includegraphics[width=1\linewidth]{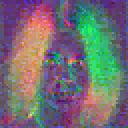}
    \centerline{\scriptsize  \textbf{layer 9}}
\end{minipage}
\hfill
\begin{minipage}{0.072\linewidth}
    \centering
    \includegraphics[width=1\linewidth]{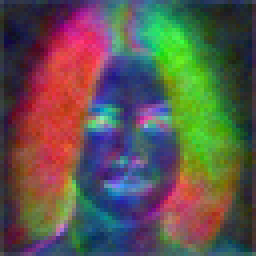}
    \centerline{\scriptsize  \textbf{layer 10}}
\end{minipage}
\hfill
\begin{minipage}{0.072\linewidth}
    \centering
    \includegraphics[width=1\linewidth]{fig/layer_pca/6_412_time_500.png}
    \centerline{\scriptsize  \textbf{layer 11}}
\end{minipage}
\vspace{-0.5em}
\caption{PCA visualization of features. The output feature of residual blocks are visualized at timestep=500. As feature goes deeper, the high-frequency information captured by the feature becomes more significant.}
\label{fig:pca}
\end{figure*}

\begin{figure}[t]
\begin{minipage}{0.18\linewidth}
    \centering
    \includegraphics[width=1\linewidth]{fig/clean/412.jpg}
    \centerline{\scriptsize  \textbf{portrait image}}
\end{minipage}
\hfill
\begin{minipage}{0.18\linewidth}
    \centering
    \includegraphics[width=1\linewidth]{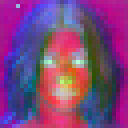}
    \centerline{\scriptsize \textbf {t $\in$ [100,200]}}
\end{minipage}
\hfill
\begin{minipage}{0.18\linewidth}
    \centering
    \includegraphics[width=1\linewidth]{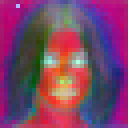}
    \centerline{\scriptsize \textbf{t $\in$ [300, 400]}}
\end{minipage}
\hfill
\begin{minipage}{0.18\linewidth}
    \centering
    \includegraphics[width=1\linewidth]{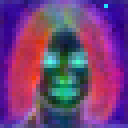}
    \centerline{\scriptsize  \textbf{t $\in$ [500, 600]}}
\end{minipage}
\hfill
\begin{minipage}{0.18\linewidth}
    \centering
    \includegraphics[width=1\linewidth]{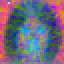}
    \centerline{\scriptsize  \textbf{t $\in$ [700, 800]}}
\end{minipage}
\vspace{-0.5em}
\caption{Diffusion feature over different timesteps. As the timestep of the noise scheduler increasing, part of the high-frequency information is lost and it is not good for adding high-frequency adversarial noise. This means that choosing an appropriate time sequence is also beneficial to feature interference loss.}
\label{fig:pca-7}
\end{figure}

\noindent\textbf{Differences at Different UNet Layers.}
Inspired by \cite{tumanyan2023plug}, we utilize PCA to visualize the output features of each layer in the U-Net decoder blocks during denoising. The decoder block consists of self-attention, cross-attention, and residual blocks, we select features of residual blocks. There are 11 layers in total, and the layers are visualized at timestep=500. As Figure~\ref{fig:pca} shows, with the output feature of UNet decoder blocks increasing, the visualized features gradually change from depicting structures and other low-frequency information to capturing texture and similar high-frequency information. Since our noise is intended to disrupt high-frequency information, the deeper features are good perturbation objects for our adversarial noises to reinforce interference which attempts to perturb high-frequency components in generated images.

\subsection{Adaptive Greedy Time Interval Selection}\label{3.4}
To achieve more effective privacy protection within a limited number of noise injection steps, we propose a fast adaptive time interval selection strategy. Firstly, based on the input image, we randomly select five timesteps from the interval $(0, MaxTrainingStep)$. For each time step, we compute the gradient concerning the input image. The absolute values of the gradients are summed. If the sum of absolute gradient values is the minimum at timestep $t$, the corresponding interval $(t-20,t+20)$ is then removed. For the timestep with maximal sum, noise is computed and added to the image and this noised image becomes the input for the next round of gradient computation. This process is repeated until the final time interval length is no greater than 100 or the training steps reach default maximum. After selection, We obtain the ultimate interval used for noise injection.

Our method relies on PGD\cite{DBLP:conf/iclr/MadryMSTV18} (Projected Gradient Descent), an iterative approach for adversarial attacks. At each iteration, computing the gradient becomes pivotal as it signifies the rate of change of the objective function concerning the perturbed images. The gradient approaching zero implies the possibility of iterative optimization but with minimal alterations to the objective function. This ultimately leads to reduced attack efficiency and diminished effectiveness. Thus, randomly selecting timesteps makes it more challenging to optimize the objective function under the same noise budget and results in decreased efficiency and effectiveness in countering customization. This highlights the crucial need for employing our adaptive, greedy time interval selection for perturbation.

\begin{algorithm}[h]
	\SetAlgoLined
	\KwIn{clean image $x$; model parameter $\theta$; number of search steps $ N $; step length $ \alpha$.}
	\KwOut{time interval $seq$.}
        Initialize $x^{'} \leftarrow x, seq=(0, MaxTrainStep)$\;
	\For{$i=1$ to $N$}{
        \eIf{length of $seq > 100$}{
                Sample timestep set $ts=(t_{1}, t_{2}, t_{3}, t_{4}, t_{5}) $ from $seq$,
                
                \For{$t$ in $ts$}{
                    $sum(\nabla_{x_{t}^{i}}\mathcal{L}_{DM}(\theta))$
                    
                    \If{$sum$ at $t$ is Minimum}{
                       delete $ (t-20, t+20) $ from $seq$
                    }
                    \If {$sum$ at $t$ is Maximum}{
                        $x^{'}=x+\alpha sgn((\nabla_{x_{t}^{i}}\mathcal{L}_{DM}(\theta))$
                    }
                }
            }
            {
            break
            }
        }
        \KwResult{$seq$}
	\caption{Adaptive Greedy Time Interval Selection}
	\label{alg:ada_select}
\end{algorithm}

\noindent\textbf{Analysis on efficacy of adaptive greedy time interval selection.}
Here we simply analyze our adaptive greedy selection for timestep $t$ in algorithm \ref{alg:ada_select}. 
Its key steps in each iteration are randomly sampling 5 timesteps $T_s=\{t_j\}_{j=1}^5$ and rescaling the timestep range by deleting the interval of $t\in T_s$, which leads to the smallest gradient absolute value. 
Thus avoiding very small or even zero gradients during optimization. 

\noindent\textbf{Theorem 1.} 
\textit{Suppose the timestep range is rescaled from $T=A\cup B$ to $T'=(A\setminus\Delta A) \cup (B\setminus\Delta B)$ in some iteration, where timestep $t\in B$ leads to zero gradients, satisfying $A\cap B=\varnothing$, and $\Delta A\cup\Delta B$ denotes the deleted interval, then we have $E_{t\sim T}|\nabla_x \mathcal{L}_{DM}| < E_{T_s\sim T}E_{t\sim T'}|\nabla_x \mathcal{L}_{DM}|$.}

\begin{proof}
We can easily obtain $E_{T_s\sim T}\frac{|\Delta A|}{|A|}<E_{T_s\sim T}\frac{|\Delta B|}{|B|}$ via proof by contradiction. For simplicity, we denote $\nabla_x \mathcal{L}_{DM}$ as $g(t)$, $LHS$ and $RHS$ denote left and right, then
\begin{align}
        LHS &= p(t\in A)E_{t\sim A}|g(t)| + p(t\in B)E_{t\sim B}|g(t)| \nonumber\\
        &= \frac{|A|}{|A|+|B|} E_{t\sim A}|g(t)|
        \approx E_{T_s\sim T}\frac{|A|\cdot E_{t\sim A\setminus\Delta A}|g(t)|}{|A|+|B|}  \nonumber\\
        &< E_{T_s\sim T}\frac{|A|-|\Delta A|}{|A|+|B|-|\Delta A| - |\Delta B|} E_{t\sim A\setminus\Delta A}|g(t)| \nonumber\\
        &= E_{T_s\sim T} E_{t\sim A\setminus\Delta A} p(t\in A\setminus\Delta A)|g(t)| = RHS \nonumber
\end{align}
\end{proof}

\subsection{Feature Interference Loss}
\label{3.5}
Based on the analysis in Sec~\ref{LDM}, we propose the feature interference loss and integrate it with the vanilla training loss as the overall objective for optimization. 
This loss calculates the layer-wise euclidian distance between the intermediate features in some specially selected layers, \ie,   
\begin{align}\label{Feature loss}           
\mathcal{L}_{f}&=\mathbb{E}\|{f_l^t}^*-f_l^t\|_2^2 
\\
\label{all loss}           
\mathcal{L} &= \mathcal{L}_{cond}  + \lambda \mathcal{L}_{f}
\end{align}
Where ${f_l^t}^*$ is the output features of the selected layers sets $l$ at timestep $t$ for the input image, $f_l^t$ is the output features for current perturbed images, and $\lambda$ represents the weighting coefficient for the feature interference loss. 

We visualized the output features of layer 7 at different time steps. As Figure~\ref{fig:pca-7} shows, when the timesteps increase, more noise is added to the latent and the response to the high-frequency components of the input image gradually decreases. Therefore, utilizing feature interference loss as the optimization objective for adversarial noise in small time intervals is a better choice to keep the effectiveness of adversarial noise disturbing high-frequency information.

\section{Experiments}
\subsection{Setup}

\noindent\textbf{Dataset} 
We utilized two facial datasets for experiments including Celeb-HQ \cite{DBLP:conf/iclr/celeb} and VGGFace2 dataset\cite{cao2018vggface2}. The dataset comprises about 50 individuals and aligns with the Anti-DreamBooth, with each individual including at least 15 clear face images for customization. Since Anti-dreambooth needs alternating training, the dataset is divided into three sets, including set A, set B, and set C and each set contains 5 portrait images.

\noindent\textbf{Model} 
Since the open-sourced stable diffusion is the most popular implementation of latent diffusion among the community, our experiments mainly use the SD-v2.1 \cite{sdv2.1}. To test the performance of our method in a black-box scenario, we assume the versions of Stable Diffusion between anti-customization and customization are the same or different.

\noindent\textbf{Baseline} We compare several open-source methods that employ adversarial attacks on diffusion models to protect user images from being misused by text-to-image diffusion models, including Photoguard \cite{DBLP:conf/icml/SalmanKLIM23}, AdvDM \cite{DBLP:conf/icml/LiangWHZXSXMG23} and Anti-DreamBooth \cite{van2023anti}. Due to the high GPU memory requirements of the complete PhotoGuard, we only utiliz the VAE encoder attack strategy in its paper for comparison.

\noindent\textbf{Metric}
We utilize a face detector named Retinaface \cite{deng2020retinaface} to detect if there is a face in the image and use the Face Detection Failure Rate (FDFR) to assess the level of disruption to the generated faces. Upon detecting a face, we encode it using ArcFace \cite{deng2019arcface} and calculate the cosine similarity between the protected image and clean input, measuring the identity resemblance between the detected face in the generated image and that of users. This matrix is defined as Identity Score Matching (ISM). In addition, the image quality is quantified by BRISQE \cite{Mittal_Moorthy_Bovik_2012}, and the quality of detected facial images is measured through SER-FIQ \cite{DBLP:conf/cvpr/TerhorstKDKK20}.

\begin{table}[t]
\centering
\scriptsize
\begin{tabular}{l|cccc}
\toprule
\multicolumn{5}{c}{CelebA-HQ} \\
\midrule
\multirow{2}{*}{Method} & \multicolumn{4}{c}{``a photo of sks person"} \\
\cmidrule{2-5}
& ISM$\downarrow$& FDFR$\uparrow$ & BRISQUE $\uparrow$& SER-FQA$\downarrow$ \\
\midrule
PhotoGurad \cite{DBLP:conf/icml/SalmanKLIM23} &\textbf{0.25} 	&41.09 	&19.38 	&0.55 	 \\
AdvDM \cite{DBLP:conf/icml/LiangWHZXSXMG23} &0.32 	&70.48 	&38.17 	&0.20 	 \\
Anti-DB \cite{van2023anti} &0.28 	&77.28 	&37.43 &	\textbf{0.20} 	 \\
Anti-DB + SimAC & 0.31	& \textbf{87.07} 	& \textbf{38.86} 	&0.21 	 \\
\bottomrule

\multirow{2}{*}{Method} & \multicolumn{4}{c}{``a dslr portrait of sks person"}\\
\cmidrule{2-5}
& ISM$\downarrow$& FDFR$\uparrow$ & BRISQUE $\uparrow$& SER-FQA$\downarrow$ \\
\midrule
PhotoGurad \cite{DBLP:conf/icml/SalmanKLIM23} &0.20 	&28.50 	&29.33 	& 0.59 \\
AdvDM \cite{DBLP:conf/icml/LiangWHZXSXMG23} &0.25 	&65.37 	&37.86 	& 0.41 \\
Anti-DB \cite{van2023anti} 	&0.19 &	86.80 	&38.90 	& 0.27 \\
Anti-DB + SimAC & \textbf{0.12} 	&	\textbf{96.87} 	& \textbf{42.10} &	\textbf{0.15} \\
\bottomrule

\multirow{2}{*}{Method} & \multicolumn{4}{c}{``a photo of sks person looking at the mirror"} \\
\cmidrule{2-5}
& ISM$\downarrow$& FDFR$\uparrow$ & BRISQUE $\uparrow$& SER-FQA$\downarrow$ \\
\midrule
PhotoGurad\cite{DBLP:conf/icml/SalmanKLIM23} &0.18 &	30.07 &	26.96 &	0.40  \\
AdvDM\cite{DBLP:conf/icml/LiangWHZXSXMG23}  &0.29 	 &35.10  &	36.46  &	0.36  
 \\
Anti-DB\cite{van2023anti} &0.22 	&42.86 &	40.34 &	0.28 
\\
Anti-DB + SimAC & \textbf{0.12} &	\textbf{91.90} 	& \textbf{43.97} 	& \textbf{0.06} 
\\
\bottomrule

\multirow{2}{*}{Method} & \multicolumn{4}{c}{``a photo of sks person in front of eiffel tower"}\\
\cmidrule{2-5}
& ISM $\downarrow$& FDFR $\uparrow$ & BRISQUE$\uparrow$& SER-FQA$\downarrow$\\
\midrule
PhotoGurad\cite{DBLP:conf/icml/SalmanKLIM23} &	0.08 	&50.95 	&32.82 	&0.40 \\
AdvDM\cite{DBLP:conf/icml/LiangWHZXSXMG23} &	0.09 	 &38.10  &	36.02 	 &0.30 
 \\
Anti-DB\cite{van2023anti} &	0.06 &	56.26 	&41.35 	&0.22 
\\
Anti-DB + SimAC &	\textbf{0.05} &	\textbf{66.19} & 	\textbf{42.77} &	\textbf{0.12} 
\\
\bottomrule
\end{tabular}
\vspace{-1em}
\caption{Comparison with other open-sourced anti-customization methods on CelebA-HQ. We evaluate the performance under four different prompts during customization.}
\label{tab:compare}
\vspace{-1em}
\end{table}

\noindent\textbf{Implementation Details}
We set the same noise budget for all methods $\eta=16/255$. Additionally, the optimization steps and step size align with the settings specified in each baseline. 
The number of training epochs is 50 and each epoch includes 3 steps for surrogate model training and 9 steps for adversarial noise optimization. 
The step size for adding noise is set as 0.005 and the learning rate for the training model is set as 5e-7 which is appropriate for human faces. 
The maximum adaptive greedy search steps are set as 50 by default. 
After training, we used the noised image for customization. 
The default base model is stable diffusion v2.1 combined with DreamBooth. 
According to the above analysis of the middle layer of the model, we select the 9, 10, and 11 layer features that can best represent the high-frequency information of the image for additional disturbance. The weight $\lambda$ of feature interference loss is set to 1. After finetuning 1000 steps, we save the model checkpoints and conduct inference. For each prompt, 30 images in .png format are generated for metric calculation.

\begin{table}[h]
\centering
\scriptsize
\begin{tabular}{l|cccc}
\toprule
\multicolumn{5}{c}{VGG-Face2} \\
\midrule
\multirow{2}{*}{Method} & \multicolumn{4}{c}{``a photo of sks person"} \\
\cmidrule{2-5}
& ISM$\downarrow$& FDFR$\uparrow$ & BRISQUE $\uparrow$& SER-FIQ$\downarrow$ \\
\midrule
PhotoGurad \cite{DBLP:conf/icml/SalmanKLIM23} &\textbf{0.29} 	&29.27 	&20.67 	&0.47 \\
AdvDM \cite{DBLP:conf/icml/LiangWHZXSXMG23}  
&0.32 &63.07 	&38.51 	&\textbf{0.21} 	\\
Anti-DB\cite{van2023anti} &0.30 	&64.67 	&37.89 	&0.22  \\
Anti-DB + SimAC &0.31 	&\textbf{80.27} 	&\textbf{40.71} 	&0.22 	\\
\bottomrule

\multirow{2}{*}{Method} & \multicolumn{4}{c}{``a dslr portrait of sks person"}\\
\cmidrule{2-5}
& ISM$\downarrow$& FDFR$\uparrow$ & BRISQUE $\uparrow$& SER-FIQ$\downarrow$ \\
\midrule
PhotoGurad \cite{DBLP:conf/icml/SalmanKLIM23} &0.25	&17.33 	&28.52 	&0.55 	
 \\
AdvDM \cite{DBLP:conf/icml/LiangWHZXSXMG23}  
&0.30 	&67.80 	&37.58 	&0.35 
  \\
Anti-DB\cite{van2023anti} &0.23 	&77.47 	&38.79 	&0.29 	
 \\
Anti-DB + SimAC &\textbf{0.11} 	&\textbf{96.33} 	&\textbf{41.78} 	&\textbf{0.12}\\
\bottomrule

\multirow{2}{*}{Method} & \multicolumn{4}{c}{``a photo of sks person looking at the mirror"} \\
\cmidrule{2-5}
& ISM$\downarrow$& FDFR$\uparrow$ & BRISQUE $\uparrow$& SER-FIQ$\downarrow$ \\
\midrule
PhotoGurad\cite{DBLP:conf/icml/SalmanKLIM23} &0.17 	&40.87 	&30.10 	&0.32  \\
AdvDM\cite{DBLP:conf/icml/LiangWHZXSXMG23}  &0.27 	&37.73 	&35.43 	&0.29 \\
Anti-DB\cite{van2023anti} &0.25 	&45.00 &	39.25 &	0.27 \\
Anti-DB + SimAC &\textbf{0.13} 	&\textbf{89.60} 	&\textbf{44.97} 	&\textbf{0.07}  \\
\bottomrule

\multirow{2}{*}{Method} & \multicolumn{4}{c}{``a photo of sks person in front of eiffel tower"}\\
\cmidrule{2-5}
& ISM$\downarrow$& FDFR$\uparrow$ & BRISQUE $\uparrow$& SER-FIQ$\downarrow$ \\
\midrule
PhotoGurad\cite{DBLP:conf/icml/SalmanKLIM23} &0.13 	&51.07 	&30.69 	&0.41\\
AdvDM\cite{DBLP:conf/icml/LiangWHZXSXMG23}  &0.14 &38.67 	&35.99 &0.31 
 \\
Anti-DB\cite{van2023anti} &	0.10 &54.93 &41.13 	&0.23 \\
Anti-DB + SimAC &\textbf{0.09} 	&\textbf{61.20} 	&\textbf{42.17} 	&\textbf{0.10}\\
\bottomrule
\end{tabular}
\vspace{-1em}
\caption{Comparison with other open-sourced anti-customization methods on VGG-Face2. We evaluate the performance under four different prompts during customization.}
\label{tab: compare vgg}
\vspace{-1em}
\end{table}

\subsection{Comparison with Baseline Methods}
\noindent\textbf{Quantitative Results}
To test the effectiveness of our approach in enhancing the protection of users' portrait images, we conduct a quantitative comparison under three prompts in Table~\ref{tab:compare} and Table~\ref{tab: compare vgg}. To be more practical, we list four text prompts for inference, the first one ``a photo of sks person" is the same as the prompt used in training, while the other three were prompts unseen before. For each prompt, we randomly select 30 generated images to compute each metric, reporting their average values. 

We can find that our method greatly improves the performance of Anti-DreamBooth and outperforms other baselines across all prompts. Due to our analysis of the unique properties of the diffusion model during denoising and effective addition of high-frequency noise, the face detection failure rates greatly increase, and the identity matching scores between the detected face and the input image are the lowest among all methods. This implies that our method is more effective in resisting abuse from customization.

\begin{figure}[h]
\begin{minipage}{0.18\linewidth}
    \centering
    \includegraphics[width=1\linewidth]{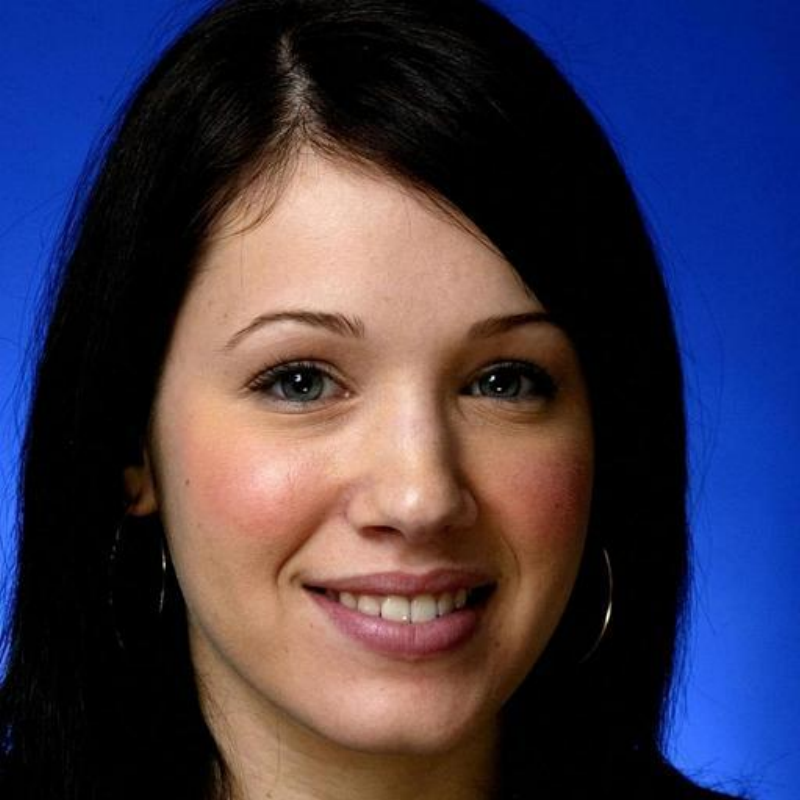}
\end{minipage}
\hfill
\begin{minipage}{0.18\linewidth}
    \centering
    \includegraphics[width=1\linewidth]{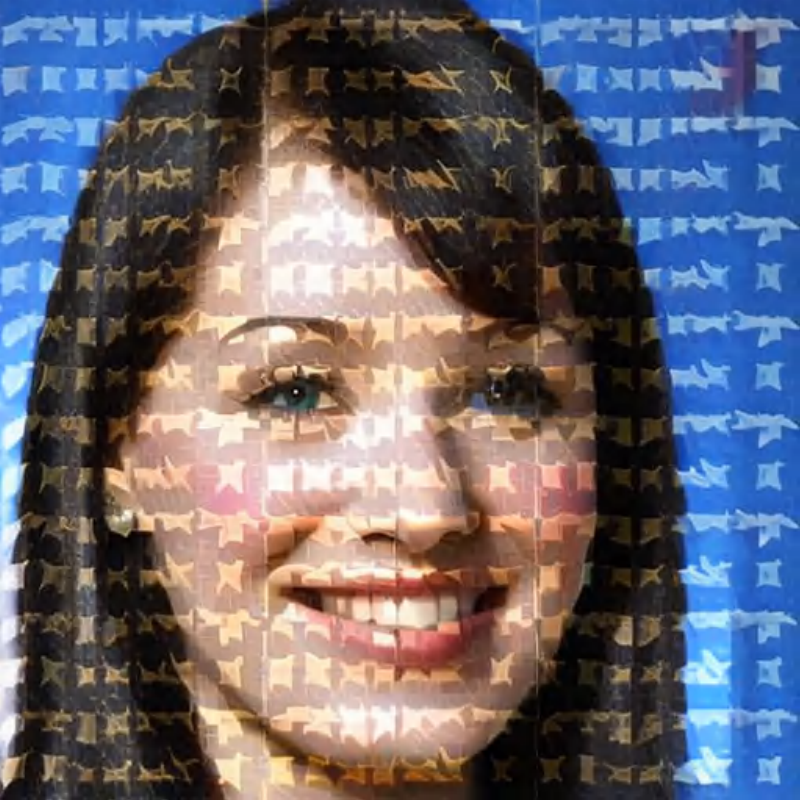}
\end{minipage}
\hfill
\begin{minipage}{0.18\linewidth}
    \centering
    \includegraphics[width=1\linewidth]{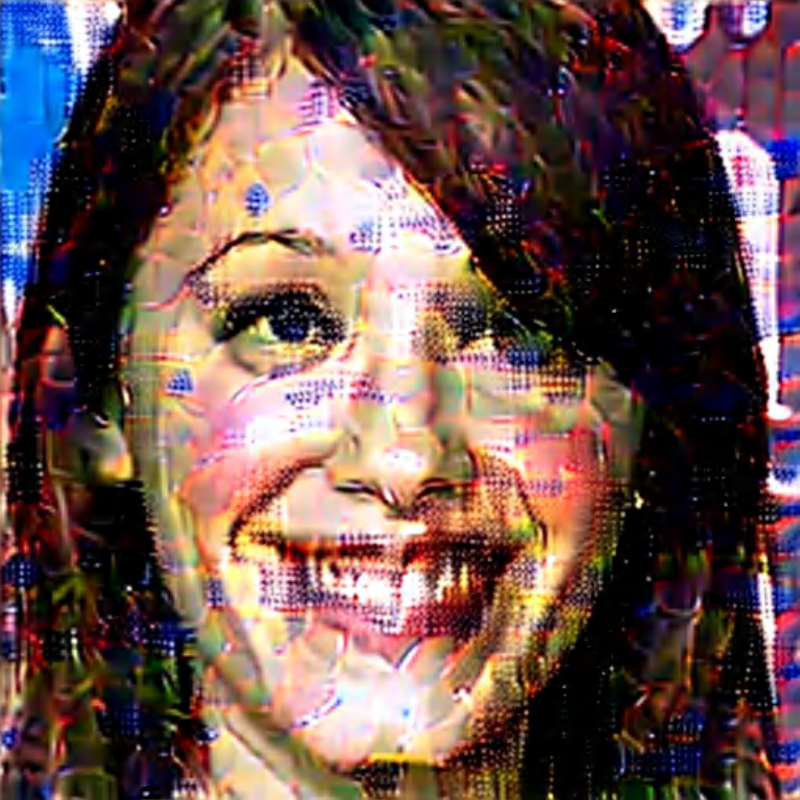}
\end{minipage}
\hfill
\begin{minipage}{0.18\linewidth}
    \centering
    \includegraphics[width=1\linewidth]{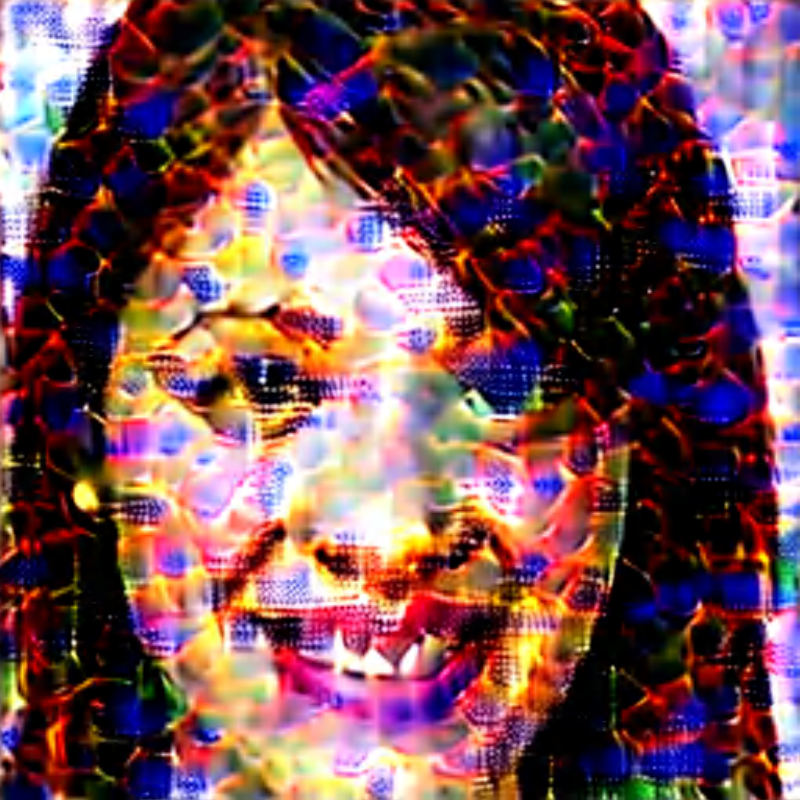}
\end{minipage}
\hfill
\begin{minipage}{0.18\linewidth}
    \centering
    \includegraphics[width=1\linewidth]{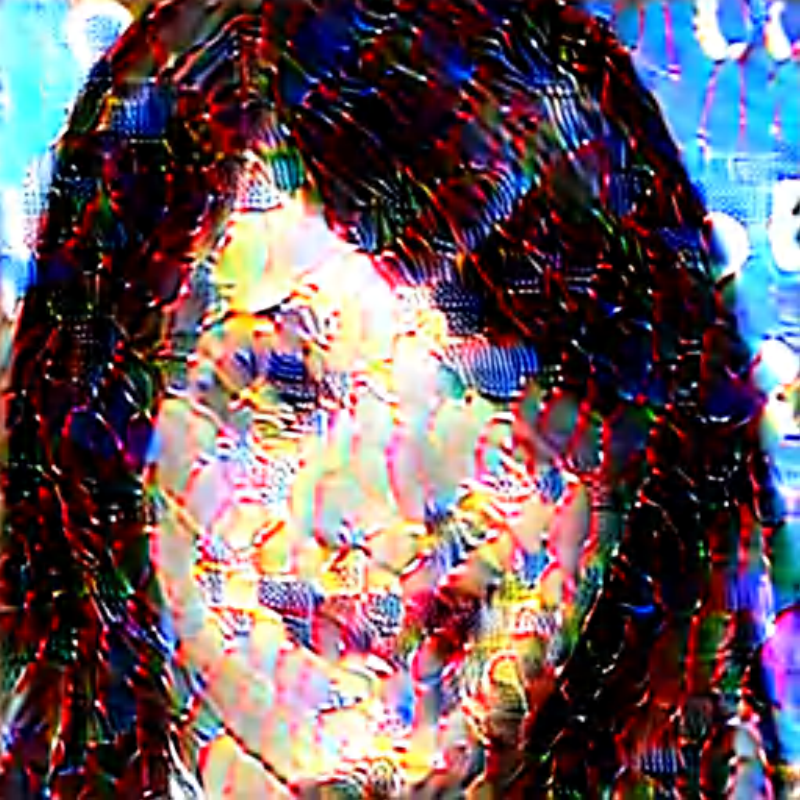}
\end{minipage}
\begin{minipage}{0.18\linewidth}
    \centering
    \includegraphics[width=1\linewidth]{fig/compare/1590.pdf}
\end{minipage}
\hfill
\begin{minipage}{0.18\linewidth}
    \centering
    \includegraphics[width=1\linewidth]{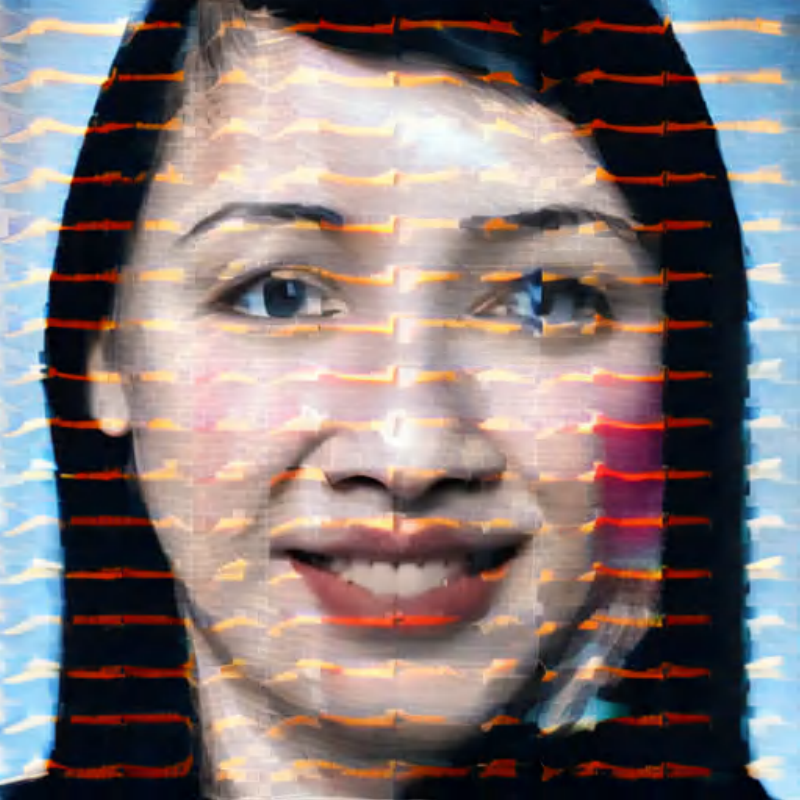}
\end{minipage}
\hfill
\begin{minipage}{0.18\linewidth}
    \centering
    \includegraphics[width=1\linewidth]{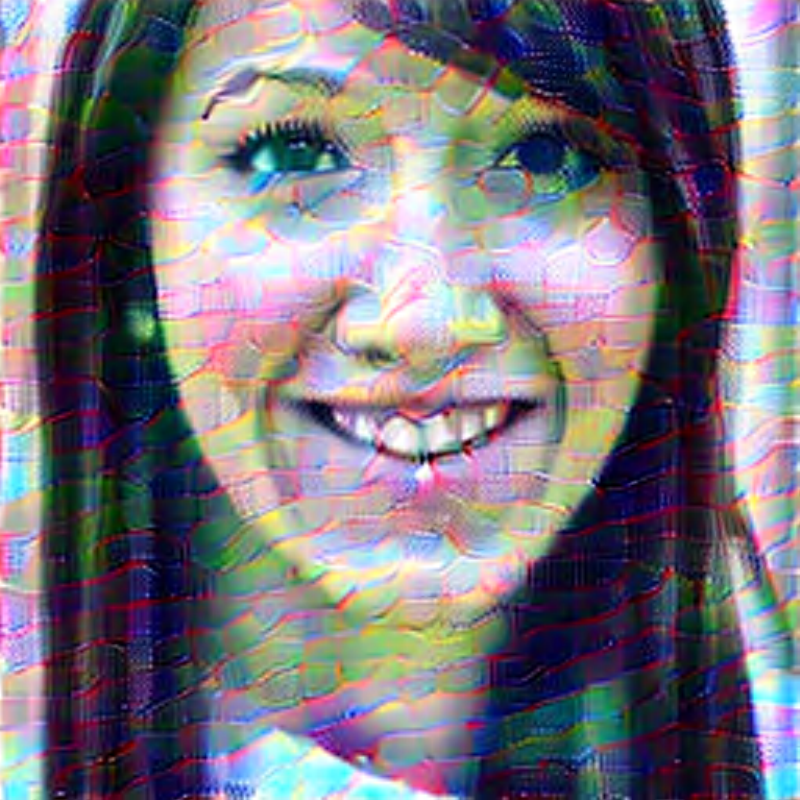}
\end{minipage}
\hfill
\begin{minipage}{0.18\linewidth}
    \centering
    \includegraphics[width=1\linewidth]{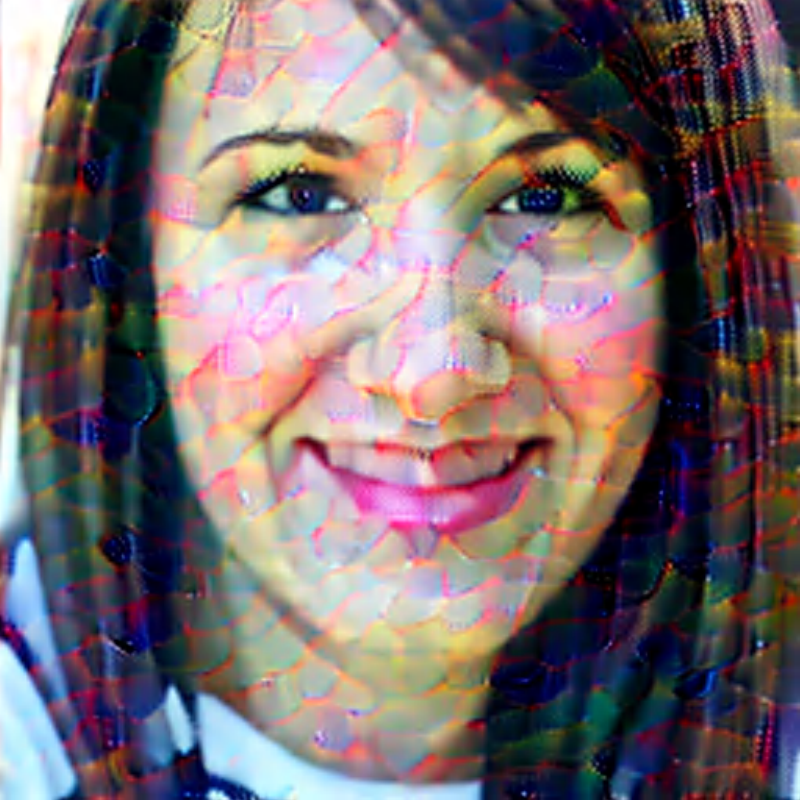}
\end{minipage}
\hfill
\begin{minipage}{0.18\linewidth}
    \centering
    \includegraphics[width=1\linewidth]{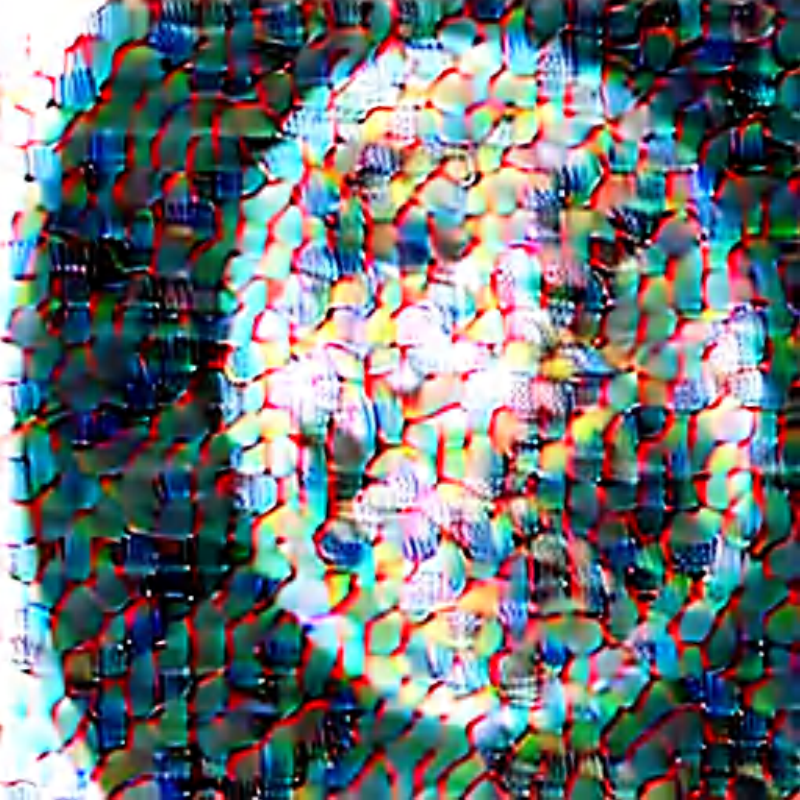}
\end{minipage}
\begin{minipage}{0.18\linewidth}
    \centering
    \includegraphics[width=1\linewidth]{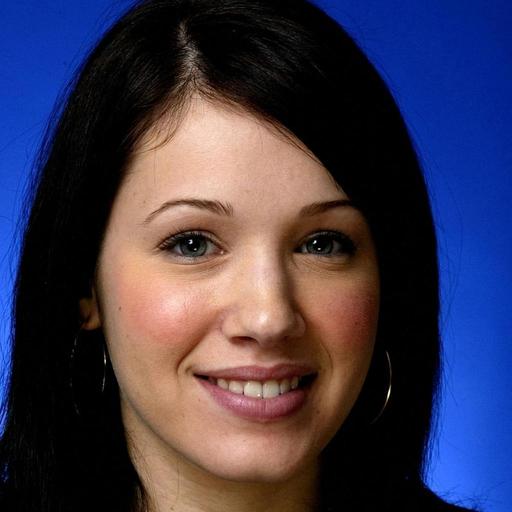}
\end{minipage}
\hfill
\begin{minipage}{0.18\linewidth}
    \centering
    \includegraphics[width=1\linewidth]{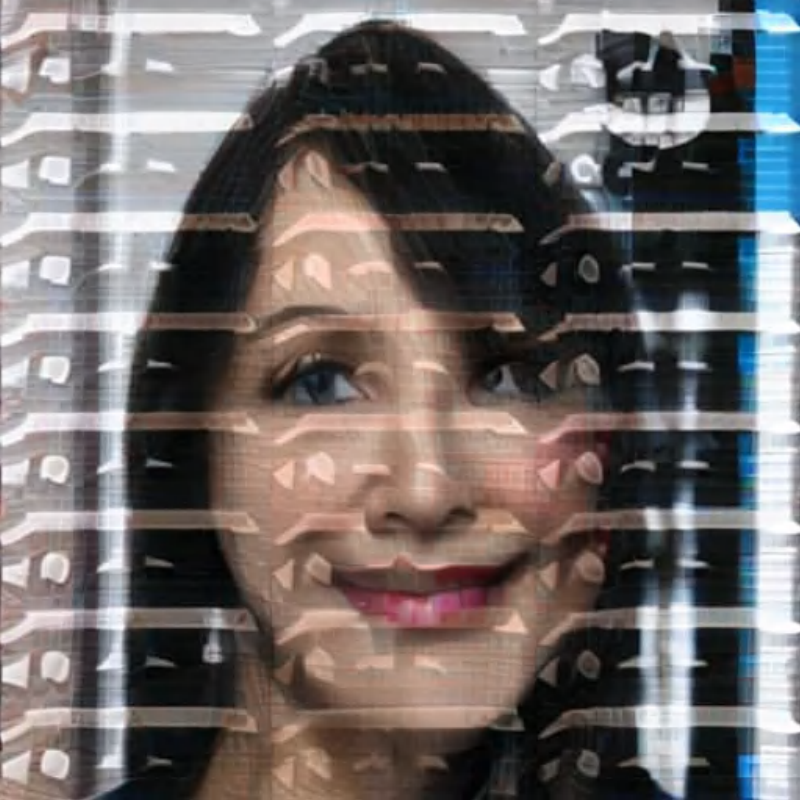}
\end{minipage}
\hfill
\begin{minipage}{0.18\linewidth}
    \centering
    \includegraphics[width=1\linewidth]{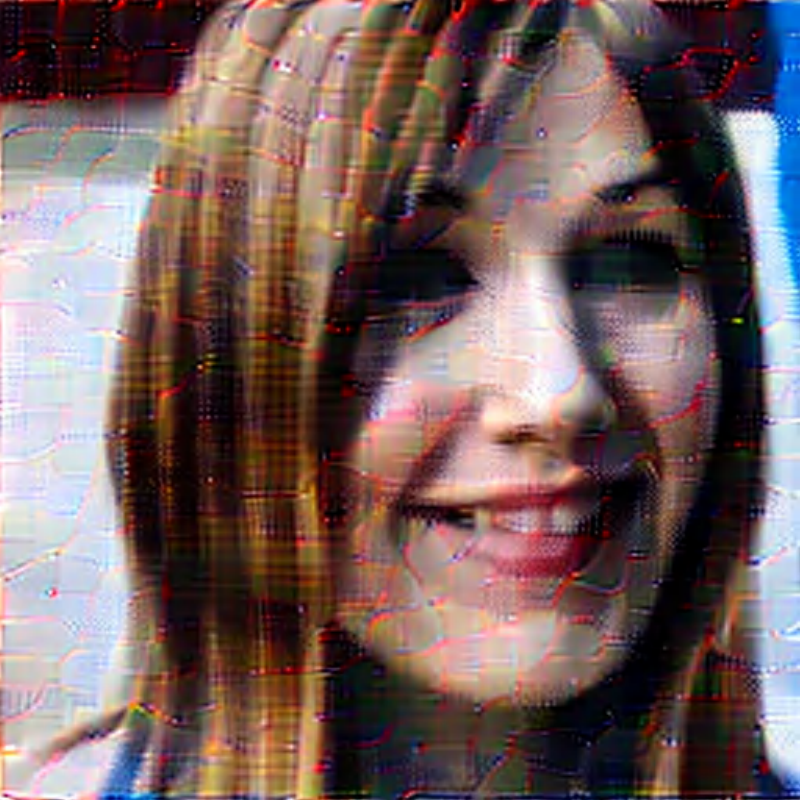}
\end{minipage}
\hfill
\begin{minipage}{0.18\linewidth}
    \centering
    \includegraphics[width=1\linewidth]{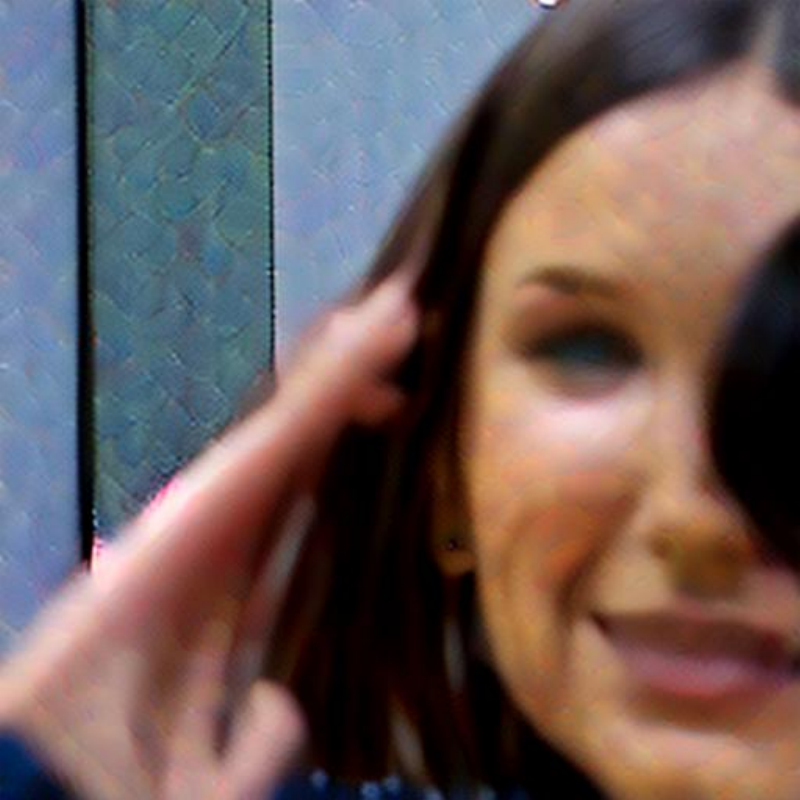}
\end{minipage}
\hfill
\begin{minipage}{0.18\linewidth}
    \centering
    \includegraphics[width=1\linewidth]{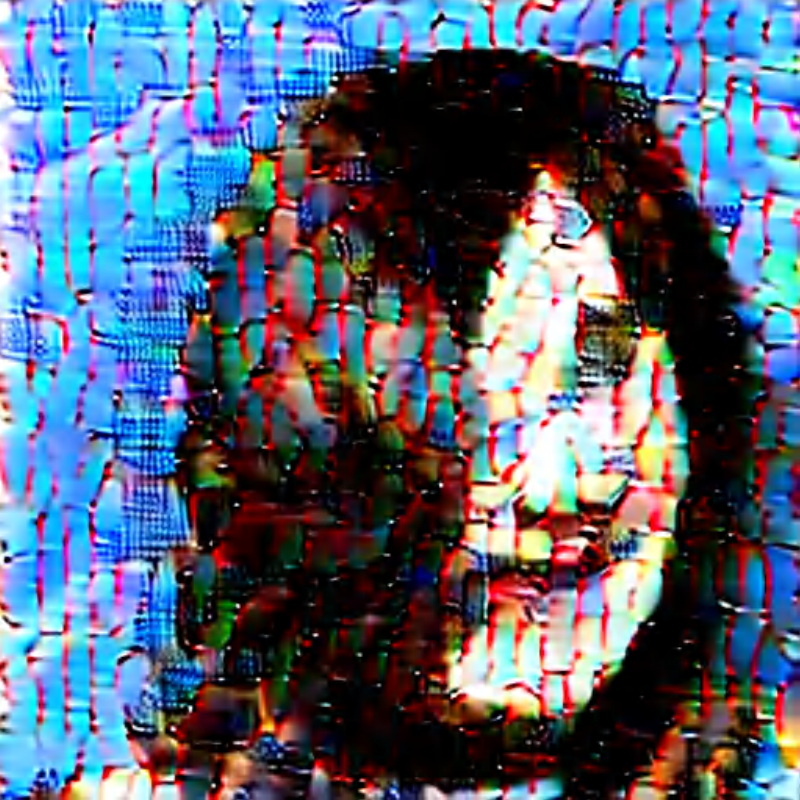}
\end{minipage}
\begin{minipage}{0.18\linewidth}
    \centering
    \includegraphics[width=1\linewidth]{fig/compare/1590.jpg}
\end{minipage}
\hfill
\begin{minipage}{0.18\linewidth}
    \centering
    \includegraphics[width=1\linewidth]{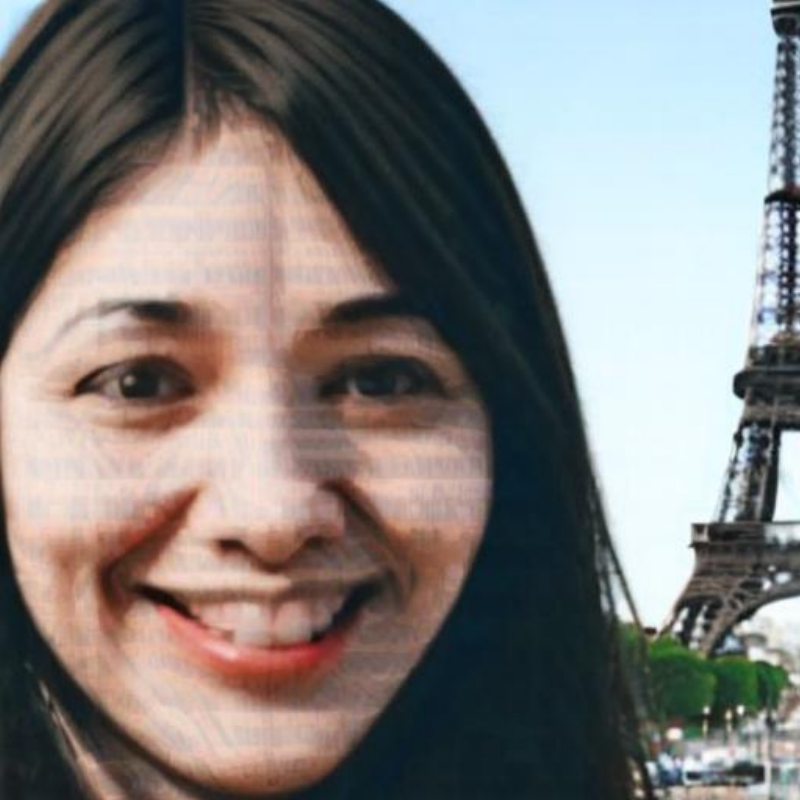}
\end{minipage}
\hfill
\begin{minipage}{0.18\linewidth}
    \centering
    \includegraphics[width=1\linewidth]{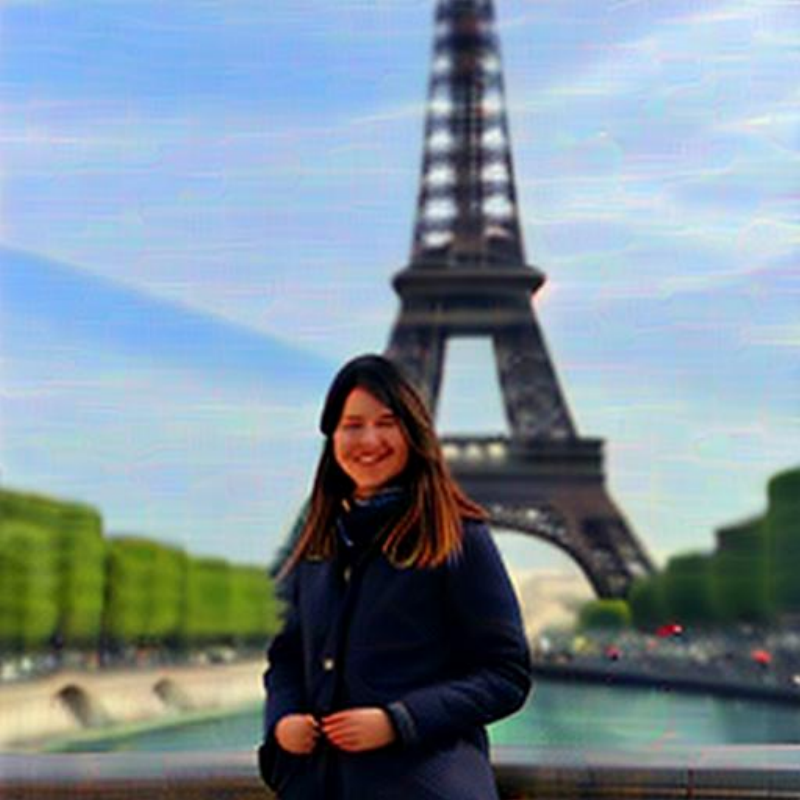}
\end{minipage}
\hfill
\begin{minipage}{0.18\linewidth}
    \centering
    \includegraphics[width=1\linewidth]{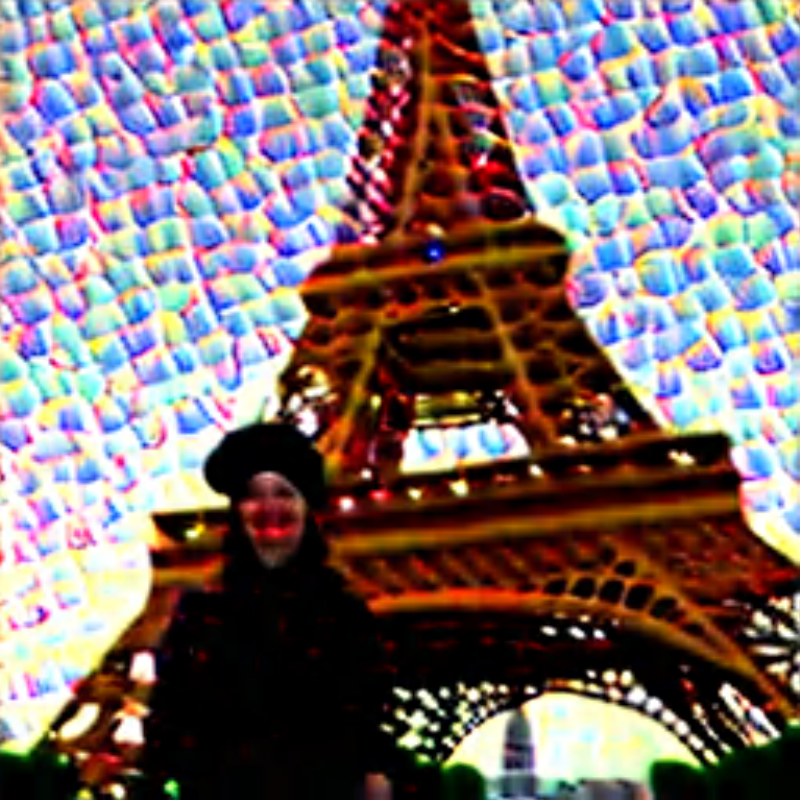}
\end{minipage}
\hfill
\begin{minipage}{0.18\linewidth}
    \centering
    \includegraphics[width=1\linewidth]{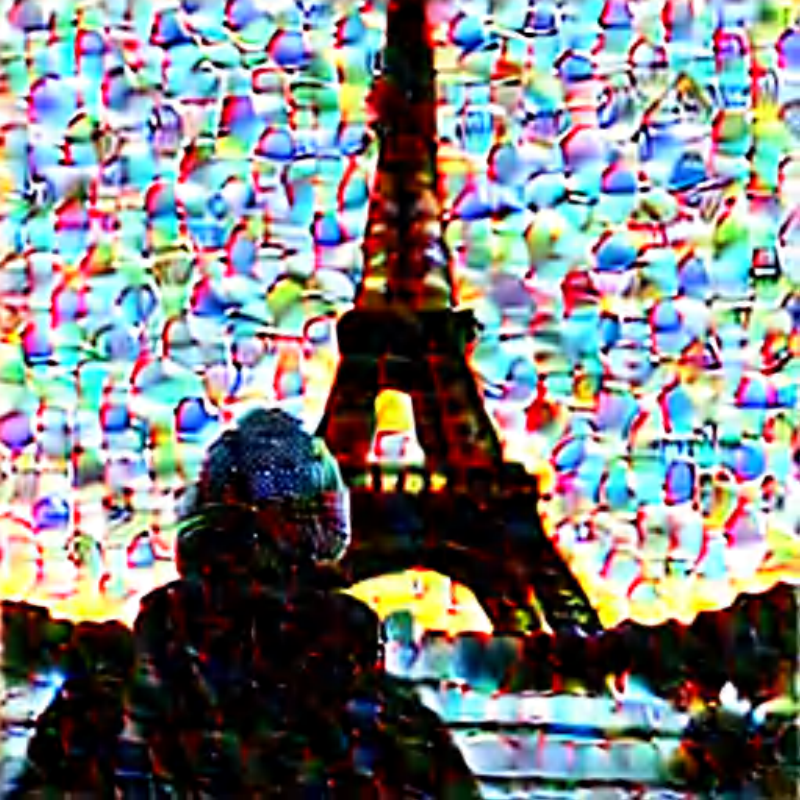}
\end{minipage}
\begin{minipage}{0.18\linewidth}
    \centering
    \includegraphics[width=1\linewidth]{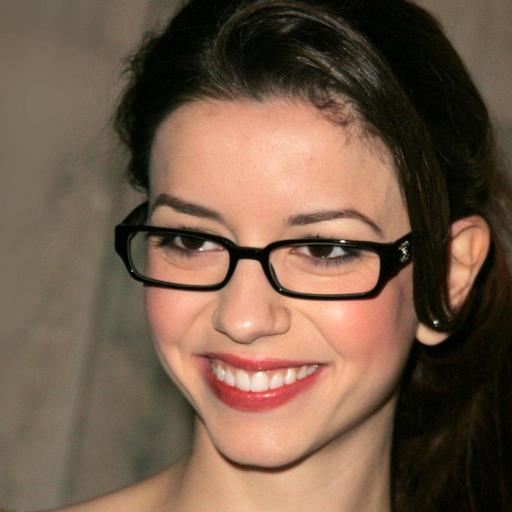}
\end{minipage}
\hfill
\begin{minipage}{0.18\linewidth}
    \centering
    \includegraphics[width=1\linewidth]{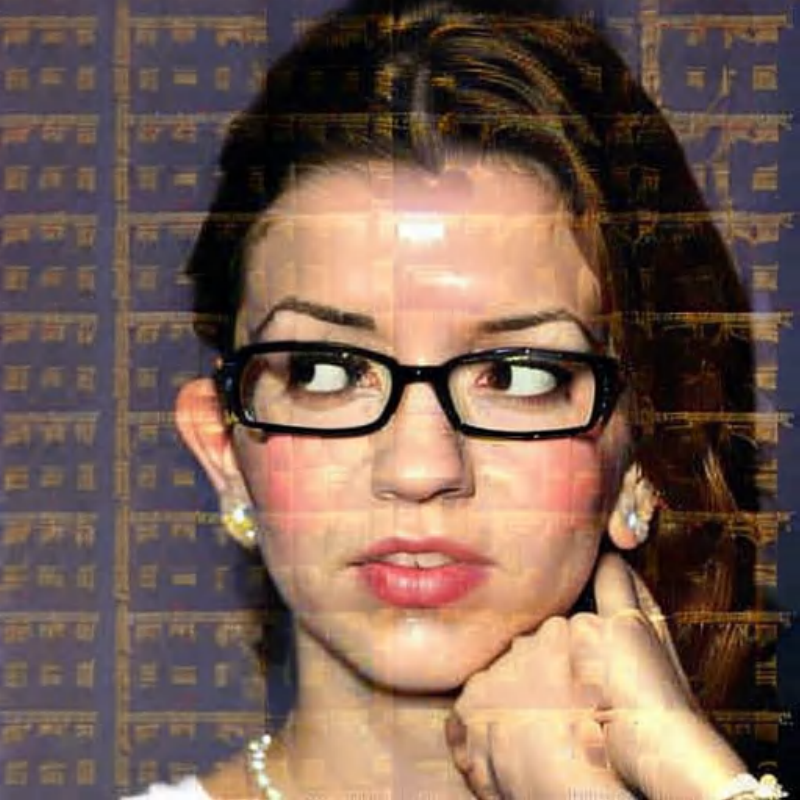}
\end{minipage}
\hfill
\begin{minipage}{0.18\linewidth}
    \centering
    \includegraphics[width=1\linewidth]{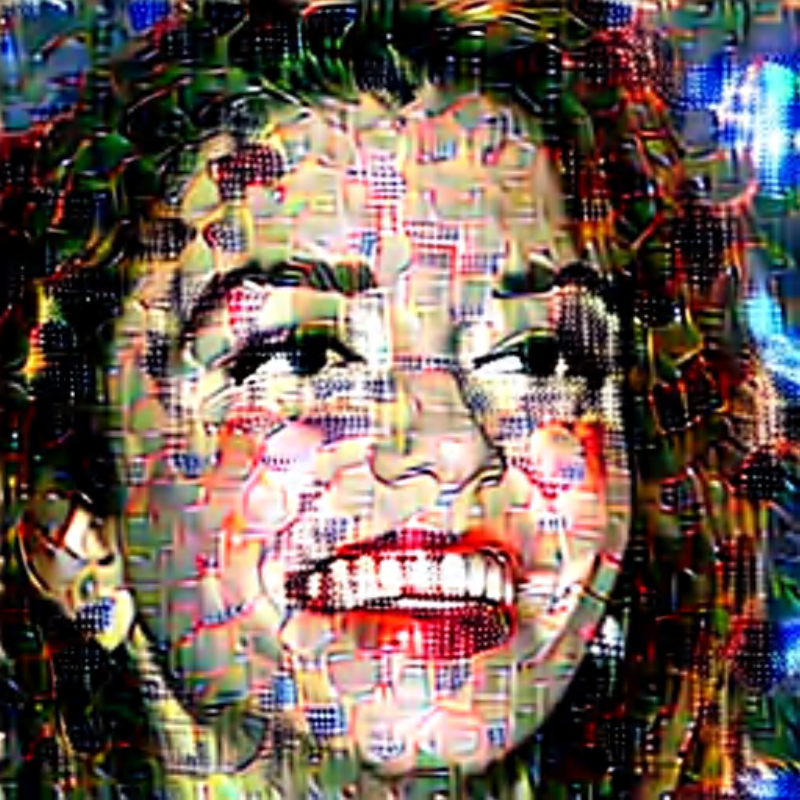}
\end{minipage}
\hfill
\begin{minipage}{0.18\linewidth}
    \centering
    \includegraphics[width=1\linewidth]{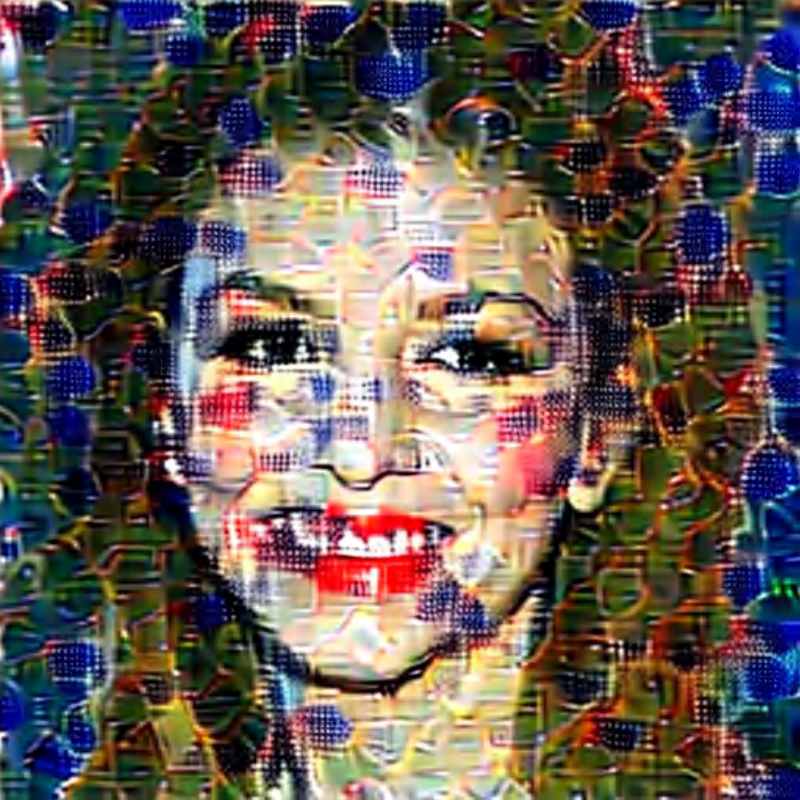}
\end{minipage}
\hfill
\begin{minipage}{0.18\linewidth}
    \centering
    \includegraphics[width=1\linewidth]{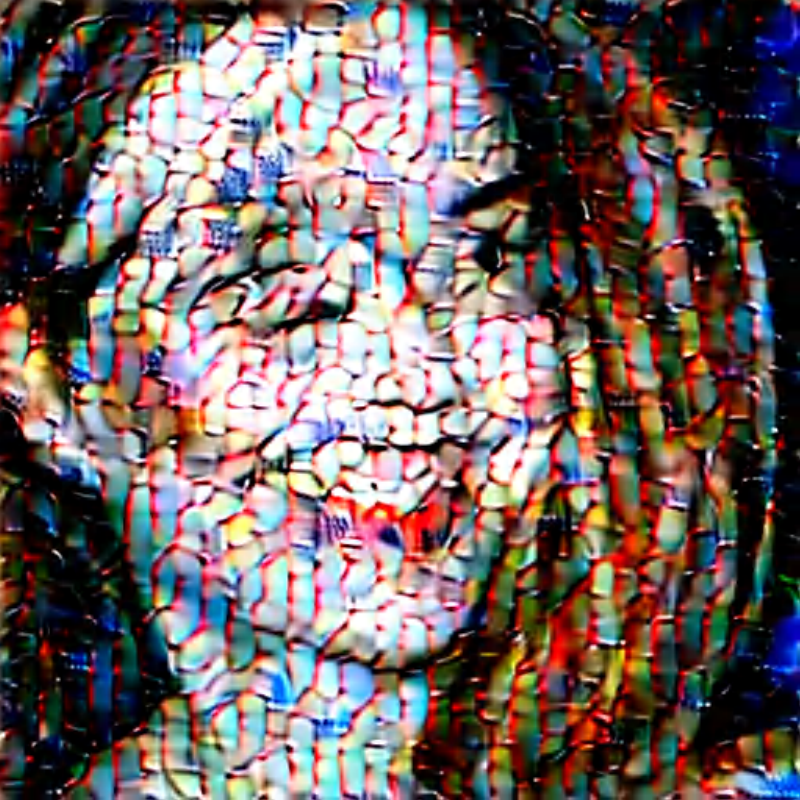}
\end{minipage}
\begin{minipage}{0.18\linewidth}
    \centering
    \includegraphics[width=1\linewidth]{fig/compare2/3731.jpg}
\end{minipage}
\hfill
\begin{minipage}{0.18\linewidth}
    \centering
    \includegraphics[width=1\linewidth]{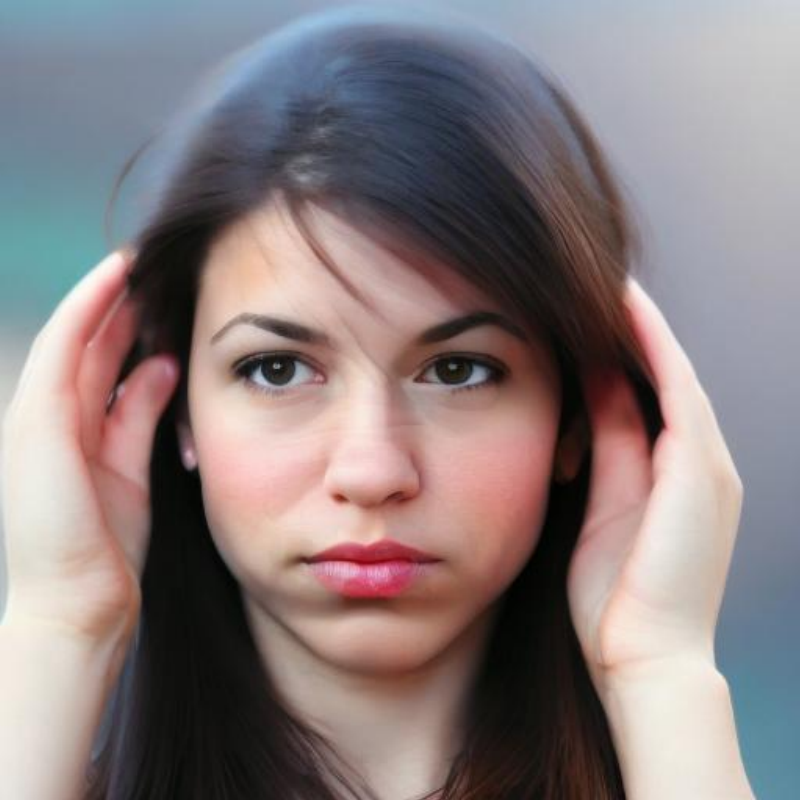}
\end{minipage}
\hfill
\begin{minipage}{0.18\linewidth}
    \centering
    \includegraphics[width=1\linewidth]{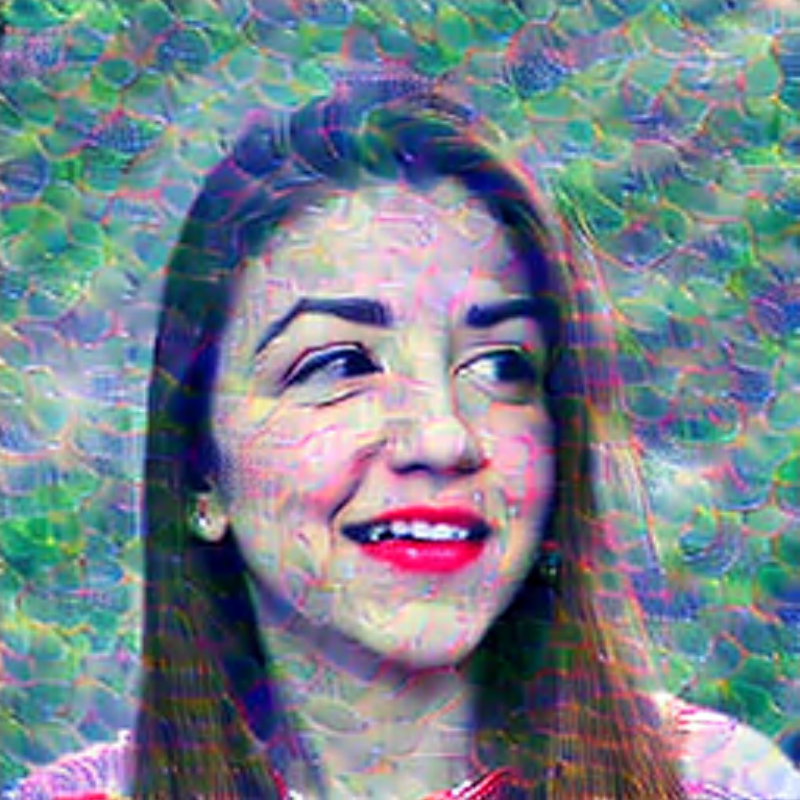}
\end{minipage}
\hfill
\begin{minipage}{0.18\linewidth}
    \centering
    \includegraphics[width=1\linewidth]{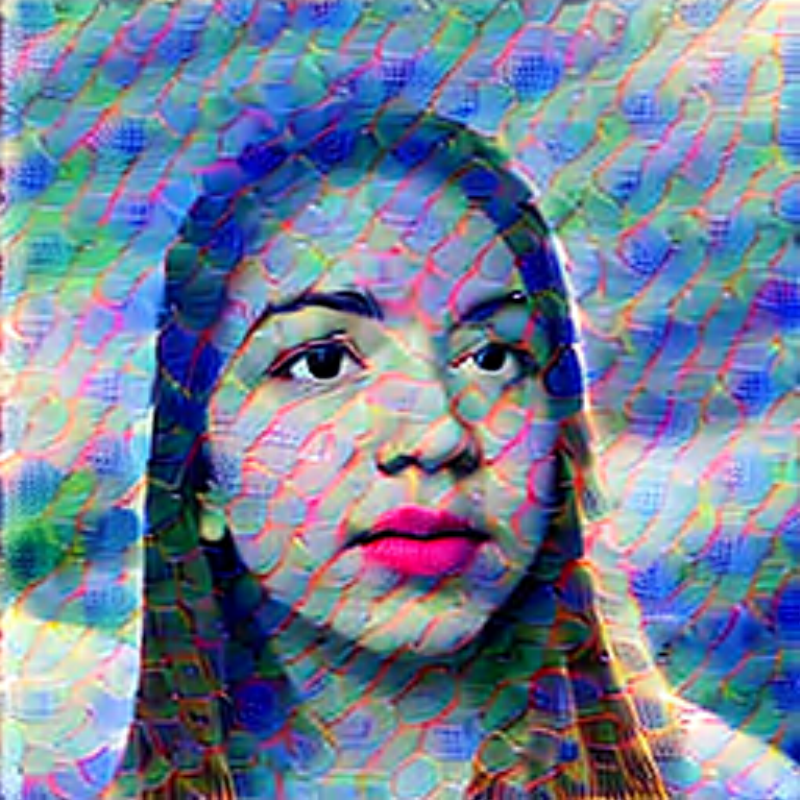}
\end{minipage}
\hfill
\begin{minipage}{0.18\linewidth}
    \centering
    \includegraphics[width=1\linewidth]{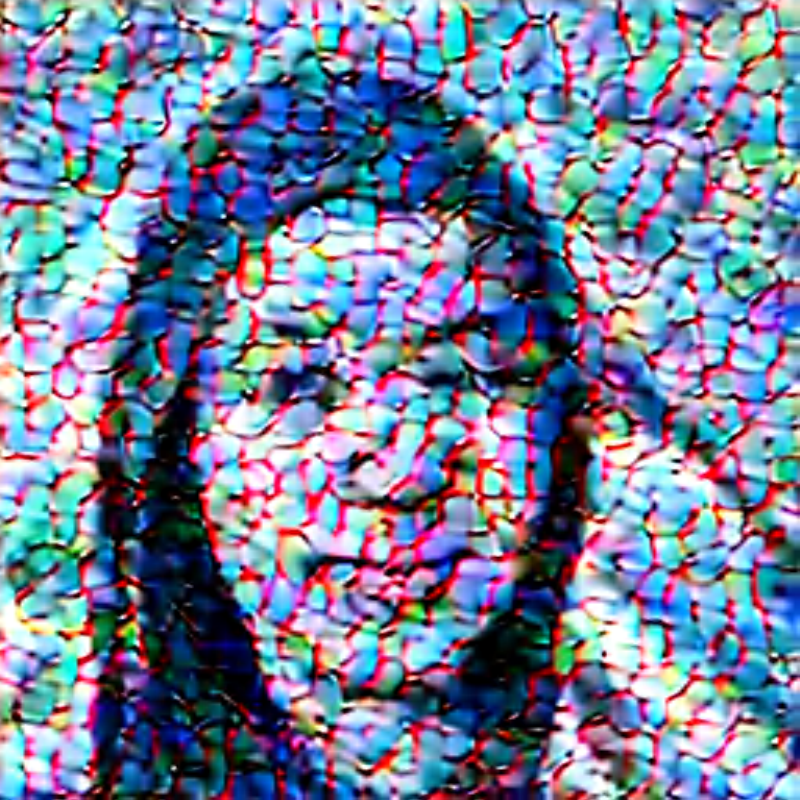}
\end{minipage}
\begin{minipage}{0.18\linewidth}
    \centering
    \includegraphics[width=1\linewidth]{fig/compare2/3731.jpg}
\end{minipage}
\hfill
\begin{minipage}{0.18\linewidth}
    \centering
    \includegraphics[width=1\linewidth]{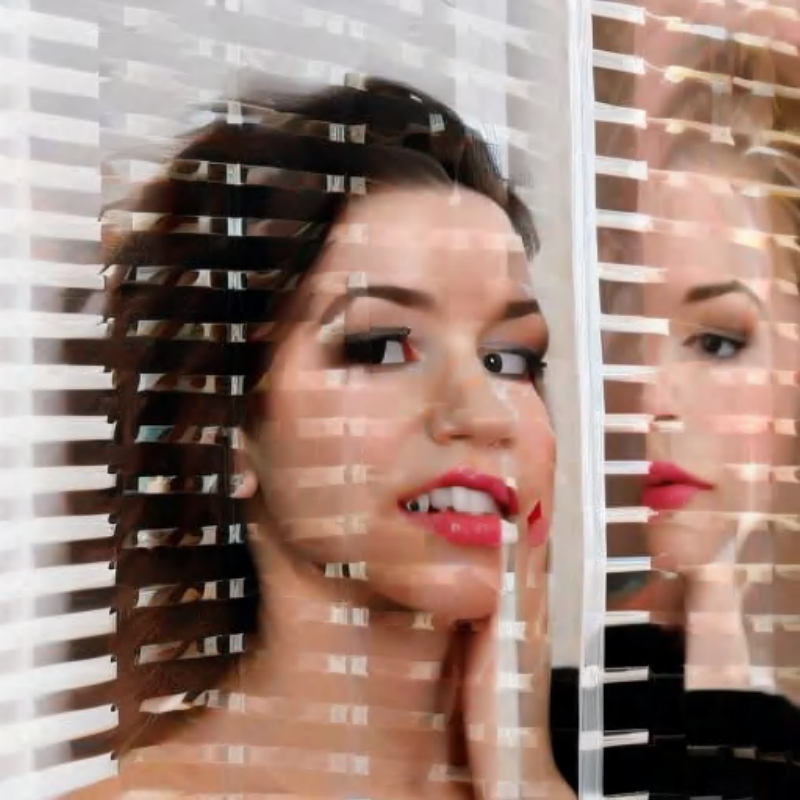}
\end{minipage}
\hfill
\begin{minipage}{0.18\linewidth}
    \centering
    \includegraphics[width=1\linewidth]{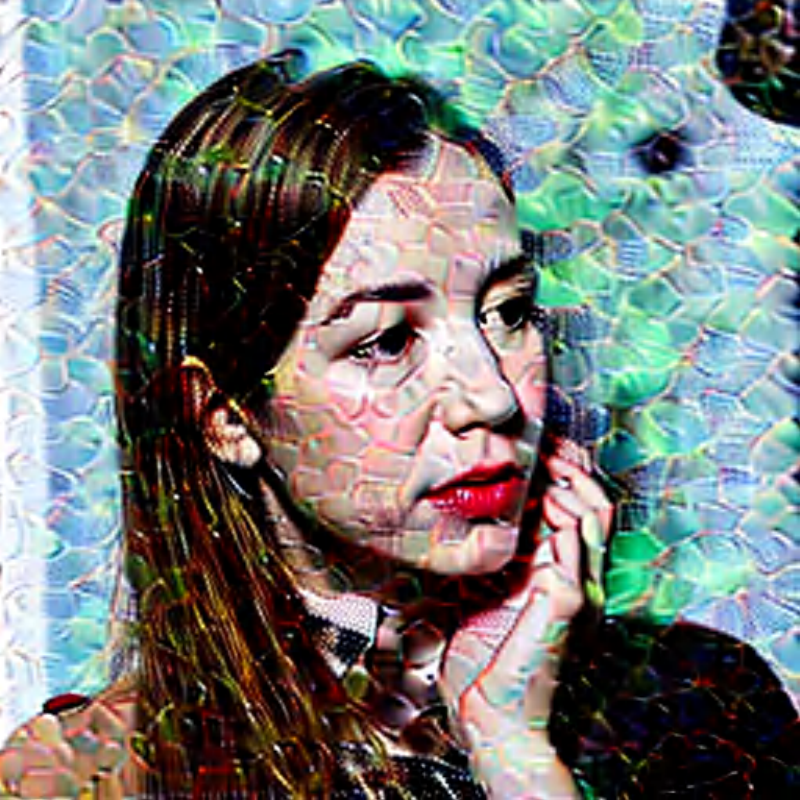}
\end{minipage}
\hfill
\begin{minipage}{0.18\linewidth}
    \centering
    \includegraphics[width=1\linewidth]{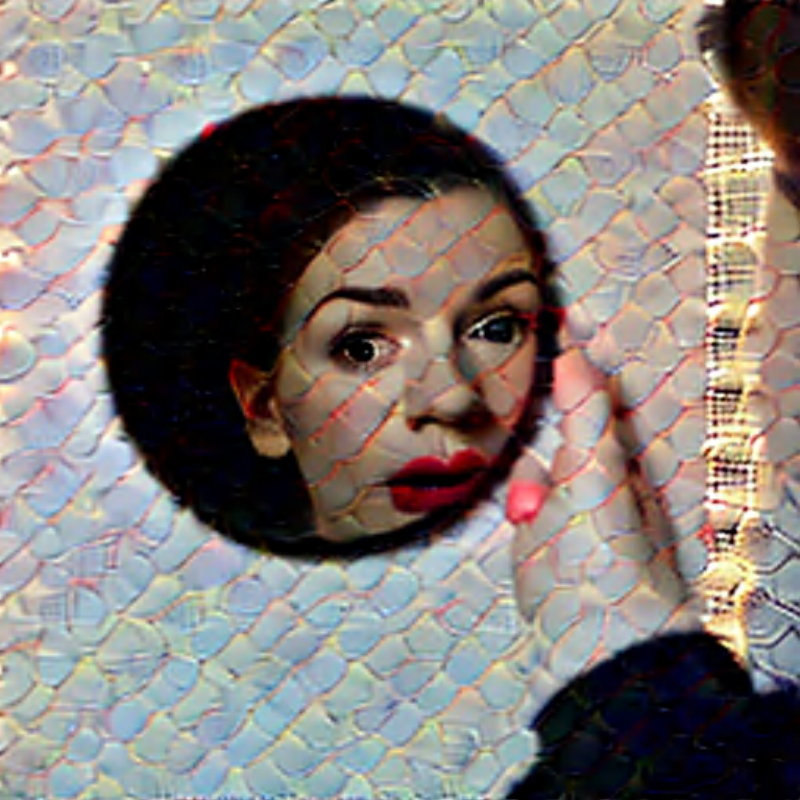}
\end{minipage}
\hfill
\begin{minipage}{0.18\linewidth}
    \centering
    \includegraphics[width=1\linewidth]{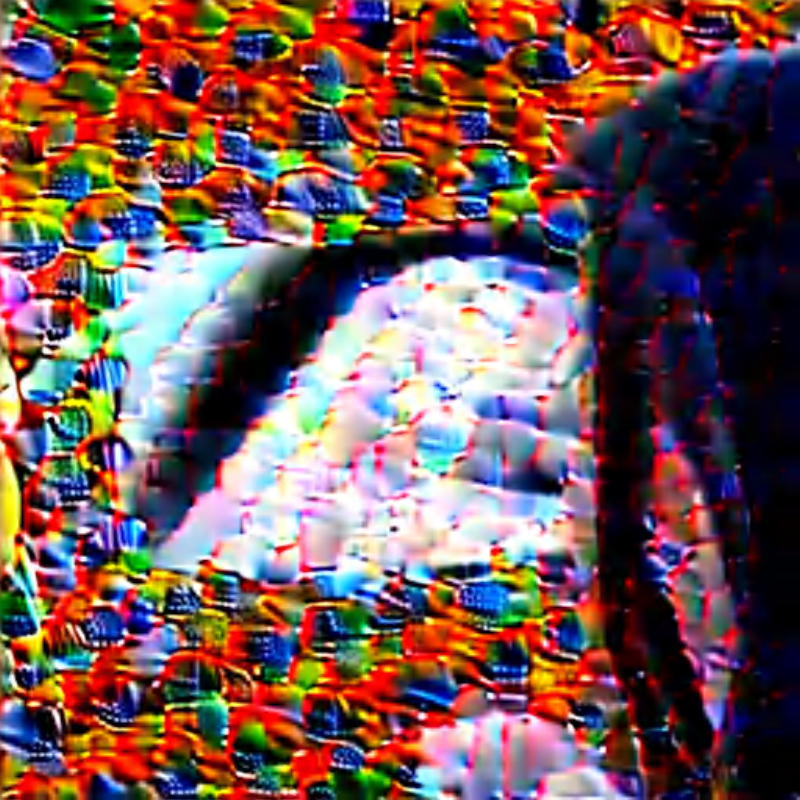}
\end{minipage}
\begin{minipage}{0.18\linewidth}
    \centering
    \includegraphics[width=1\linewidth]{fig/compare2/3731.jpg}
    \centerline{\scriptsize  \textbf{Portrait Image}}
\end{minipage}
\hfill
\begin{minipage}{0.18\linewidth}
    \centering
    \includegraphics[width=1\linewidth]{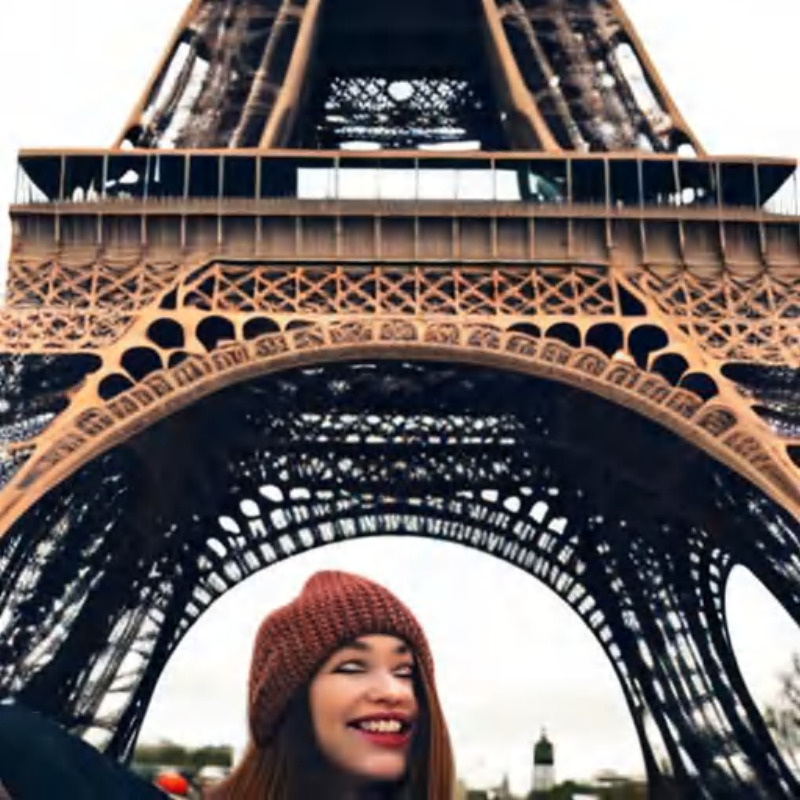}
    \centerline{\scriptsize \textbf {PhotoGuard}}
\end{minipage}
\hfill
\begin{minipage}{0.18\linewidth}
    \centering
    \includegraphics[width=1\linewidth]{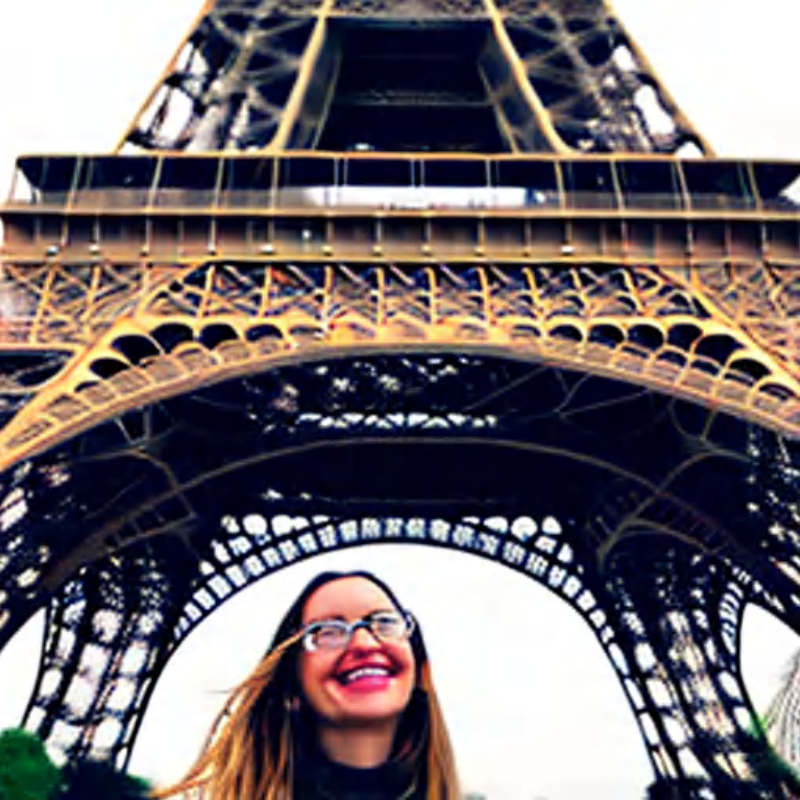}
    \centerline{\scriptsize \textbf{AdvDM}}
\end{minipage}
\hfill
\begin{minipage}{0.18\linewidth}
    \centering
    \includegraphics[width=1\linewidth]{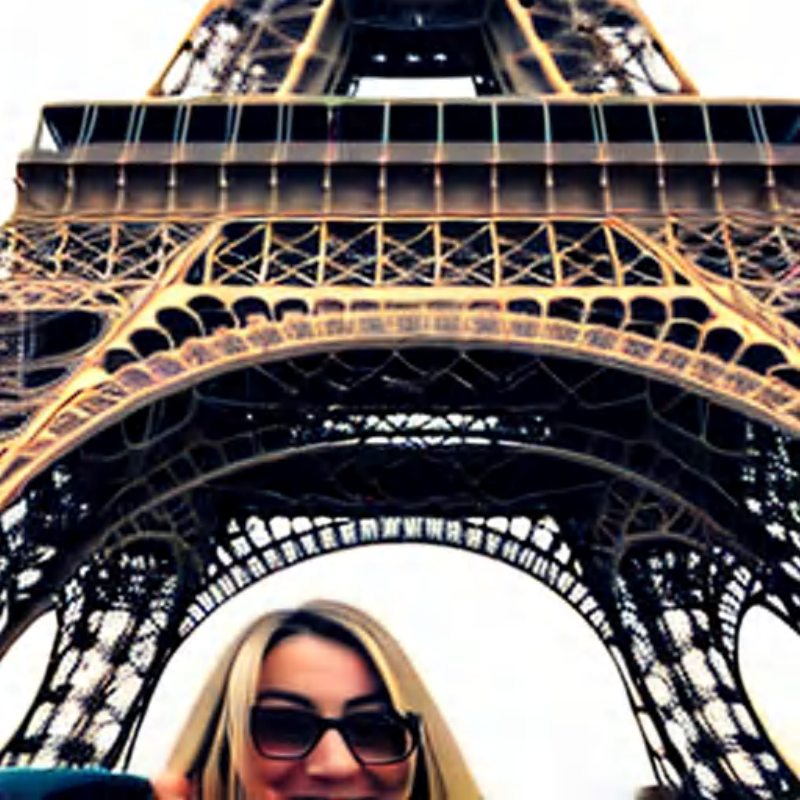}
    \centerline{\scriptsize  \textbf{Anti-DB}}
\end{minipage}
\hfill
\begin{minipage}{0.18\linewidth}
    \centering
    \includegraphics[width=1\linewidth]{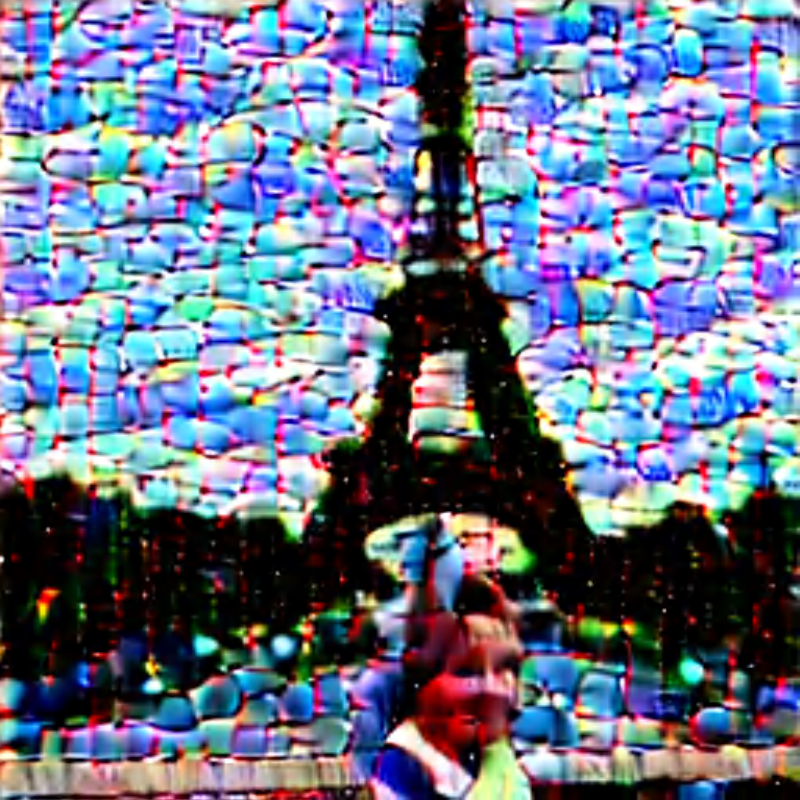}
    \centerline{\scriptsize  \textbf{Anti-DB+SimAC}}
\end{minipage}
\vspace{-0.5em}
\caption{Visualization results (four prompts) on Celeb-HQ. From the first row to the last row is ``a photo of sks person'',``a dslr portrait of sks person'', ``a photo of sks person looking at the mirror'', ``a photo of sks person in front of eiffel tower". }
\label{fig:Quantitative}
\vspace{-1em}
\end{figure}

\begin{figure}[h]
\centering
\begin{minipage}{0.18\linewidth}
    \centering
    \includegraphics[width=1\linewidth]{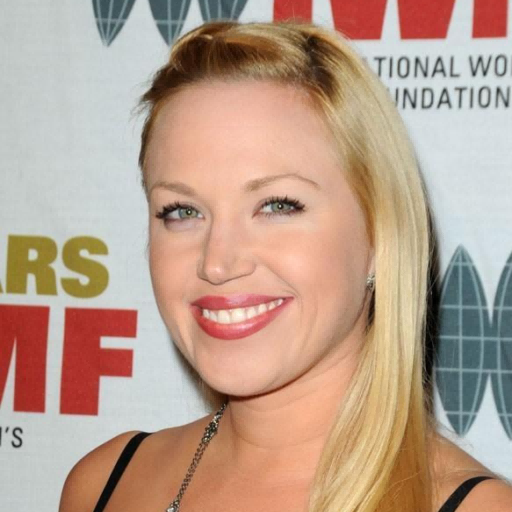}
\end{minipage}
\begin{minipage}{0.18\linewidth}
    \centering
    \includegraphics[width=1\linewidth]{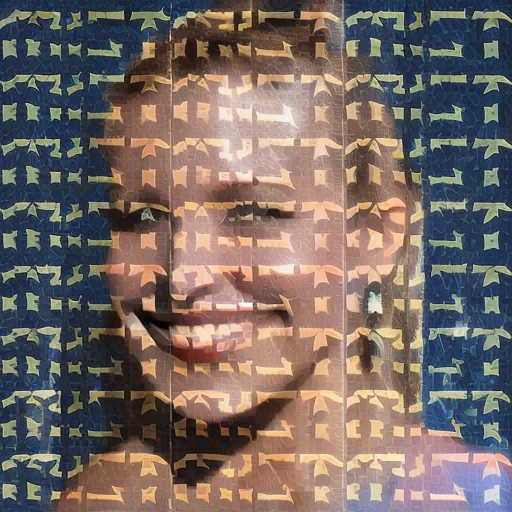}
\end{minipage}
\begin{minipage}{0.18\linewidth}
    \centering
    \includegraphics[width=1\linewidth]{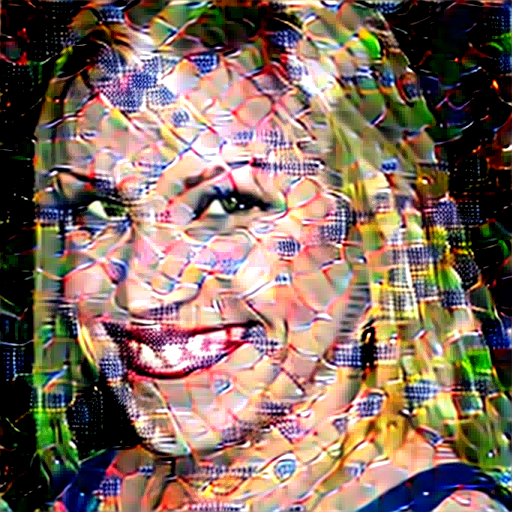}
\end{minipage}
\begin{minipage}{0.18\linewidth}
    \centering
    \includegraphics[width=1\linewidth]{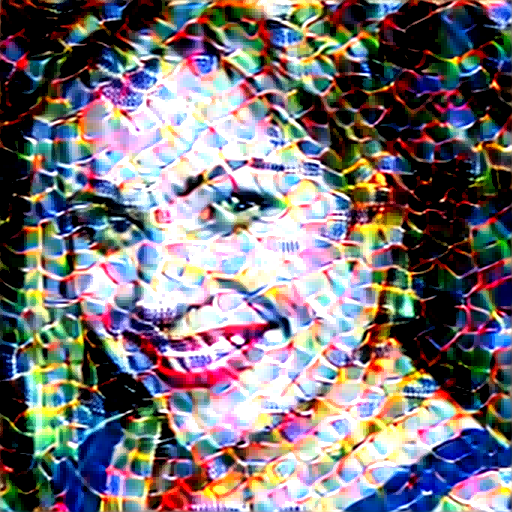}
\end{minipage}
\begin{minipage}{0.18\linewidth}
    \centering
    \includegraphics[width=1\linewidth]{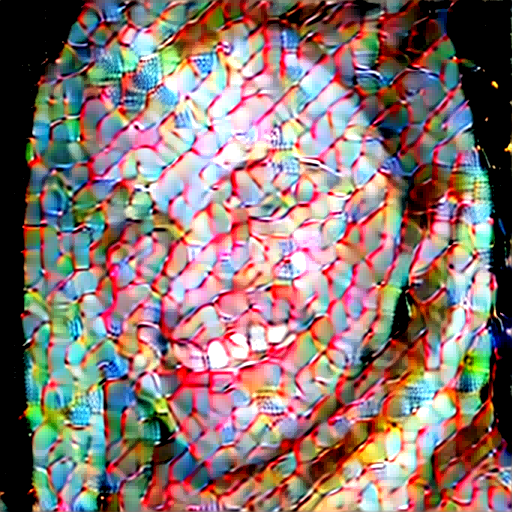}
\end{minipage}
\\
\begin{minipage}{0.18\linewidth}
    \centering
    \includegraphics[width=1\linewidth]{fig/supp/vgg_compare/0027_01.png}
\end{minipage}
\begin{minipage}{0.18\linewidth}
    \centering
    \includegraphics[width=1\linewidth]{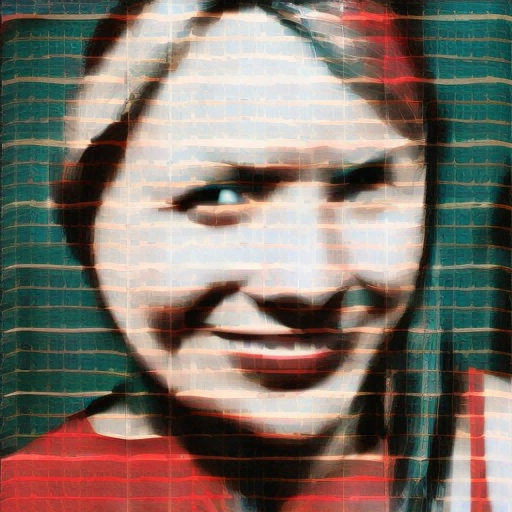}
\end{minipage}
\begin{minipage}{0.18\linewidth}
    \centering
    \includegraphics[width=1\linewidth]{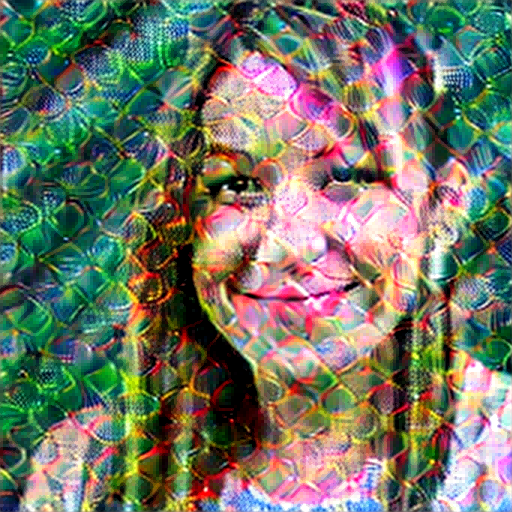}
\end{minipage}
\begin{minipage}{0.18\linewidth}
    \centering
    \includegraphics[width=1\linewidth]{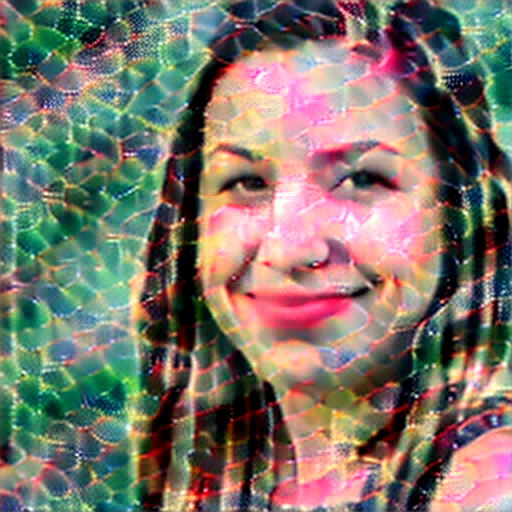}
\end{minipage}
\begin{minipage}{0.18\linewidth}
    \centering
    \includegraphics[width=1\linewidth]{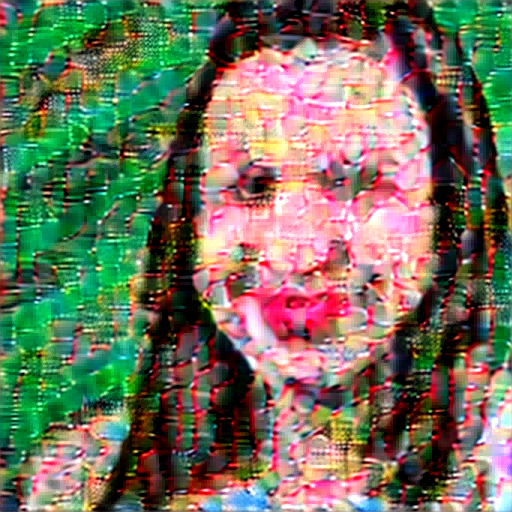}
\end{minipage}
\\
\begin{minipage}{0.18\linewidth}
    \centering
    \includegraphics[width=1\linewidth]{fig/supp/vgg_compare/0027_01.png}
\end{minipage}
\begin{minipage}{0.18\linewidth}
    \centering
    \includegraphics[width=1\linewidth]{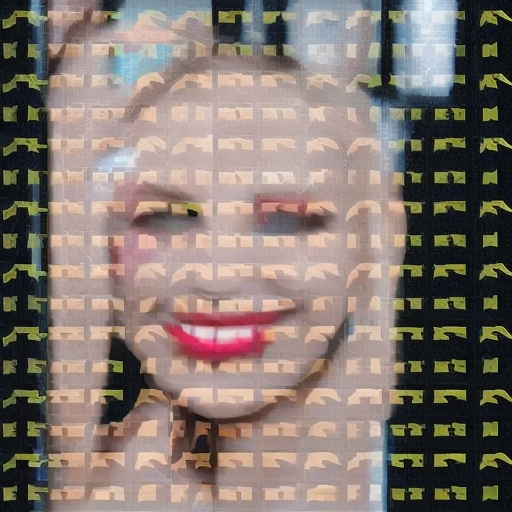}
\end{minipage}
\begin{minipage}{0.18\linewidth}
    \centering
    \includegraphics[width=1\linewidth]{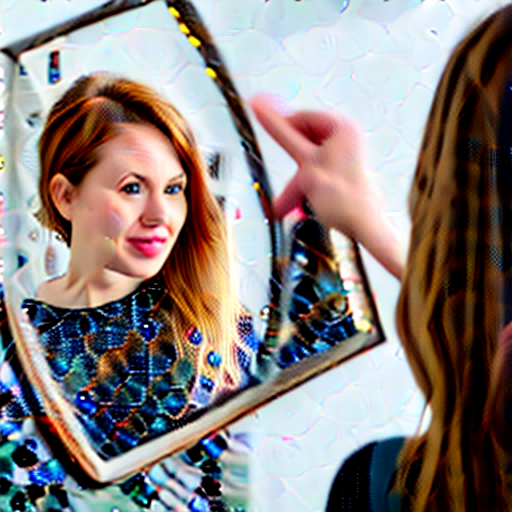}
\end{minipage}
\begin{minipage}{0.18\linewidth}
    \centering
    \includegraphics[width=1\linewidth]{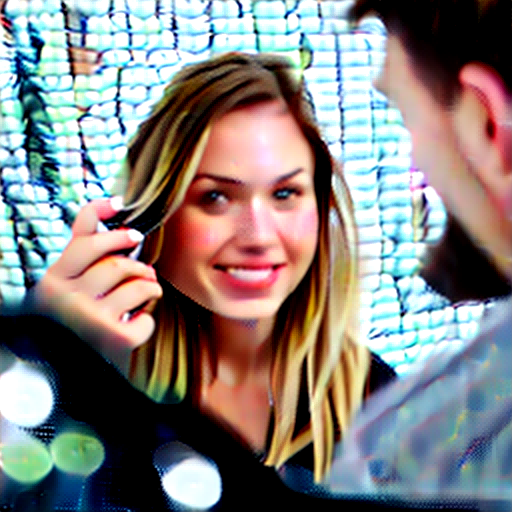}
\end{minipage}
\begin{minipage}{0.18\linewidth}
    \centering
    \includegraphics[width=1\linewidth]{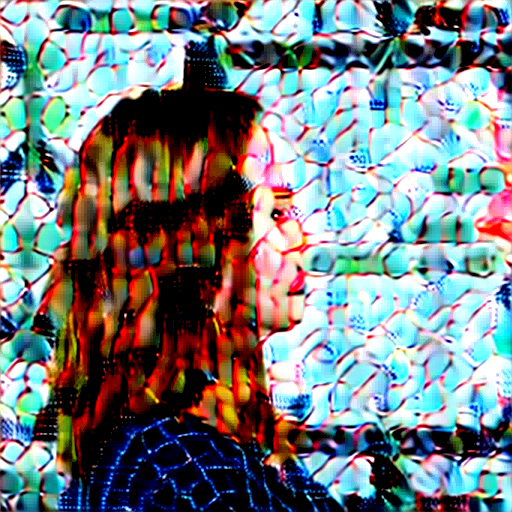}
\end{minipage}
\\
\begin{minipage}{0.18\linewidth}
    \centering
    \includegraphics[width=1\linewidth]{fig/supp/vgg_compare/0027_01.png}
\end{minipage}
\begin{minipage}{0.18\linewidth}
    \centering
    \includegraphics[width=1\linewidth]{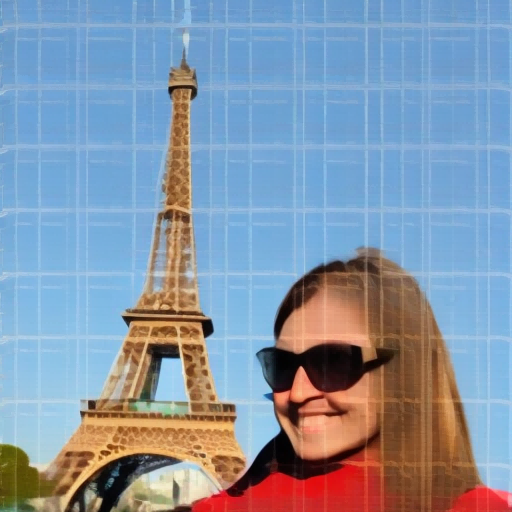}
\end{minipage}
\begin{minipage}{0.18\linewidth}
    \centering
    \includegraphics[width=1\linewidth]{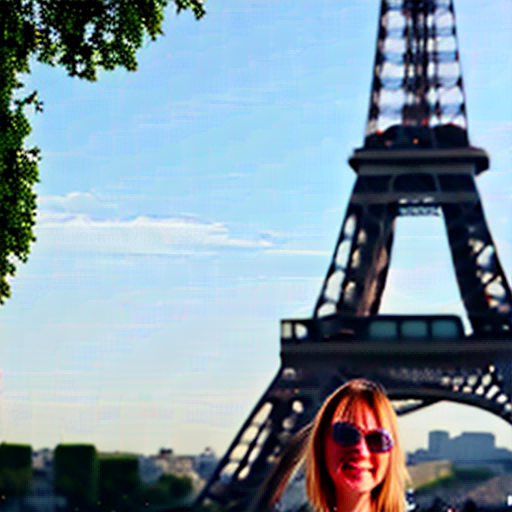}
\end{minipage}
\begin{minipage}{0.18\linewidth}
    \centering
    \includegraphics[width=1\linewidth]{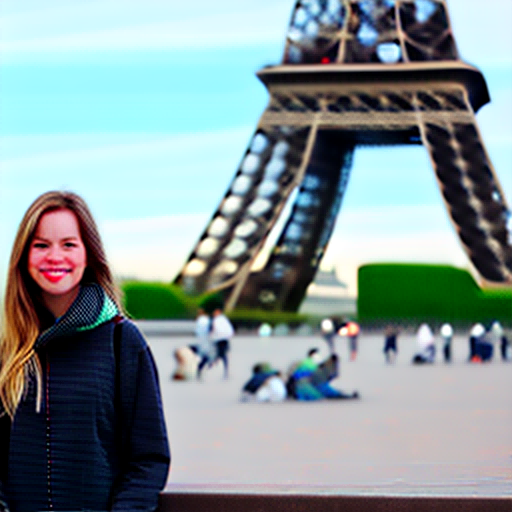}
\end{minipage}
\begin{minipage}{0.18\linewidth}
    \centering
    \includegraphics[width=1\linewidth]{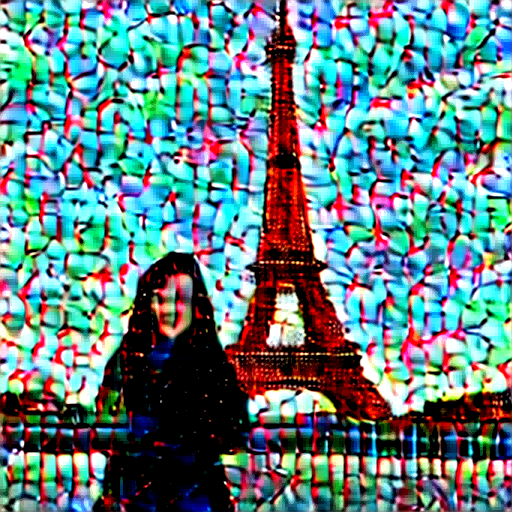}
\end{minipage}
\\
\begin{minipage}{0.18\linewidth}
    \centering
    \includegraphics[width=1\linewidth]{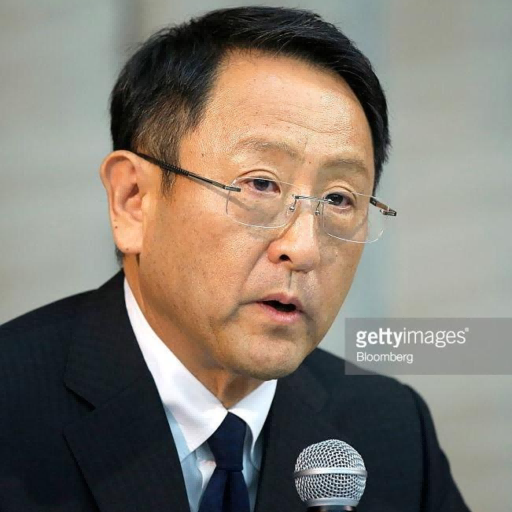}
\end{minipage}
\begin{minipage}{0.18\linewidth}
    \centering
    \includegraphics[width=1\linewidth]{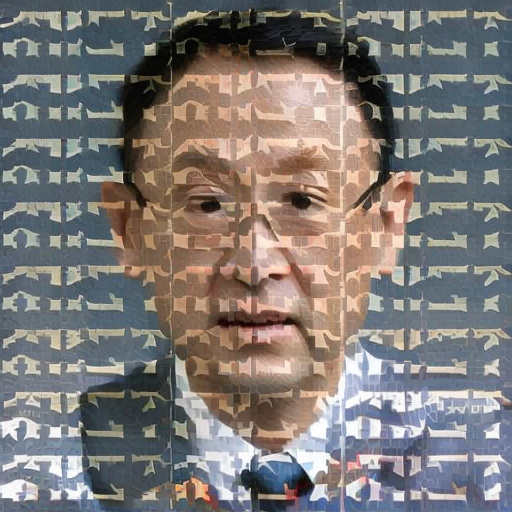}
\end{minipage}
\begin{minipage}{0.18\linewidth}
    \centering
    \includegraphics[width=1\linewidth]{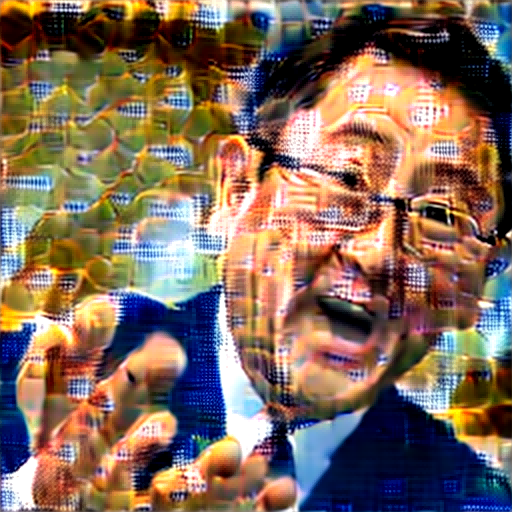}
\end{minipage}
\begin{minipage}{0.18\linewidth}
    \centering
    \includegraphics[width=1\linewidth]{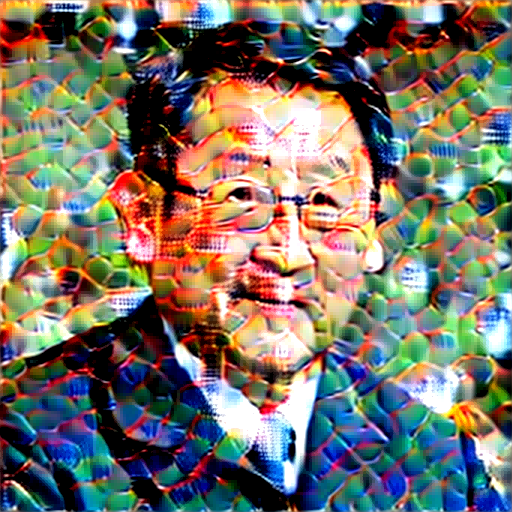}
\end{minipage}
\begin{minipage}{0.18\linewidth}
    \centering
    \includegraphics[width=1\linewidth]{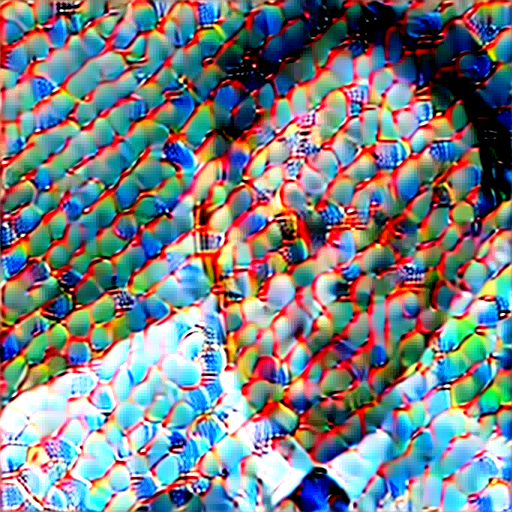}
\end{minipage}
\\
\begin{minipage}{0.18\linewidth}
    \centering
    \includegraphics[width=1\linewidth]{fig/supp/vgg_compare2/0069_01.png}
\end{minipage}
\begin{minipage}{0.18\linewidth}
    \centering
    \includegraphics[width=1\linewidth]{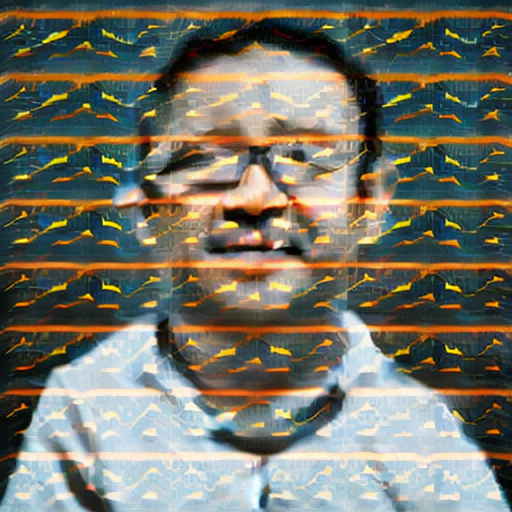}
\end{minipage}
\begin{minipage}{0.18\linewidth}
    \centering
    \includegraphics[width=1\linewidth]{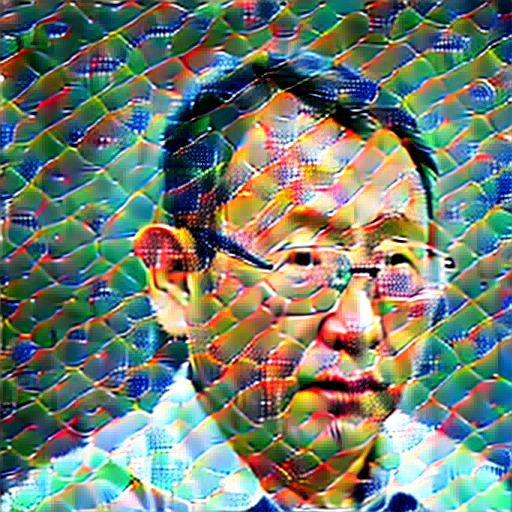}
\end{minipage}
\begin{minipage}{0.18\linewidth}
    \centering
    \includegraphics[width=1\linewidth]{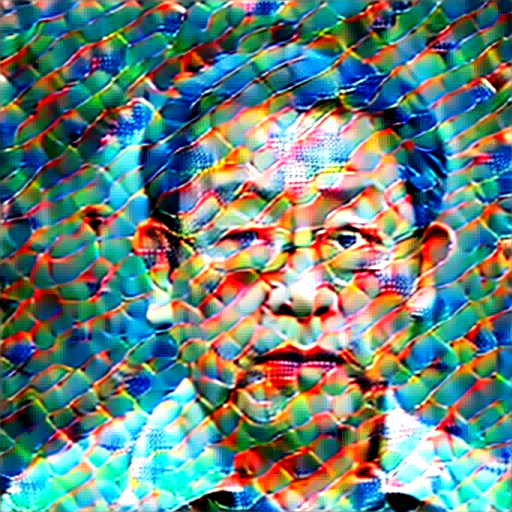}
\end{minipage}
\begin{minipage}{0.18\linewidth}
    \centering
    \includegraphics[width=1\linewidth]{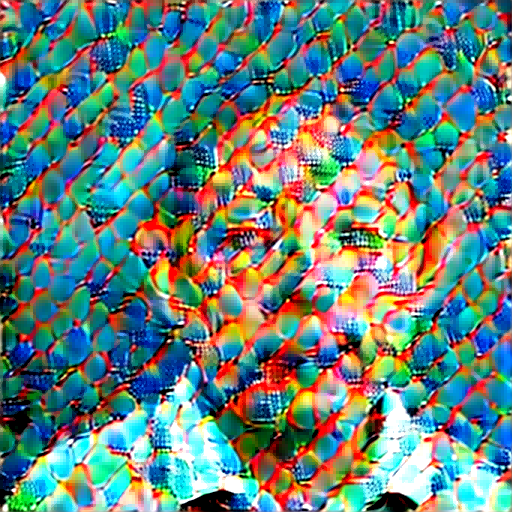}
\end{minipage}
\\
\begin{minipage}{0.18\linewidth}
    \centering
    \includegraphics[width=1\linewidth]{fig/supp/vgg_compare2/0069_01.png}
\end{minipage}
\begin{minipage}{0.18\linewidth}
    \centering
    \includegraphics[width=1\linewidth]{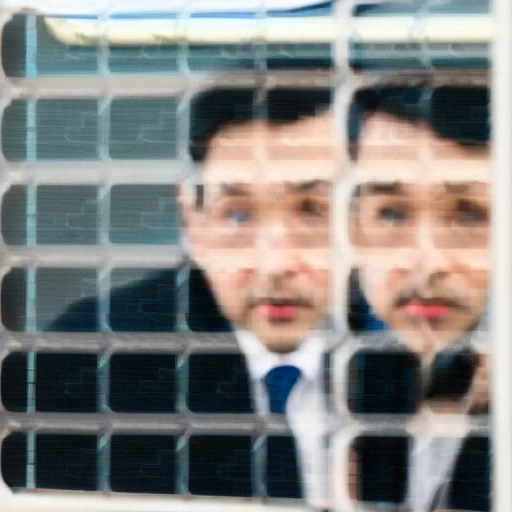}
\end{minipage}
\begin{minipage}{0.18\linewidth}
    \centering
    \includegraphics[width=1\linewidth]{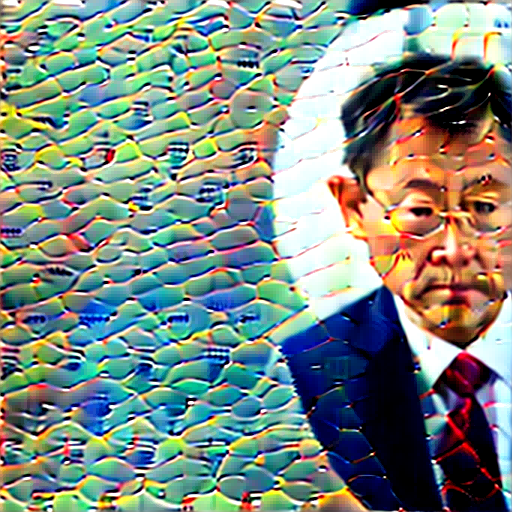}
\end{minipage}
\begin{minipage}{0.18\linewidth}
    \centering
    \includegraphics[width=1\linewidth]{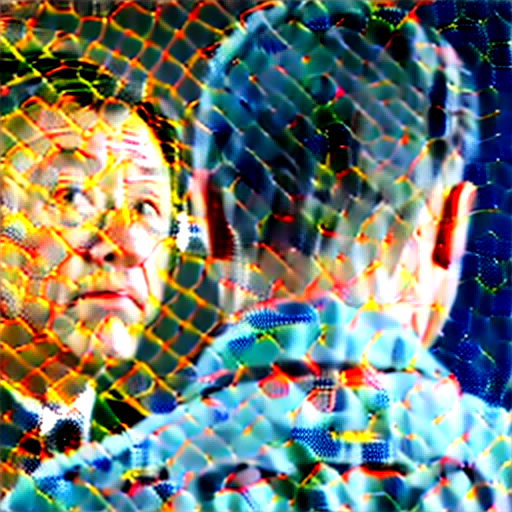}
\end{minipage}
\begin{minipage}{0.18\linewidth}
    \centering
    \includegraphics[width=1\linewidth]{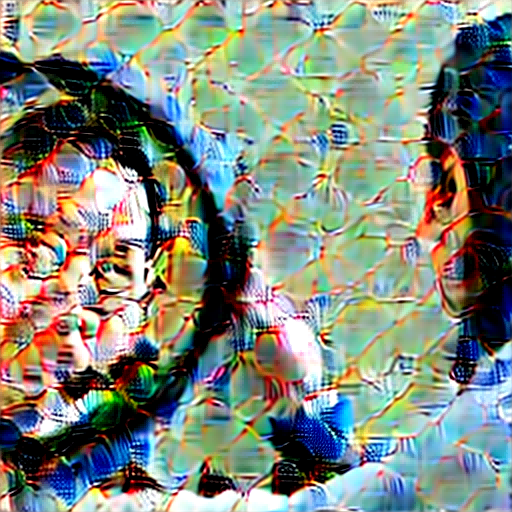}
\end{minipage}
\\
\begin{minipage}{0.18\linewidth}
    \centering
    \includegraphics[width=1\linewidth]{fig/supp/vgg_compare2/0069_01.png}
    \centerline{\scriptsize  \textbf{Portrait image}}
\end{minipage}
\begin{minipage}{0.18\linewidth}
    \centering
    \includegraphics[width=1\linewidth]{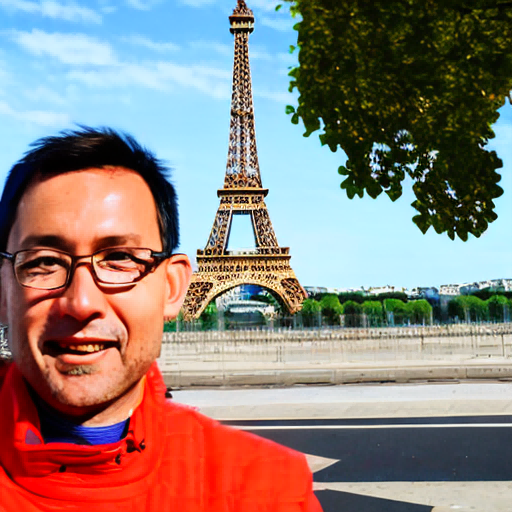}
    \centerline{\scriptsize \textbf {PhotoGuard}}
\end{minipage}
\begin{minipage}{0.18\linewidth}
    \centering
    \includegraphics[width=1\linewidth]{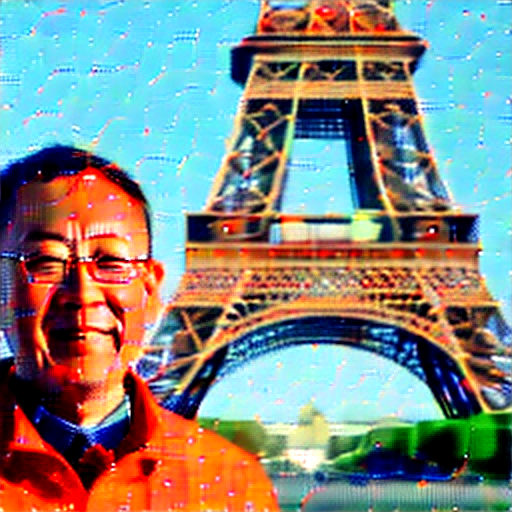}
    \centerline{\scriptsize \textbf{AdvDM}}
\end{minipage}
\begin{minipage}{0.18\linewidth}
    \centering
    \includegraphics[width=1\linewidth]{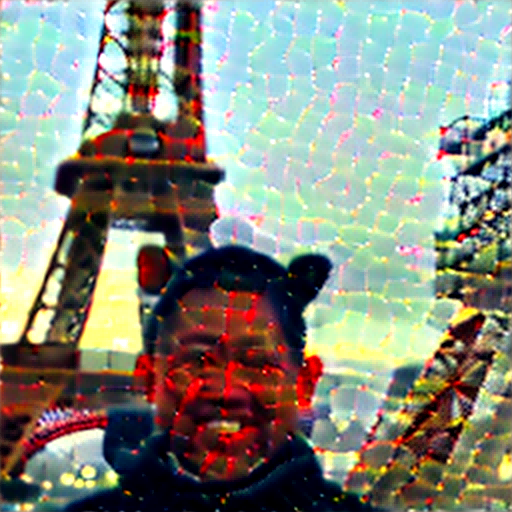}
    \centerline{\scriptsize  \textbf{Anti-DB}}
\end{minipage}
\begin{minipage}{0.18\linewidth}
    \centering
    \includegraphics[width=1\linewidth]{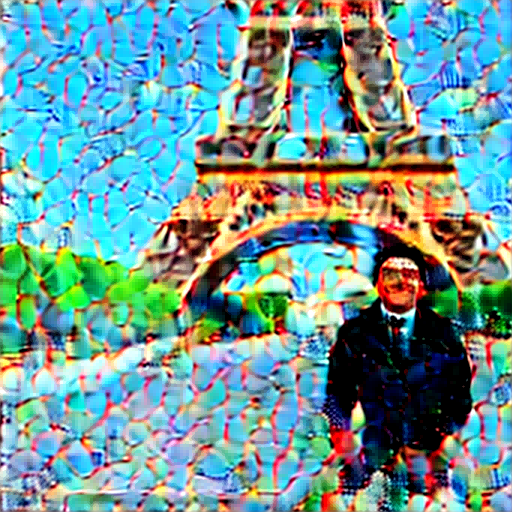}
    \centerline{\scriptsize  \textbf{Anti-DB+SimAC}}
\end{minipage}
\caption{Quantitative results of two people under four prompts on VGG-Face2 dataset. For each person, the first row is ``a photo of sks person'', the second row is ``a dslr portrait of sks person'', the third row is ``a photo of sks person looking at the mirror'' and the last row is ``a photo of sks person in front of eiffel tower". }
\label{fig:vgg qualitative}
\end{figure}

\noindent\textbf{Qualitative Results} 
We show some results in Figure~\ref{fig:Quantitative}. It's evident that SimAC combined with Anti-DreamBooth on the Celeb-HQ dataset achieves a strong image disruption effect, providing the best privacy protection for the input portrait. Since PhotoGuard operates an attack on the latent space, the resulting images tend to generate patterns similar to the target latent. This attack doesn't effectively reduce the probability of facial appearances or improve the quality of generated images. 

Both AdvDM and Anti-DreamBooth use DM's training loss as an objective for optimizing noise, which, as been seen, yields similar results. The two methods aim to disrupt high-frequency components in images via adversarial noise to create artifacts and raise the bar for potential misuse of users' photos. Although the quality of generated images decreases, the results of both AdvDM and Anti-DreamBooth retain many details from the user input images. This indicates a leak of privacy for some users and isn't user-friendly. 
According to our analysis, both of these methods follow the training process of DM, randomly selecting timestamps for optimization. This leads to a waste of some noise addition steps, thereby reducing the efficiency of the attack and the effectiveness of protection. 

On the VGG-Face2 dataset, we select two different individuals for baseline comparison. As shown in Fig ~\ref{fig:vgg qualitative}, although Photoguard, ADVDM, and Anti-Dreambooth perturb the generated images, many characteristics of the user-provided portrait are still present, leading to potential privacy leaks. SimAC+Anti-DreamBooth, however, obscures more facial features and details from the user input images, resulting in the highest level of facial distortion and thereby achieving the most effective user information concealment.

\begin{table}[t]
\scriptsize
\centering
\begin{tabular}{l|l|cccc}
\toprule
\multirow{2}{*}{Train [v]} &\multirow{2}{*}{Test [v]} &\multicolumn{4}{c}{``a photo of [v] person"} \\
\cmidrule{3-6}
&& ISM$\downarrow$& FDFR$\uparrow$ & BRISQUE $\uparrow$& SER-FIQ$\downarrow$ \\
\midrule
sks &sks&0.31 	&87.07 	&38.86 	&0.21 	
  \\
sks &t@t&0.30 	&81.36 	&39.67 	&0.31 
 \\
\bottomrule
\multirow{2}{*}{Train [v]} &\multirow{2}{*}{Test [v]} & \multicolumn{4}{c}{``a dslr portrait of [v] person"}\\
\cmidrule{3-6}
&& ISM$\downarrow$& FDFR$\uparrow$ & BRISQUE $\uparrow$& SER-FIQ$\downarrow$ \\
\midrule
sks &sks&0.12 	&96.87 	&42.10 	&0.15
  \\
sks &t@t&0.23 	&54.42 	&37.77 	&0.52 
 \\
\bottomrule

\multirow{2}{*}{Train [v]} &\multirow{2}{*}{Test [v]}& \multicolumn{4}{c}{``a photo of [v] person looking at the mirror"} \\
\cmidrule{3-6}
&& ISM$\downarrow$& FDFR$\uparrow$ & BRISQUE $\uparrow$& SER-FIQ$\downarrow$ \\
\midrule
sks &sks&0.12 	&91.90 	&43.97 	&0.06 	
 \\
sks &t@t&0.16 &	30.54 	&40.63 	&0.25 \\
\bottomrule
\multirow{2}{*}{Train [v]} &\multirow{2}{*}{Test [v]} & \multicolumn{4}{c}{``a photo of [v] person in front of eiffel tower"}\\
\cmidrule{3-6}
&& ISM$\downarrow$& FDFR$\uparrow$ & BRISQUE $\uparrow$& SER-FIQ$\downarrow$ \\
\midrule
sks &sks&0.05 	&66.19 	&42.77 	&0.12 
 \\
sks &st@t&0.06 	&23.20 &	37.74 &	0.28  \\
\bottomrule
\end{tabular}
\vspace{-1em}
\caption{Prompt mismatch between training and testing on CelebA-HQ. The training prompt is ``a photo of sks person" and the inference prompt uses rare identifier ``sks" or ``t@t".}
\label{tab: prompt mismatch}
\vspace{-1em}
\end{table}

\begin{table}[t]
\scriptsize
\centering
\begin{tabular}{l|l|cccc}
\toprule
\multirow{2}{*}{Train} & \multirow{2}{*}{Test } &\multicolumn{4}{c}{``a photo of sks person"}\\
\cmidrule{3-6}
&& ISM$\downarrow$& FDFR$\uparrow$ & BRISQUE $\uparrow$& SER-FIQ$\downarrow$ \\
\midrule
v2.1 &v2.1&0.31 	&87.07 	&38.86 	&0.21 	
  \\
 v1.4 & v2.1 & 0.38 &70.07 &39.22 &0.35 
\\
v2.1 &v1.4&0.01 	&99.86 	&62.09 &	0.02 	
 \\
\bottomrule
\multirow{2}{*}{Train} & \multirow{2}{*}{Test } & \multicolumn{4}{c}{``a dslr portrait of sks person"}\\
\cmidrule{3-6}
&& ISM$\downarrow$& FDFR$\uparrow$ & BRISQUE $\uparrow$& SER-FIQ$\downarrow$ \\
\midrule
v2.1 &v2.1&0.12 	&96.87 	&42.10 	&0.15
  \\
 v1.4 & v2.1 &0.14 	&95.17 &	40.95 	&0.23 
\\
v2.1 &v1.4&0.26	&20.54 	&16.44 	&0.57 
 \\
\bottomrule

\multirow{2}{*}{Train} & \multirow{2}{*}{Test }& \multicolumn{4}{c}{``a photo of sks person looking at the mirror"} \\
\cmidrule{3-6}
&& ISM$\downarrow$& FDFR$\uparrow$ & BRISQUE $\uparrow$& SER-FIQ$\downarrow$ \\
\midrule
v2.1 &v2.1&0.12 	&91.90 	&43.97 	&0.06 	
 \\
v1.4&v2.1&0.16 	&79.39 	&42.52& 	0.13 	
\\
v2.1 &v1.4& 0.20 	&82.72 	&40.12 	&0.16 	
\\
\bottomrule
\multirow{2}{*}{Train} & \multirow{2}{*}{Test }& \multicolumn{4}{c}{``a photo of sks person in front of eiffel tower"}\\
\cmidrule{3-6}
&& ISM$\downarrow$& FDFR$\uparrow$ & BRISQUE $\uparrow$& SER-FIQ$\downarrow$ \\
\midrule
v2.1 &v2.1&0.05 	&66.19 	&42.77 	&0.12 
 \\
v1.4&v2.1&0.06	&61.36 &	41.51 &	0.10 
\\
v2.1 &v1.4&0.08 	&53.67 	&20.56 	&0.22 
\\
\bottomrule

\end{tabular}
\vspace{-1em}
\caption{Model version mismatch during training and testing on CelebA-HQ. The training prompt is ``a photo of sks person"}
\label{tab: version mismatch}
\vspace{-1em}
\end{table}

\begin{figure}[h]
\centering
\includegraphics[width=1.0\linewidth]{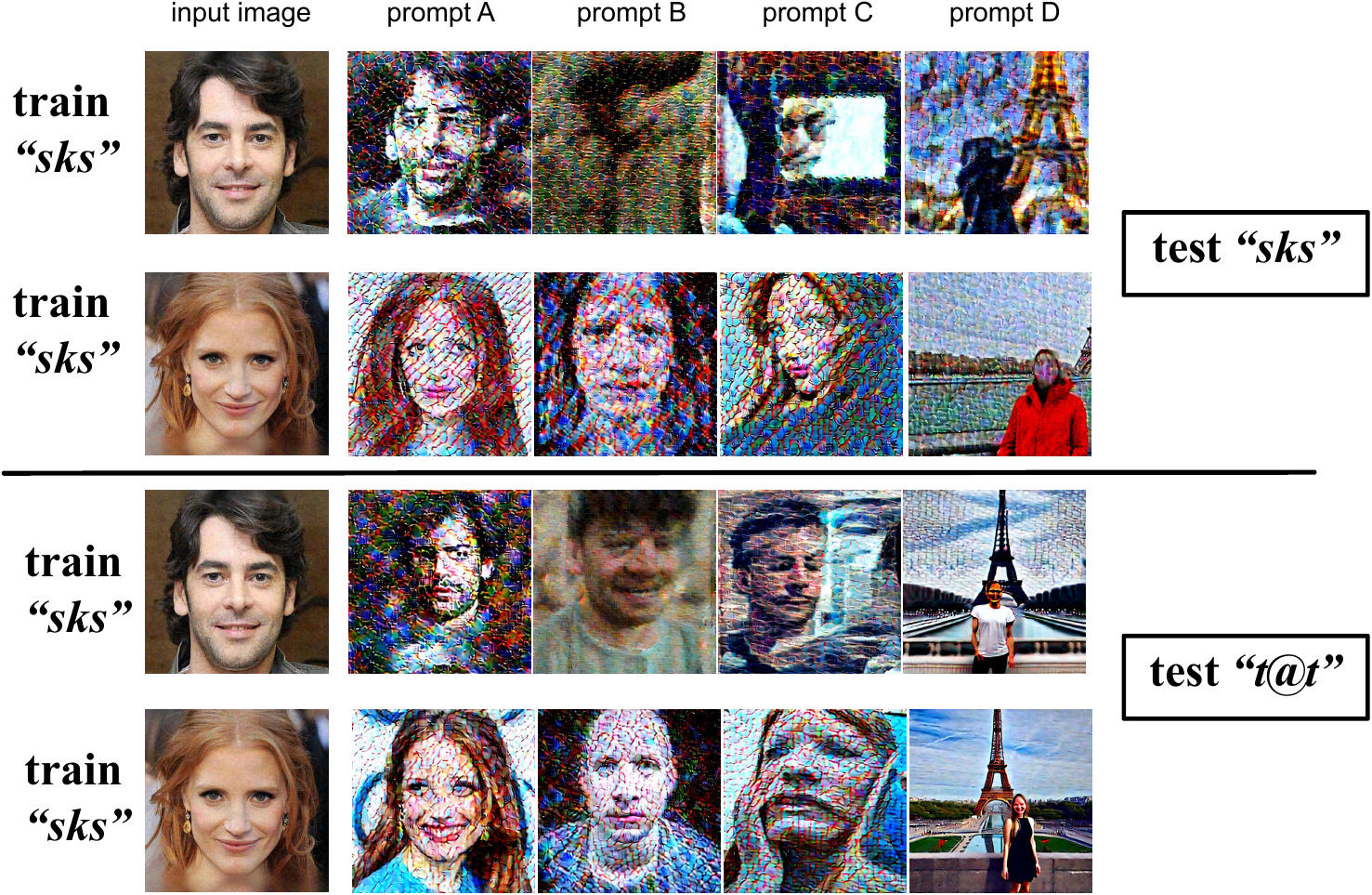}
\vspace{-1em}
\caption{Quantitative results of prompt mismatch under four prompts on CelebA-HQ dataset. The training rare identifier [v] is ``sks" and the customization test rare identifier is ``t@t". This aims to test performance when the prompt used for customization isn't foreseen.}
\label{fig:prompt mismatch}
\end{figure}

\begin{figure}[t]
\centering
\includegraphics[width=1.0\linewidth]{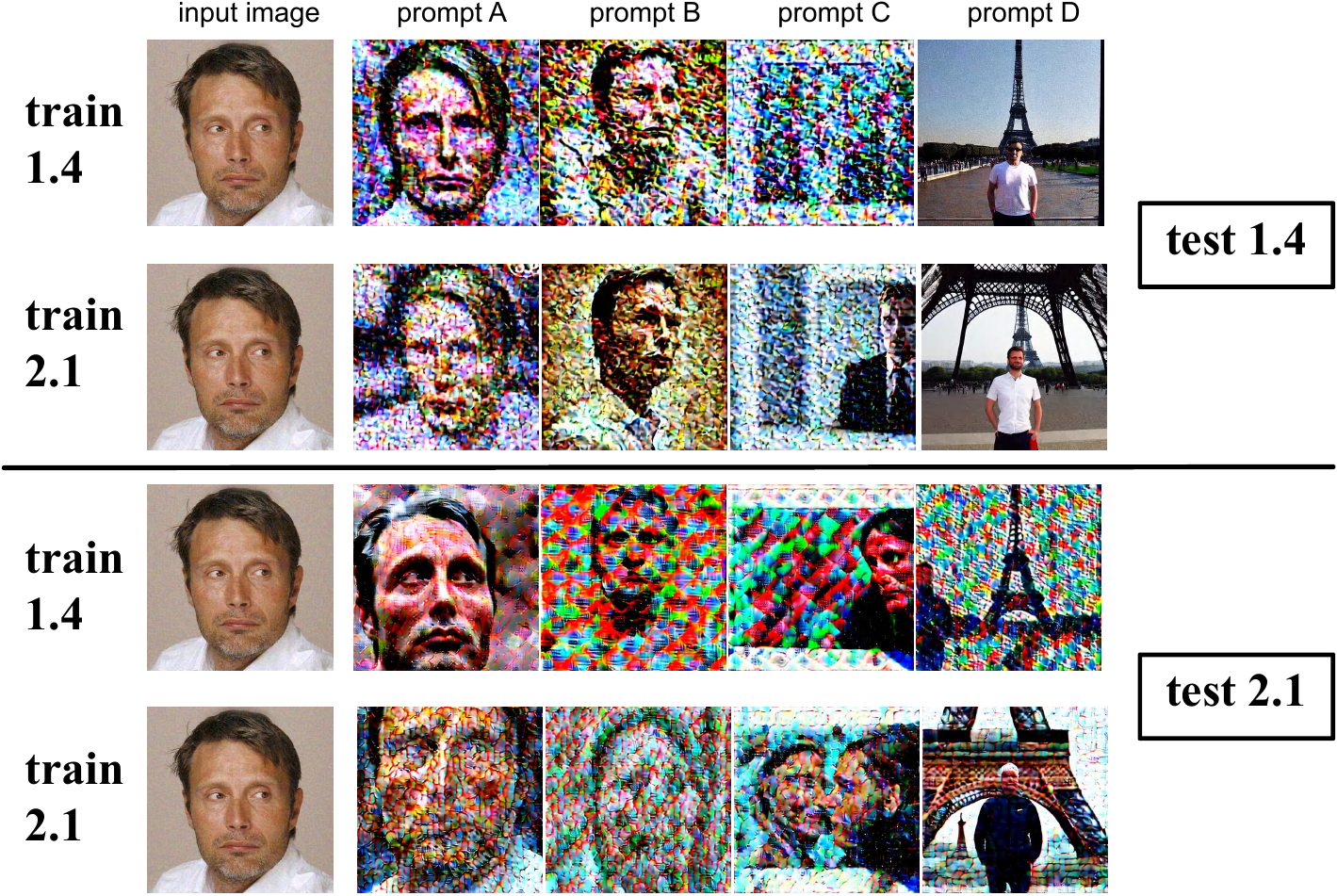}
\vspace{-1em}
\caption{Quantitative results of models mismatch under four prompts on CelebA-HQ dataset. The training uses stable diffusion v1.4 or v2.1, and the testing uses v1.4 and v2.1 in combination with the training model, respectively, to test the sensitivity of the method to the model version.}
\label{fig:model mismatch}
\end{figure}

\begin{figure}[t]
\includegraphics[width=1.0\linewidth]{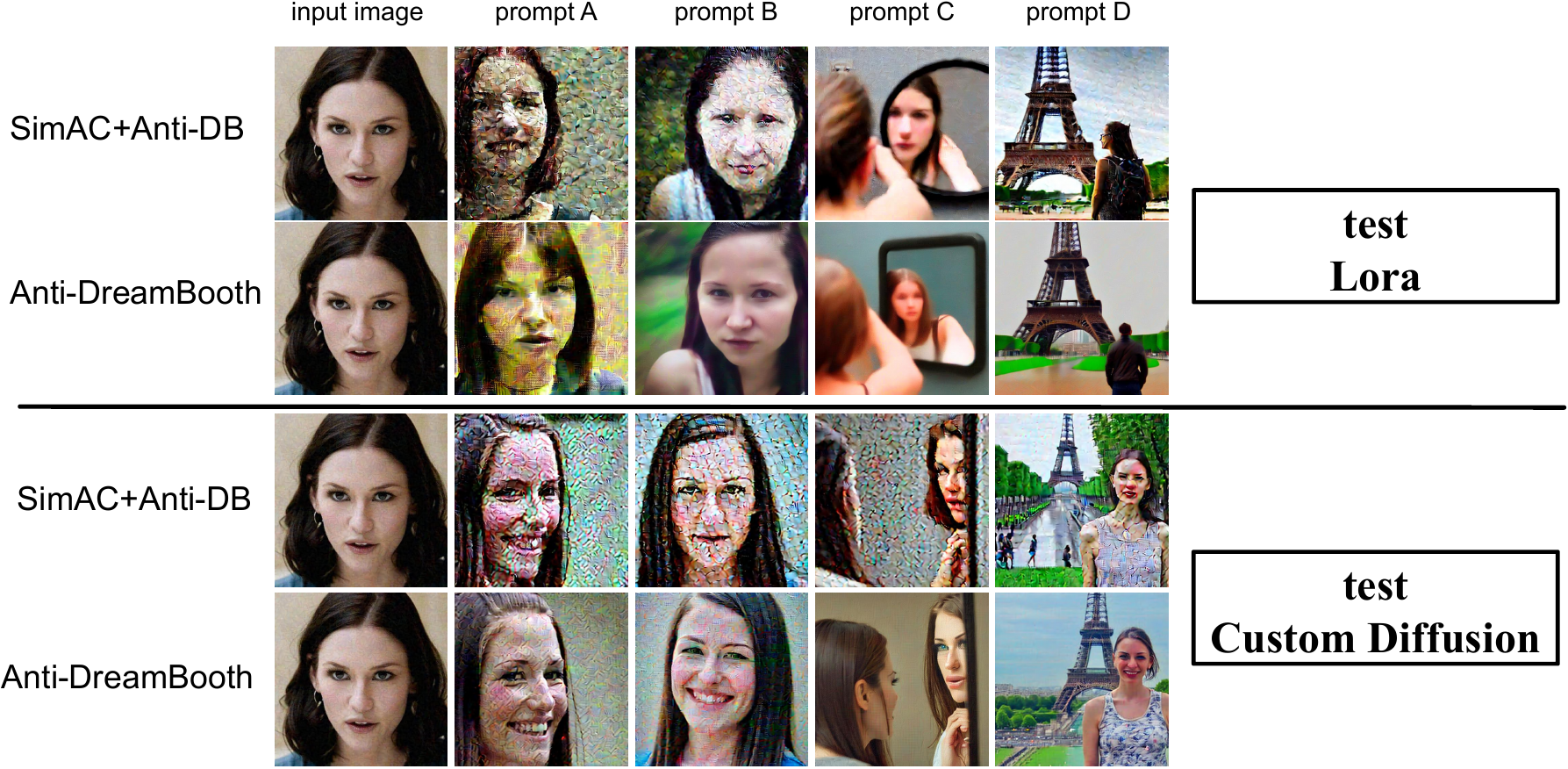}
\vspace{-1em}
\caption{Quantitative results of customization mismatch under four prompts on CelebA-HQ dataset with Anti-DreamBooth and SimAC+Anti-DreamBooth. The training is based on Dreambooth and the customization test is based on Lora or Custom Diffusion. The experiment simulates an attacker using an unknown customization method.}
\label{fig:customization mismatch }
\end{figure}

\subsection{Black-Box Performance}
\subsubsection{Quantitative Results} 
\noindent\textbf{Prompt Mismatch} When attackers use stable diffusion to customize concepts, the prompt they use might differ from our assumptions when adding noise. Thus, we use the ``a photo of sks person" during perturbation learning and change the rare identifier ``sks" to ``t@t" during DreamBooth model finetuing. The results in Table ~\ref{tab: prompt mismatch} indicate a decrease in performance in this scenario, but we notice that the Identity Score Matching(ISM), representing identity similarity remains consistent.

\noindent\textbf{Model Mismatch} The models used to add adversarial noise may also mismatch with the model fine-tuned by DreamBooth. We examine the effectiveness of adversarial noise learned on stable diffusion v2.1 against customization based on stable diffusion v2.1 and v1.4, or vice versa in Table ~\ref{tab: version mismatch}. The conclusion of model mismatch resembles those of prompt mismatch. Although there is an overall decline in performance, the critical matrix, ISM, remains considerable. This also implies that in a model mismatch scenario, SimAC + Anti-DreamBooth can still protect user portrait privacy.

\noindent\textbf{Customization Method Mismatch}
In the previous context, we assume that malicious users fine-tune models based on DreamBooth. However, due to the high training computational cost of DreamBooth, subsequent efforts have been made to reduce the required memory for training while maintaining the fidelity of customization, such as Dreambooth+Lora\cite{hu2021lora} and Custom Diffusion\cite{kumari2023multi}. Hence, we conduct a evaluation for Anti-DreamBooth and SimAC+Anti-DreamBooth when the customization methods mismatch during training and testing. The results in Table~\ref{tab:custom mismatch} are obvious that SimAC+Anti-DreamBooth brings more safeguards than Anti-DreamBooth across different customization strategies. 

\subsubsection{Qualitative result}
In the following pictures, the prompt A is ``a photo of [v] person'', the prompt B is ``a dslr portrait of [v] person'', the prompt C is ``a photo of [v] person looking at the mirror'' and the prompt D is ``a photo of [v] person in front of eiffel tower"

\noindent\textbf{Prompt Mismatch}
We show results in Figure~\ref{fig:prompt mismatch} when different prompts are used for learning noise and fine-tuning stable diffusion (the rare identifiers change from ``sks" to ``t@t"). Despite a slight decline in performance, SimAC+Anti-dreambooth is still able to generate sufficient artifacts to protect users' portrait privacy.

\noindent\textbf{Model Mismatch} 
 We exhibit the results in Figure~\ref{fig:model mismatch}, considering whether the model used for customization is same or different from the model used for adversarial noise. It is evident that the protection is significant when these two model versions are consistent. However, when the versions are different, as seen in cases train v1.4 and test v2.1, where added adversarial noise needs to transfer across different model versions, the protection effect is weakened. Nevertheless, overall, whether the model versions are the same or different, strong distortion is observed in most prompts.

\noindent\textbf{Customization Method Mismatch} Learning adversarial noise based on Dreambooth, and then customizing the protected images based on either Dreambooth+Lora or Custom Diffusion, yields performance decline. However, the identity similarity (ISM) remains at a lower level, offering a degree of user portrait protection. In this setup, enhancing transferred defense against unauthorized customization is an area we will explore in future research. The results, as depicted in Figure~\ref{fig:customization mismatch }, indicate that our method, when transferred to attack Lora-based customization, still performs well. However, when transferred to attack Custom Diffusion-based customization, the artifact effect is not so pronounced. It is worth noting that SimAC+Anti-DreamBooth enhances improve the transer performance of adversarial noise optimized by Anti-DreamBooth in both resisting Lora or Custom Diffusion.

\begin{table}[htbp]
\centering
\begin{tabular}{l|c|c}
\toprule
Method& epoch=50 & epoch=20 \\
\midrule
Anti-DB &3.36$\times 10^{-6}$ &3.26$\times 10^{-6}$ 
  \\
SimAC+Anti-DB &1.23$\times 10^{-5}$&1.14$\times 10^{-5}$
 \\
\bottomrule
\end{tabular}
\caption{The comparison of the averaged mean of absolute gradient values of perturbed images during the training process between SimAC+Anti-DreamBooth and Anti-Dreambooth with training epochs 20 and 50 on the CelebA-HQ dataset.}
\label{tab: gradient comparison}
\end{table}

\begin{table*}[htbp]
\centering
\footnotesize
\begin{tabular}{l|l|cccc|cccc}
\toprule
\multirow{2}{*}{Test } & \multirow{2}{*}{Method } &\multicolumn{4}{c|}{``a photo of sks person"} & \multicolumn{4}{c}{``a dslr portrait of sks person"}\\
\cmidrule{3-10}
&& ISM$\downarrow$& FDFR$\uparrow$ & BRISQUE $\uparrow$& SER-FIQ$\downarrow$ & ISM $\downarrow$& FDFR $\uparrow$ & BRISQUE$\uparrow$& SER-FIQ$\downarrow$\\
\midrule
 \multirow{2}{*}{DreamBooth\cite{ruiz2023dreambooth} }&Anti-DreamBooth &0.28 	&77.28 &	37.43 &	0.20 	&0.19 	&86.80 	&38.90 	&0.27 
  \\
&SimAC+Anti-DreamBooth&0.31 	&87.07 	&38.86 	&0.21 	&0.12 	&96.87 	&42.10 	&0.15
\\
\hline
\multirow{2}{*}{Lora\cite{hu2021lora}}&Anti-DreamBooth& 0.30 	&44.63 &	33.02 &	0.51 	&0.22 	&16.73 	&10.43 	&0.60 

\\
&SimAC+Anti-DreamBooth& 0.19 	&82.86 	&39.99 	&0.35 &	0.20 	&46.94 	&20.19 	&0.54 
\\
\hline
\multirow{2}{*}{Custom Diffusion\cite{kumari2023multi}}&Anti-DreamBooth& 0.60 &	6.39 	&38.90 	&0.73 &	0.46 	&5.99 	&9.83 	&0.75 

\\
&SimAC+Anti-DreamBooth &0.27 &	85.71 &	39.69 	&0.47 &	0.45 	&18.37 &	17.73 	&0.73 

\\
\bottomrule
\end{tabular}
\begin{tabular}{l|l|cccc|cccc}
\toprule
\multirow{2}{*}{Test } & \multirow{2}{*}{Method}& \multicolumn{4}{c|}{``a photo of sks person looking at the mirror"} & \multicolumn{4}{c}{``a photo of sks person in front of eiffel tower"}\\
\cmidrule{3-10}
&& ISM$\downarrow$& FDFR$\uparrow$ & BRISQUE $\uparrow$& SER-FIQ$\downarrow$ & ISM $\downarrow$& FDFR $\uparrow$ & BRISQUE$\uparrow$& SER-FIQ$\downarrow$\\
\midrule
\multirow{2}{*}{DreamBooth \cite{ruiz2023dreambooth}}&Anti-DreamBooth& 0.23 	&42.86 &	40.34 	&0.28 	&0.06 	&56.26 &	41.35 &	0.22 

 \\
&SimAC+Anti-DreamBooth&0.12 	&91.90 	&43.97 	&0.06 	&0.05	&66.19 	&42.77 	&0.12 
\\
\hline
\multirow{2}{*}{Lora\cite{hu2021lora}}	&Anti-DreamBooth& 0.15 	&26.12 	&21.60 	&0.32 	&0.08	&63.61 	&15.31 	&0.27 

\\
&SimAC+Anti-DreamBooth& 0.13 	&18.10 	&21.52 &	0.33 	&0.05 &	64.35 	&26.15 &	0.12 
\\
\hline
\multirow{2}{*}{Custom Diffusion\cite{kumari2023multi}}& Anti-DreamBooth &0.27 &	14.83 &	21.42 	&0.42 &	0.12 	&15.37& 	26.50 &	0.43 
\\
&SimAC+Anti-DreamBooth &0.27 &	33.33 &	32.04 	&0.34 &	0.11 &	17.14 &	34.55 	&0.36 
\\
\bottomrule
\end{tabular}
\caption{Customization strategy mismatch during training and testing on CelebA-HQ dataset, where lower ISM and SER-FIQ are better and higher FDFR and BRISQUE are better. The training is based on Dreambooth and the customization test is based on Lora or Custom Diffusion. }
\label{tab:custom mismatch}
\end{table*}


\begin{figure}[ht]
 \centering
\begin{minipage}{0.48\linewidth}
    \centering
    \includegraphics[width=1.0\linewidth]{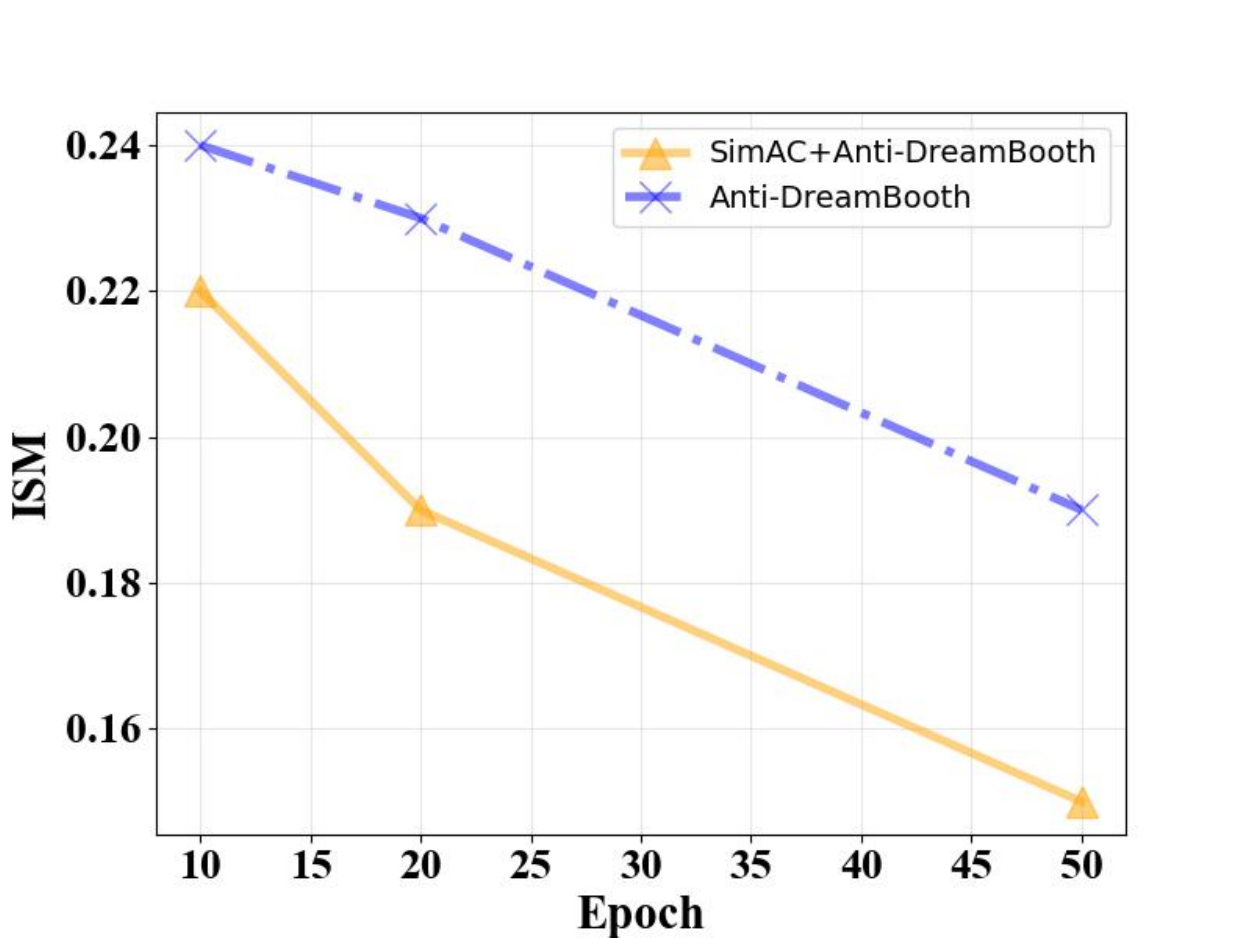}
\end{minipage}
\hfill
\begin{minipage}{0.48\linewidth}
    \centering
    \includegraphics[width=1.0\linewidth]{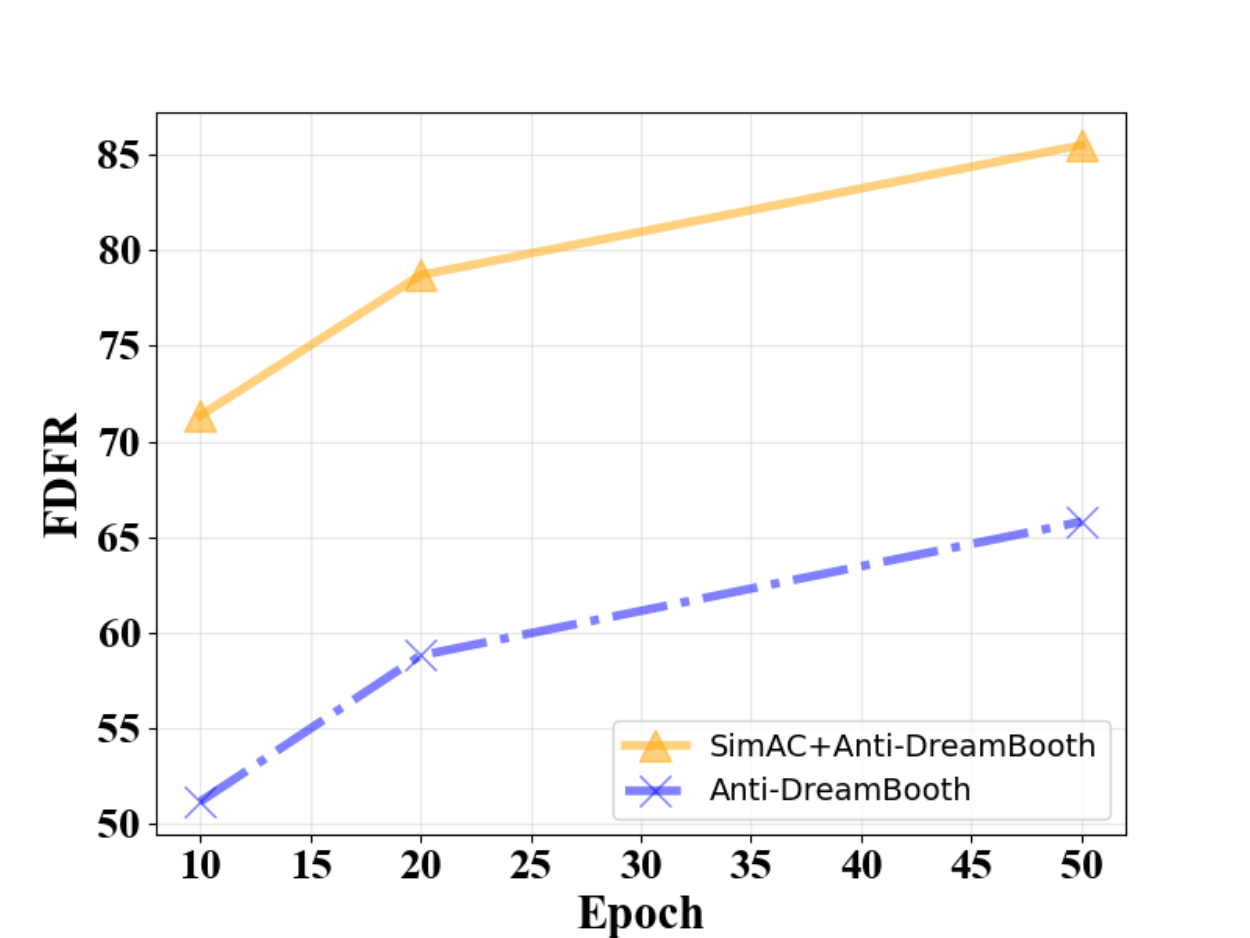}
\end{minipage}
\hfill
\begin{minipage}{0.48\linewidth}
    \centering
    \includegraphics[width=1.0\linewidth]{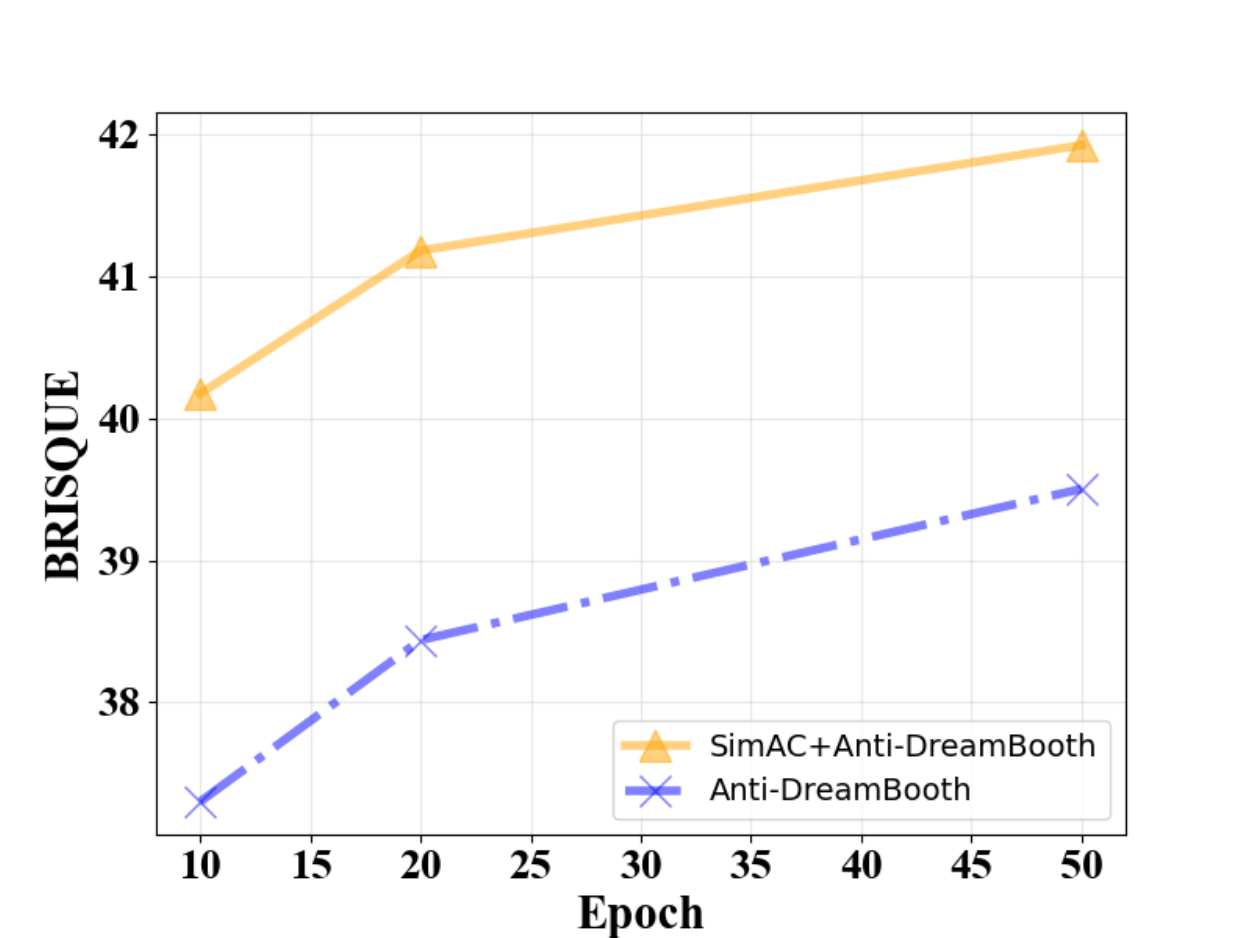}
\end{minipage}
\hfill
\begin{minipage}{0.48\linewidth}
    \centering
    \includegraphics[width=1.0\linewidth]{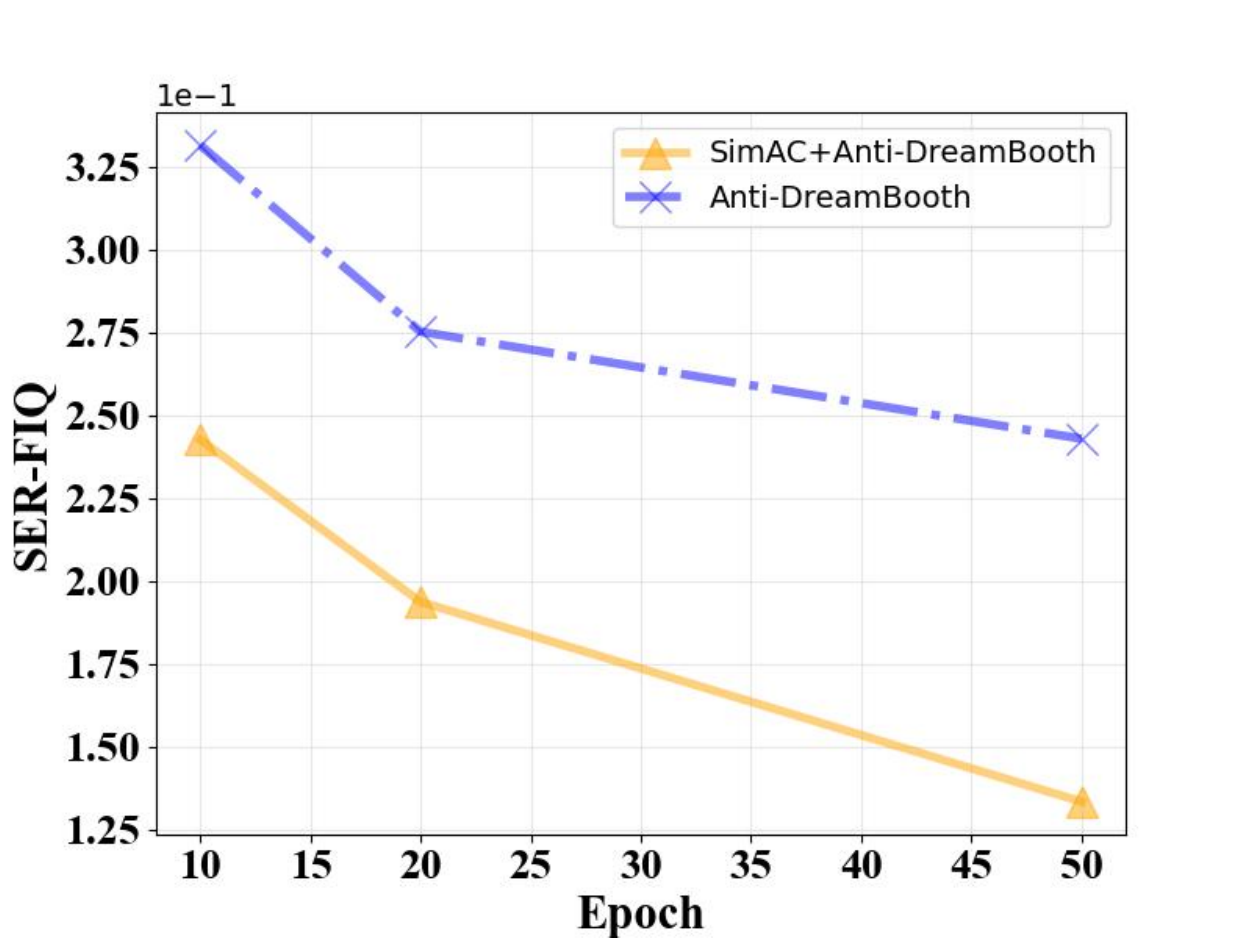}
\end{minipage}
\caption{Setting different training epochs to compare the performance of Anti-DreamBooth and SimAC+Anti-DreamBooth on CelebA-HQ dataset, where lower ISM and SER-FIQ are better and higher FDFR and BRISQUE are better. The comparison aims to examine the training epoch required for both to achieve similar performance, which indicates their efficiency.}
\label{fig:efficiency}
\end{figure}

\section{Anti-Customization Efficiency}
Our method, SimAC+Anti-DreamBooth, outperforms Anti-DreamBooth despite both using the same training epochs. To highlight the effectiveness of SimAC+Anti-DreamBooth by demonstrating its ability to achieve comparable protection to Anti-DreamBooth with fewer training steps, we conduct a comparative study employing training epochs of $(10, 20, 50)$. We ensure a comprehensive evaluation by averaging metrics across four different prompts.

Our findings highlight the significant efficiency boost that SimAC brings to Anti-Dreambooth. As depicted in Figure~\ref{fig:efficiency}, the combined impact of SimAC and Anti-Dreambooth after only 20 training epochs mirrors the performance achieved by Anti-Dreambooth trained for 50 epochs. This observation supports our conclusion that within Anti-Dreambooth's random timestep sampling, some larger timesteps may be selected and cause gradients to approach zero. Consequently, this leads to a failure to sufficiently disrupt the reconstruction fidelity of the original image.

To visually demonstrate the impact of our adaptive greedy time interval selection on enhancing perturbed image gradients during the attack process, we've recorded the averaged mean of absolute gradients for both SimAC+Anti-DreamBooth and Anti-DreamBooth across all training timesteps.

As depicted in Table~\ref{tab: gradient comparison}, whether trained for 20 or 50 epochs, integrating the adaptive time selection approach noticeably enhances gradients during training. This enhancement significantly improves training efficiency and enhances the effectiveness of protection measures countering malicious customization.

\section{GPT-4V(ision) Evaluation}
Beyond employing the conventional deep face recognition technique ArcFace\cite{deng2019arcface} to determine identity similarity between the generated image and the user-provided one, we leverage GPT-4V(ision) to more accurately simulate human judgment regarding identity similarity.

\begin{figure}[h]
\centering
\includegraphics[width=0.6\linewidth]{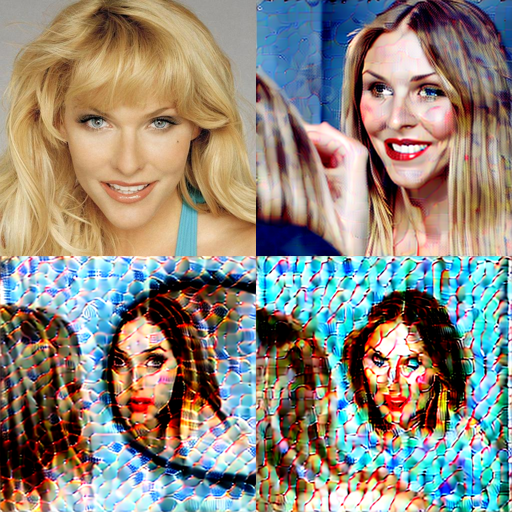}
\vspace{-1em}
\caption{The concatenated image input to GPT-4V(ision) which are generated with the inference prompt ``a photo of sks person looking at the mirror”.}
\label{fig:fourimages}
\end{figure}

We randomly selected images generated from prompts such as ``a photo of sks person", ``a dslr portrait of sks person" and ``a photo of sks person looking at the mirror". And then, selected images are filtered to retain those detectable by Retinaface\cite{deng2020retinaface}.
These images were then categorized based on the corresponding anti-customization methods. For each input, we compile sets of four images, as depicted in Figure ~\ref{fig:fourimages}. The top-left image represents the clean user portrait, while the top-right, bottom-left, and bottom-right sections illustrate outcomes from AdvDM, Anti-DreamBooth, and the combined effect of SimAC and Anti-DreamBooth, respectively.
Subsequently, we obtain a total of 50 images, all of which are fed into GPT-4V(ision)\cite{openai2023gpt4}. The objective is to compare identity similarity with the clean user portrait across the varied results.

We craft the prompts which are the input, as follows in Table~\ref{tab:GPT4}.  The model needs to assess similarity based on a person's facial features and requests a similarity rating within the range of 0 to 10 along with corresponding explanations or justifications. 

The final assessment process 50 images, resulting in valid responses for 35 of them, as detailed in Table ~\ref{tab: gpt-score}. Despite SimAC+Anti-DreamBooth showing a higher facial detection failure rate (FDFR) in comparison to AdvDM and Anti-DreamBooth, its performance within the subset of images containing detectable faces stands out significantly. SimAC+Anti-DreamBooth exhibits the lowest similarity scores and most profound disruption of identity information in portraits among the three methods within this evaluated subset under the judgment of large-scale pre-trained multimodal models GPT-4V(ision).
\begin{table}[h]
\centering
\footnotesize
\begin{tabular}{p{80mm}}
    \toprule
    GPT-4V(ision) Prompt
    \\
    \midrule
    Given the four-square grid image that contains four face images (i.e., the top left image, the top right image, the bottom left image, and the bottom right image), you are required to score the identity similarity between the top left image and each of the remaining three images, respectively. You should pay extra attention to the identity similarity between the persons that appear in the two images, which refers to the similarity of the person's appearance such as facial features, expressions, skin textures, facial ratio, etc.\\
    Please rate the identity similarity on a scale of 0 to 10, where a higher score indicates a higher similarity. The scores are required to have a certain degree of difference.\\
    Please output the scores for the top right image, the bottom left image, and the bottom right image when compared with the top left image. The three scores are separated by a space. Following the scores, please provide an explanation of your evaluation, avoiding any potential bias and ensuring that the order in which the face images were presented does not affect your judgment.\\ \\
    Output format:\\
    Similarity: $<$Scores of the top right image, the bottom left image, and the bottom right image$>$\\
    Reason:\\ 
    \bottomrule
\end{tabular}
\caption{The prompt used for GPT-4V(ision) evaluation.}
\label{tab:GPT4}
\end{table}

\begin{table}[h]
\centering
\begin{tabular}{l|c}
\toprule
Method& GPT-4V(ision) score (total) \\
\midrule
AdvDM\cite{DBLP:conf/icml/LiangWHZXSXMG23} & 86
\\
Anti-DB\cite{van2023anti} &74
  \\
SimAC+Anti-DB &61
 \\
\bottomrule
\end{tabular}
\caption{The comparison of the total similarity scores given by GPT-4V(vision) between AdvDM, Anti-Dreambooth and SimAC+Anti-DreamBooth.}
\label{tab: gpt-score}
\end{table}

\section{Generalizing to Other Anti-Customization Methods}
Our examination delves into the influence of SimAC on other anti-customization methods. Due to PhotoGuard\cite{DBLP:conf/icml/SalmanKLIM23} primarily targeting the VAE encoder attack without involving the selection of timesteps for noise addition during training, we opted to scrutinize the impact of SimAC specifically on AdvDM\cite{DBLP:conf/icml/LiangWHZXSXMG23}.

\begin{table}[ht]
\centering
\footnotesize
\begin{tabular}{l|cccc}
\toprule
\multirow{2}{*}{Method} & \multicolumn{4}{c}{\textcolor{gray}{``a photo of sks person"}} \\
\cmidrule{2-5}
& ISM$\downarrow$& FDFR$\uparrow$ & BRISQUE $\uparrow$& SER-FIQ$\downarrow$ \\
\midrule
AdvDM \cite{DBLP:conf/icml/LiangWHZXSXMG23}&0.32 	&70.48 &	38.17 &	0.20 	
 \\
AdvDM + SimAC &0.45 	&72.18 	&38.03 	&0.41 
	
\\
\bottomrule
\multirow{2}{*}{Method} & \multicolumn{4}{c}{``a dslr portrait of sks person"}\\
\cmidrule{2-5}
& ISM$\downarrow$& FDFR$\uparrow$ & BRISQUE $\uparrow$& SER-FIQ$\downarrow$ \\
\midrule
AdvDM \cite{DBLP:conf/icml/LiangWHZXSXMG23}&0.25	&65.37 &	37.86 &	0.41 
 \\
AdvDM + SimAC &0.27 	&80.48 	&39.21 &	0.44 

\\
\bottomrule
\multirow{2}{*}{Method} & \multicolumn{4}{c}{``a photo of sks person looking at the mirror"} \\
\cmidrule{2-5}
& ISM$\downarrow$& FDFR$\uparrow$ & BRISQUE $\uparrow$& SER-FIQ$\downarrow$ \\
\midrule
AdvDM\cite{DBLP:conf/icml/LiangWHZXSXMG23}  &0.29 &	35.10 &	36.46 	&0.36 
\\
AdvDM + SimAC  &0.26 	&73.61 	&41.18 	&0.21  	 
\\
\bottomrule
\multirow{2}{*}{Method} & \multicolumn{4}{c}{``a photo of sks person in front of eiffel tower"}\\
\cmidrule{2-5}
& ISM$\downarrow$& FDFR$\uparrow$ & BRISQUE $\uparrow$& SER-FIQ$\downarrow$\\
\midrule
AdvDM\cite{DBLP:conf/icml/LiangWHZXSXMG23}  &0.09 &	38.10 &	36.02 	&0.30 
\\
AdvDM + SimAC & 0.08 &	48.10 &	41.79 &	0.19 
\\
\bottomrule
\end{tabular}
\caption{Combining SimAC with AdvDM on the CelebA-HQ dataset improves the performance of AdvDM, where lower ISM and SER-FIQ are better and higher FDFR and BRISQUE are better. The inference prompt in \textcolor{gray}{gray} color is the same as the training prompt.}
\label{tab:adv_simac}
\end{table}

As indicated in Table ~\ref{tab:adv_simac}, our SimAC demonstrates consistent improvements in face detection failure rates across four prompts. Notably, in the latter two prompts, SimAC effectively enhances both identity protection metrics (ISM and FDFR) and the degradation of image quality metrics (BRISQUE and SER-FIQ) achieved by AdvDM. In the second prompt, its performance is comparable to AdvDM's outcomes. However, in the initial prompt, ``a photo of sks person", the combination of AdvDM and SimAC displays less effective performance. This discrepancy in the first inference prompt, which is also utilized during training, suggests that AdvDM might be better suited for this specific prompt, indicating a more obvious reduction in protection effectiveness when encountering unseen prompts. Considering that inferring with unseen prompts aligns more closely with real-world scenarios, the comprehensive integration of SimAC notably enhances the privacy protection performance of AdvDM.

\section{Ablation Study}
\noindent\textbf{Different Layers when Computing The Feature Interference Loss}
To verify our analysis of the properties of LDMs, that is, in UNet decoder blocks, as the layers go deeper, the features pay more attention to high-frequency information and are more prone to adversarial noise, we compare different layer selection in Table~\ref{tab:layer}. The results show that the choice of layer 9,10,11 has the best performance among all groups in four prompts comprehensively and this is consistent with our conclusion that perturbation of deeper features in the UNet decoder will bring more anti-customization performance gains.
\begin{table*}[t]
\centering
\begin{tabular}{l|cccc|cccc}
\toprule
\multirow{2}{*}{layer} & \multicolumn{4}{c|}{``a photo of sks person"} & \multicolumn{4}{c}{``a dslr portrait of sks person"}\\
\cline{2-9}
& ISM$\downarrow$& FDFR$\uparrow$ & BRISQUE $\uparrow$& SER-FQA$\downarrow$ & ISM $\downarrow$& FDFR $\uparrow$ & BRISQUE$\uparrow$& SER-FQA$\downarrow$ \\
\midrule
0,1,2& 0.28 	&86.53 	&39.26 &	0.20 &	0.09 	&96.33 	&42.52 &	0.13 

\\
3,4,5 &0.30 	&84.69 	&38.01 	&0.20 &	0.10 &	96.80 	&42.14 &	0.15 

\\
6,7,8 & 0.30 	&87.35 	&40.14 	&0.21 &	0.15 	&95.17 	&42.00 &	0.16 

 \\
9,10,11&0.31 	&87.07 &	38.86 &	0.21 &	0.12 &	96.87 &	42.10 &	0.15 

\\
\bottomrule
\multirow{2}{*}{layer} & \multicolumn{4}{c|}{``a photo of sks person looking at the mirror"}& \multicolumn{4}{c}{``a photo of sks person in front of eiffel tower"}\\
\cline{2-9}
& ISM$\downarrow$& FDFR$\uparrow$ & BRISQUE $\uparrow$& SER-FQA$\downarrow$ & ISM $\downarrow$& FDFR $\uparrow$ & BRISQUE$\uparrow$& SER-FQA$\downarrow$ \\
\midrule
0,1,2&0.14&88.23&44.68&0.08&0.05&63.95&42.98 &0.12 
\\
3,4,5 &0.12 &89.80 &43.93 &0.08 &0.05&62.93 	&43.12 	&0.11 
\\
6,7,8 &0.14 &90.20 	&44.66 	&0.08 &0.06&65.24&43.73&0.11 
 \\
9,10,11 &0.12 &91.90 	&43.97 &0.06&0.05 &66.19 &42.77 &0.12 
\\
\bottomrule
\end{tabular}
\caption{Comparison of different layer combinations for feature interference loss on CelebA-HQ dataset. We evaluate the performance under four different prompts during customization, where lower ISM and SER-FQA are better and higher FDFR and BRISQUE are better.}
\label{tab:layer}
\end{table*}

\noindent\textbf{Noise Budget} We adjust the noise budget to observe the impact of $\eta$ on defense performance. The results in Table ~\ref{tab: noise budget} show that as the noise budget increases, the defense effectiveness improves. To ensure the image quality of the input image, we select 16/255 as the default noise budget.

\begin{figure}[h]
\centering
\includegraphics[width=1.0\linewidth]{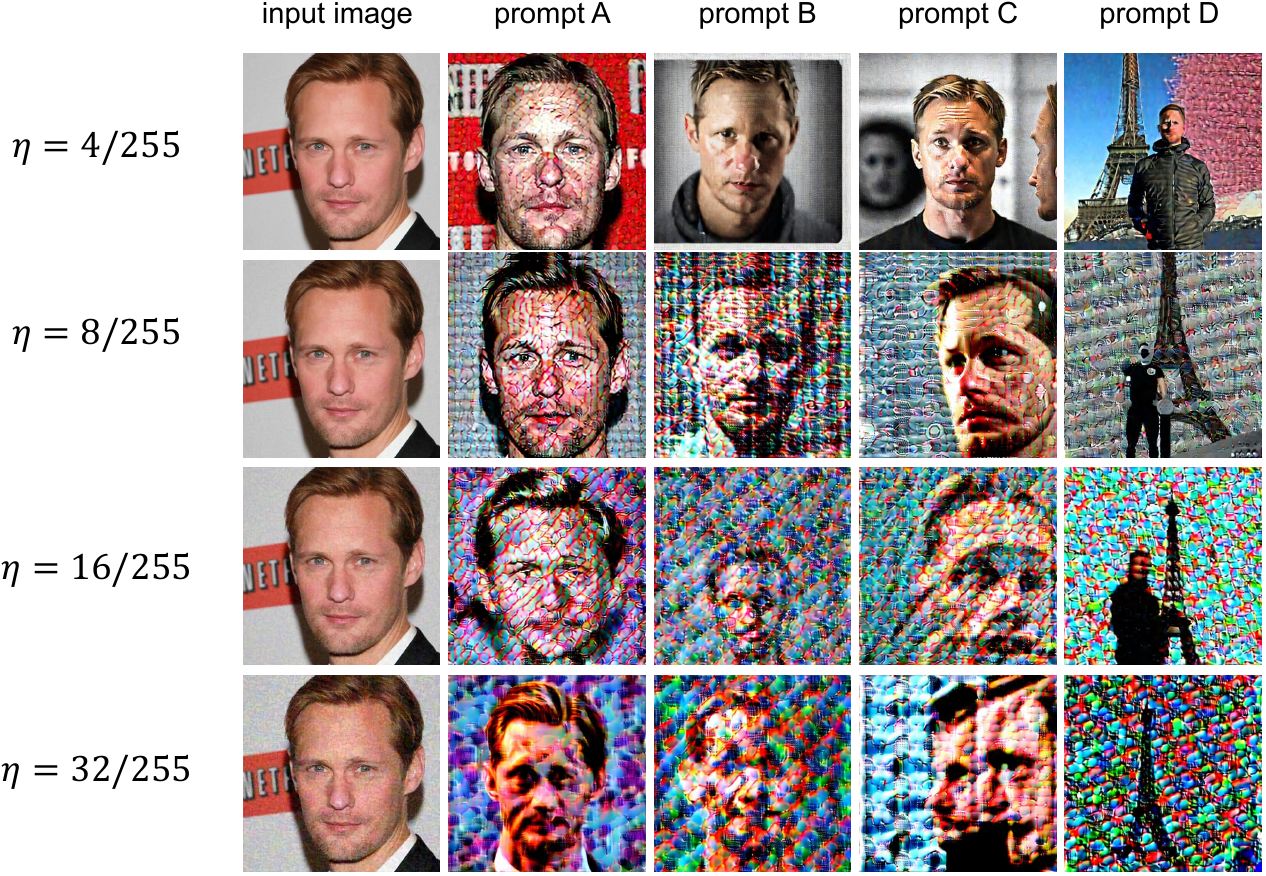}
\vspace{-1em}
\caption{Quantitative results of different noise budgets under four prompts on CelebA-HQ dataset. The training noise budget is constrained to 4/255, 8/255, 16/255 and 32/255. The prompt A is ``a photo of sks person'', the prompt B is ``a dslr portrait of sks person'', the prompt C is ``a photo of sks person looking at the mirror'' and the prompt D is ``a photo of sks person in front of eiffel tower". }
\label{fig:noise budget }
\end{figure}

The quantitative results for the noise budget already demonstrate that with an increase in the noise budget, the performance of anti-customization improves. From a qualitative standpoint, in Figure~\ref{fig:noise budget }, protected user portrait images under different noise budgets are depicted after customization. It can be observed that with a smaller noise budget of 4/255, the protective effect is minimal. However, at a noise budget of 16/255, the effect becomes highly pronounced, meeting our objectives for user identity privacy protection. To strike a balance between image quality and the effect of disruption, we opt for 16/255 instead of choosing a higher level of image degradation at 32/255.

\begin{table}[h]
\centering
\footnotesize
\begin{tabular}{l|cccc}
\toprule
\multirow{2}{*}{$\eta$} & \multicolumn{4}{c}{``a photo of sks person"} \\
\cmidrule{2-5}
& ISM$\downarrow$& FDFR$\uparrow$ & BRISQUE $\uparrow$& SER-FIQ$\downarrow$ \\
\midrule
4/255   &0.60 &	14.15 &	36.11 &	0.70 	
  \\
8/255 &0.47 	&73.06 	&38.81 	&0.38 	 
 \\
16/255 & 0.31 	& 87.07 	& 38.86 	&0.21 	
\\
32/255 &0.22 	&87.21 &	40.86 	&0.14

\\
\bottomrule
\multirow{2}{*}{$\eta$}& \multicolumn{4}{c}{``a dslr portrait of sks person"}\\
\cmidrule{2-5}
& ISM $\downarrow$& FDFR $\uparrow$ & BRISQUE$\uparrow$& SER-FIQ$\downarrow$\\
\midrule
4/255   &0.48 	&29.86 &	17.66 	&0.74 
  \\
8/255 &0.32 	&74.56 	&41.14 	&0.45 
 \\
16/255& 0.12 &	96.87 	& 42.10 &	0.15
\\
32/255 &0.04 &	98.71 	&41.28 	&0.04 
\\
\bottomrule
\multirow{2}{*}{$\eta$} & \multicolumn{4}{c}{``a photo of sks person looking at the mirror"} \\
\cmidrule{2-5}
& ISM$\downarrow$& FDFR$\uparrow$ & BRISQUE $\uparrow$& SER-FIQ$\downarrow$ \\
\midrule
4/255  &0.42 	&17.69 &	28.21 &	0.56 	
 \\
8/255 &0.29 	&64.08 	&42.58 	&0.26 	

\\
16/255 & 0.12 &	91.90 	& 43.97 	& 0.06 
\\
32/255 &0.07 	&94.29 &	44.02 	&0.03 
\\
\bottomrule
\multirow{2}{*}{$\eta$} & \multicolumn{4}{c}{``a photo of sks person in front of eiffel tower"} 
\\
\cmidrule{2-5}
& ISM$\downarrow$& FDFR$\uparrow$ & BRISQUE $\uparrow$& SER-FIQ$\downarrow$ \\
\midrule
4/255  &0.14 &	29.46 	&30.45 &	0.37 
 \\
8/255 &0.08	&40.88 	&41.45 	&0.22 

\\
16/255 &	0.05 &	66.19 & 	42.77 &	0.12
\\
32/255 &	0.04 	&71.90 &	41.78 &	0.07 
\\
\bottomrule
\end{tabular}
\caption{Different noise budget based on SimAC on CelebA-HQ dataset, where lower ISM and SER-FIQ are better and higher FDFR and BRISQUE are better. As the noise budget increases, the deteriorating effect on the generated image increases.}
\label{tab: noise budget}
\end{table}

\section{Conclusion}
Current anti-customization methods, primarily relying on adversarial attacks, often overlook crucial internal properties of the diffusion model, leading to ineffective optimization. 
This paper addresses this issue by exploring inherent properties to enhance anti-customization. 
Two key aspects are examined:
Analyzing the relationship between timestep selection and the model's perception in the frequency domain inspires an adaptive greedy search for optimal timesteps.
Scrutinizing feature roles during denoising, resulting in a sophisticated feature-based optimization framework. 
Experiments show increased identity disruption, enhancing user privacy and security.

{
    \small
    \bibliographystyle{ieeenat_fullname}
    \bibliography{main}
}

\end{document}

%% file: preamble.tex
\usepackage[dvipsnames]{xcolor}


%% file: main.bbl
\begin{thebibliography}{39}
\providecommand{\natexlab}[1]{#1}
\providecommand{\url}[1]{\texttt{#1}}
\expandafter\ifx\csname urlstyle\endcsname\relax
  \providecommand{\doi}[1]{doi: #1}\else
  \providecommand{\doi}{doi: \begingroup \urlstyle{rm}\Url}\fi

\bibitem[sdv()]{sdv2.1}
Stable-diffusion-2-1. \url{https://huggingface.co/stabilityai/stable-diffusion-2-1}.

\bibitem[tru()]{trump}
Making pictures of trump getting arrested while waiting for trump's arrest. \url{https://twitter.com/EliotHiggins/status/1637927681734987777}.

\bibitem[Cao et~al.(2018)Cao, Shen, Xie, Parkhi, and Zisserman]{cao2018vggface2}
Qiong Cao, Li Shen, Weidi Xie, Omkar~M Parkhi, and Andrew Zisserman.
\newblock Vggface2: A dataset for recognising faces across pose and age.
\newblock In \emph{2018 13th IEEE international conference on automatic face \& gesture recognition (FG 2018)}, pages 67--74. IEEE, 2018.

\bibitem[Deng et~al.(2019)Deng, Guo, Xue, and Zafeiriou]{deng2019arcface}
Jiankang Deng, Jia Guo, Niannan Xue, and Stefanos Zafeiriou.
\newblock Arcface: Additive angular margin loss for deep face recognition.
\newblock In \emph{Proceedings of the IEEE/CVF Conference on Computer Vision and Pattern Recognition}, pages 4690--4699, 2019.

\bibitem[Deng et~al.(2020)Deng, Guo, Ververas, Kotsia, and Zafeiriou]{deng2020retinaface}
Jiankang Deng, Jia Guo, Evangelos Ververas, Irene Kotsia, and Stefanos Zafeiriou.
\newblock Retinaface: Single-shot multi-level face localisation in the wild.
\newblock In \emph{Proceedings of the IEEE/CVF Conference on Computer Vision and Pattern Recognition}, pages 5203--5212, 2020.

\bibitem[Dosovitskiy et~al.(2020)Dosovitskiy, Beyer, Kolesnikov, Weissenborn, Zhai, Unterthiner, Dehghani, Minderer, Heigold, Gelly, et~al.]{dosovitskiy2020image}
Alexey Dosovitskiy, Lucas Beyer, Alexander Kolesnikov, Dirk Weissenborn, Xiaohua Zhai, Thomas Unterthiner, Mostafa Dehghani, Matthias Minderer, Georg Heigold, Sylvain Gelly, et~al.
\newblock An image is worth 16x16 words: Transformers for image recognition at scale.
\newblock \emph{arXiv preprint arXiv:2010.11929}, 2020.

\bibitem[Gal et~al.(2022)Gal, Alaluf, Atzmon, Patashnik, Bermano, Chechik, and Cohen-Or]{gal2022image}
Rinon Gal, Yuval Alaluf, Yuval Atzmon, Or Patashnik, Amit~H Bermano, Gal Chechik, and Daniel Cohen-Or.
\newblock An image is worth one word: Personalizing text-to-image generation using textual inversion.
\newblock \emph{arXiv preprint arXiv:2208.01618}, 2022.

\bibitem[Goodfellow et~al.(2014)Goodfellow, Pouget-Abadie, Mirza, Xu, Warde-Farley, Ozair, Courville, and Bengio]{goodfellow2014generative}
Ian Goodfellow, Jean Pouget-Abadie, Mehdi Mirza, Bing Xu, David Warde-Farley, Sherjil Ozair, Aaron Courville, and Yoshua Bengio.
\newblock Generative adversarial nets.
\newblock \emph{Advances in neural information processing systems}, 27, 2014.

\bibitem[Goodfellow et~al.(2015)Goodfellow, Shlens, and Szegedy]{goodfellow2015explaining}
Ian~J. Goodfellow, Jonathon Shlens, and Christian Szegedy.
\newblock Explaining and harnessing adversarial examples.
\newblock In \emph{International Conference on Learning Representations (ICLR)}, 2015.

\bibitem[Gulrajani et~al.(2017)Gulrajani, Ahmed, Arjovsky, Dumoulin, and Courville]{gulrajani2017improved}
Ishaan Gulrajani, Faruk Ahmed, Martin Arjovsky, Vincent Dumoulin, and Aaron~C Courville.
\newblock Improved training of wasserstein gans.
\newblock \emph{Advances in neural information processing systems}, 30, 2017.

\bibitem[Ho and Salimans(2021)]{ho2021classifier}
Jonathan Ho and Tim Salimans.
\newblock Classifier-free diffusion guidance.
\newblock In \emph{NeurIPS 2021 Workshop on Deep Generative Models and Downstream Applications}, 2021.

\bibitem[Ho et~al.(2020)Ho, Jain, and Abbeel]{ho2020denoising}
Jonathan Ho, Ajay Jain, and Pieter Abbeel.
\newblock Denoising diffusion probabilistic models.
\newblock \emph{Advances in neural information processing systems}, 33:\penalty0 6840--6851, 2020.

\bibitem[Hu et~al.(2021)Hu, Shen, Wallis, Allen-Zhu, Li, Wang, Wang, and Chen]{hu2021lora}
Edward~J Hu, Yelong Shen, Phillip Wallis, Zeyuan Allen-Zhu, Yuanzhi Li, Shean Wang, Lu Wang, and Weizhu Chen.
\newblock Lora: Low-rank adaptation of large language models.
\newblock \emph{arXiv preprint arXiv:2106.09685}, 2021.

\bibitem[Huang et~al.(2021)Huang, Zhang, Zhou, Zhang, and Yu]{huang2021initiative}
Qidong Huang, Jie Zhang, Wenbo Zhou, Weiming Zhang, and Nenghai Yu.
\newblock Initiative defense against facial manipulation.
\newblock In \emph{Proceedings of the AAAI Conference on Artificial Intelligence}, pages 1619--1627, 2021.

\bibitem[Huang et~al.(2022{\natexlab{a}})Huang, Dong, Chen, Zhou, Zhang, and Yu]{huang2022shape}
Qidong Huang, Xiaoyi Dong, Dongdong Chen, Hang Zhou, Weiming Zhang, and Nenghai Yu.
\newblock Shape-invariant 3d adversarial point clouds.
\newblock In \emph{Proceedings of the IEEE/CVF conference on computer vision and pattern recognition}, pages 15335--15344, 2022{\natexlab{a}}.

\bibitem[Huang et~al.(2022{\natexlab{b}})Huang, Dong, Chen, Zhou, Zhang, Zhang, Hua, and Yu]{huang2022pointcat}
Qidong Huang, Xiaoyi Dong, Dongdong Chen, Hang Zhou, Weiming Zhang, Kui Zhang, Gang Hua, and Nenghai Yu.
\newblock Pointcat: Contrastive adversarial training for robust point cloud recognition.
\newblock \emph{arXiv preprint arXiv:2209.07788}, 2022{\natexlab{b}}.

\bibitem[Huang et~al.(2023{\natexlab{a}})Huang, Dong, Chen, Chen, Yuan, Hua, Zhang, and Yu]{huang2023improving}
Qidong Huang, Xiaoyi Dong, Dongdong Chen, Yinpeng Chen, Lu Yuan, Gang Hua, Weiming Zhang, and Nenghai Yu.
\newblock Improving adversarial robustness of masked autoencoders via test-time frequency-domain prompting.
\newblock In \emph{Proceedings of the IEEE/CVF International Conference on Computer Vision}, pages 1600--1610, 2023{\natexlab{a}}.

\bibitem[Huang et~al.(2023{\natexlab{b}})Huang, Dong, Chen, Zhang, Wang, Hua, and Yu]{huang2023diversity}
Qidong Huang, Xiaoyi Dong, Dongdong Chen, Weiming Zhang, Feifei Wang, Gang Hua, and Nenghai Yu.
\newblock Diversity-aware meta visual prompting.
\newblock In \emph{Proceedings of the IEEE/CVF Conference on Computer Vision and Pattern Recognition}, pages 10878--10887, 2023{\natexlab{b}}.

\bibitem[Huang et~al.(2023{\natexlab{c}})Huang, Dong, Zhang, Wang, He, Wang, Lin, Zhang, and Yu]{huang2023opera}
Qidong Huang, Xiaoyi Dong, Pan Zhang, Bin Wang, Conghui He, Jiaqi Wang, Dahua Lin, Weiming Zhang, and Nenghai Yu.
\newblock Opera: Alleviating hallucination in multi-modal large language models via over-trust penalty and retrospection-allocation.
\newblock \emph{arXiv preprint arXiv:2311.17911}, 2023{\natexlab{c}}.

\bibitem[Karras et~al.(2018)Karras, Aila, Laine, and Lehtinen]{DBLP:conf/iclr/celeb}
Tero Karras, Timo Aila, Samuli Laine, and Jaakko Lehtinen.
\newblock Progressive growing of gans for improved quality, stability, and variation.
\newblock In \emph{6th International Conference on Learning Representations, {ICLR} 2018, Vancouver, BC, Canada, April 30 - May 3, 2018, Conference Track Proceedings}. OpenReview.net, 2018.

\bibitem[Kingma and Welling(2013)]{kingma2013auto}
Diederik~P Kingma and Max Welling.
\newblock Auto-encoding variational bayes.
\newblock \emph{arXiv preprint arXiv:1312.6114}, 2013.

\bibitem[Kumari et~al.(2023)Kumari, Zhang, Zhang, Shechtman, and Zhu]{kumari2023multi}
Nupur Kumari, Bingliang Zhang, Richard Zhang, Eli Shechtman, and Jun-Yan Zhu.
\newblock Multi-concept customization of text-to-image diffusion.
\newblock In \emph{Proceedings of the IEEE/CVF Conference on Computer Vision and Pattern Recognition}, pages 1931--1941, 2023.

\bibitem[Liang et~al.()Liang, Wu, Hua, Zhang, Xue, Song, Xue, Ma, and Guan]{DBLP:conf/icml/LiangWHZXSXMG23}
Chumeng Liang, Xiaoyu Wu, Yang Hua, Jiaru Zhang, Yiming Xue, Tao Song, Zhengui Xue, Ruhui Ma, and Haibing Guan.
\newblock Adversarial example does good: Preventing painting imitation from diffusion models via adversarial examples.
\newblock In \emph{International Conference on Machine Learning, {ICML} 2023}, pages 20763--20786.

\bibitem[Madry et~al.()Madry, Makelov, Schmidt, Tsipras, and Vladu]{DBLP:conf/iclr/MadryMSTV18}
Aleksander Madry, Aleksandar Makelov, Ludwig Schmidt, Dimitris Tsipras, and Adrian Vladu.
\newblock Towards deep learning models resistant to adversarial attacks.
\newblock In \emph{6th International Conference on Learning Representations, {ICLR} 2018}.

\bibitem[Mirza and Osindero(2014)]{mirza2014conditional}
Mehdi Mirza and Simon Osindero.
\newblock Conditional generative adversarial nets.
\newblock \emph{arXiv preprint arXiv:1411.1784}, 2014.

\bibitem[Mittal et~al.(2012)Mittal, Moorthy, and Bovik]{Mittal_Moorthy_Bovik_2012}
A. Mittal, A.~K. Moorthy, and A.~C. Bovik.
\newblock No-reference image quality assessment in the spatial domain.
\newblock \emph{IEEE Transactions on Image Processing}, page 4695–4708, 2012.

\bibitem[OpenAI(2023)]{openai2023gpt4}
OpenAI.
\newblock Gpt-4 technical report, 2023.

\bibitem[Rombach et~al.(2022)Rombach, Blattmann, Lorenz, Esser, and Ommer]{rombach2022high}
Robin Rombach, Andreas Blattmann, Dominik Lorenz, Patrick Esser, and Bj{\"o}rn Ommer.
\newblock High-resolution image synthesis with latent diffusion models.
\newblock In \emph{Proceedings of the IEEE/CVF conference on computer vision and pattern recognition}, pages 10684--10695, 2022.

\bibitem[Ruiz et~al.(2023)Ruiz, Li, Jampani, Pritch, Rubinstein, and Aberman]{ruiz2023dreambooth}
Nataniel Ruiz, Yuanzhen Li, Varun Jampani, Yael Pritch, Michael Rubinstein, and Kfir Aberman.
\newblock Dreambooth: Fine tuning text-to-image diffusion models for subject-driven generation.
\newblock In \emph{Proceedings of the IEEE/CVF Conference on Computer Vision and Pattern Recognition}, pages 22500--22510, 2023.

\bibitem[Salman et~al.()Salman, Khaddaj, Leclerc, Ilyas, and Madry]{DBLP:conf/icml/SalmanKLIM23}
Hadi Salman, Alaa Khaddaj, Guillaume Leclerc, Andrew Ilyas, and Aleksander Madry.
\newblock Raising the cost of malicious ai-powered image editing.
\newblock In \emph{International Conference on Machine Learning, {ICML} 2023}, pages 29894--29918.

\bibitem[Shan et~al.(2023)Shan, Cryan, Wenger, Zheng, Hanocka, and Zhao]{shan2023glaze}
Shawn Shan, Jenna Cryan, Emily Wenger, Haitao Zheng, Rana Hanocka, and Ben~Y Zhao.
\newblock Glaze: Protecting artists from style mimicry by text-to-image models.
\newblock \emph{arXiv preprint arXiv:2302.04222}, 2023.

\bibitem[Sohl-Dickstein et~al.(2015)Sohl-Dickstein, Weiss, Maheswaranathan, and Ganguli]{sohl2015deep}
Jascha Sohl-Dickstein, Eric Weiss, Niru Maheswaranathan, and Surya Ganguli.
\newblock Deep unsupervised learning using nonequilibrium thermodynamics.
\newblock In \emph{International conference on machine learning}, pages 2256--2265. PMLR, 2015.

\bibitem[Song et~al.(2021)Song, Meng, and Ermon]{DBLP:conf/iclr/SongME21}
Jiaming Song, Chenlin Meng, and Stefano Ermon.
\newblock Denoising diffusion implicit models.
\newblock In \emph{9th International Conference on Learning Representations, {ICLR}}, 2021.

\bibitem[Szegedy et~al.(2013)Szegedy, Zaremba, Sutskever, Bruna, Erhan, Goodfellow, and Fergus]{szegedy2013intriguing}
Christian Szegedy, Wojciech Zaremba, Ilya Sutskever, Joan Bruna, Dumitru Erhan, Ian Goodfellow, and Rob Fergus.
\newblock Intriguing properties of neural networks.
\newblock In \emph{International Conference on Learning Representations (ICLR)}, 2013.

\bibitem[Terh{\"{o}}rst et~al.(2020)Terh{\"{o}}rst, Kolf, Damer, Kirchbuchner, and Kuijper]{DBLP:conf/cvpr/TerhorstKDKK20}
Philipp Terh{\"{o}}rst, Jan~Niklas Kolf, Naser Damer, Florian Kirchbuchner, and Arjan Kuijper.
\newblock {SER-FIQ:} unsupervised estimation of face image quality based on stochastic embedding robustness.
\newblock In \emph{2020 {IEEE/CVF} Conference on Computer Vision and Pattern Recognition, {CVPR} 2020, Seattle, WA, USA, June 13-19, 2020}, pages 5650--5659. {IEEE}, 2020.

\bibitem[Tumanyan et~al.(2023)Tumanyan, Geyer, Bagon, and Dekel]{tumanyan2023plug}
Narek Tumanyan, Michal Geyer, Shai Bagon, and Tali Dekel.
\newblock Plug-and-play diffusion features for text-driven image-to-image translation.
\newblock In \emph{Proceedings of the IEEE/CVF Conference on Computer Vision and Pattern Recognition}, pages 1921--1930, 2023.

\bibitem[Van~Le et~al.(2023)Van~Le, Phung, Nguyen, Dao, Tran, and Tran]{van2023anti}
Thanh Van~Le, Hao Phung, Thuan~Hoang Nguyen, Quan Dao, Ngoc~N Tran, and Anh Tran.
\newblock Anti-dreambooth: Protecting users from personalized text-to-image synthesis.
\newblock In \emph{Proceedings of the IEEE/CVF International Conference on Computer Vision}, pages 2116--2127, 2023.

\bibitem[Zhang et~al.(2023)Zhang, Han, Ghosh, Metaxas, and Ren]{zhang2023sine}
Zhixing Zhang, Ligong Han, Arnab Ghosh, Dimitris~N Metaxas, and Jian Ren.
\newblock Sine: Single image editing with text-to-image diffusion models.
\newblock In \emph{Proceedings of the IEEE/CVF Conference on Computer Vision and Pattern Recognition}, pages 6027--6037, 2023.

\bibitem[Zhu et~al.(2017)Zhu, Park, Isola, and Efros]{zhu2017unpaired}
Jun-Yan Zhu, Taesung Park, Phillip Isola, and Alexei~A Efros.
\newblock Unpaired image-to-image translation using cycle-consistent adversarial networks.
\newblock In \emph{Proceedings of the IEEE international conference on computer vision}, pages 2223--2232, 2017.

\end{thebibliography}
